\documentclass[preprint,authoryear]{elsarticle}
\journal{Applied Energy}
\usepackage[top=3cm, bottom=3cm, left=3cm, right=3cm]{geometry}
\usepackage{lmodern}
\usepackage[utf8]{inputenc}
\usepackage[english]{babel}
\usepackage[T1]{fontenc}
\usepackage{algorithm}
\usepackage{algorithmic}
\usepackage{ dsfont }
\usepackage{amsmath}
\usepackage{amsthm}
\usepackage{float}
\usepackage{amsfonts}
\usepackage{amssymb}
\usepackage{todonotes}
\usepackage{graphicx}
\usepackage{ mathrsfs }
\usepackage[hidelinks]{hyperref}
\usepackage{cancel}
\usepackage{comment}
\usepackage{ marvosym }
\usepackage{bm} 
\usepackage{enumitem} 
\usepackage{comment}
\usepackage{etoolbox}
\usepackage{subcaption}
\usepackage{tikz}
\usepackage{multirow}
\usetikzlibrary{arrows.meta, positioning, shapes,fit, calc, decorations.markings}

\usetikzlibrary{arrows}
\usepackage{comment}

\newcommand{\Rr}{{\mathbb{R}}}

\date{}

\begin{document}

\begin{frontmatter}

\title{Parametric and Generative Forecasts of EPEX Day\char45 Ahead Energy Market Curves}
\tnotetext[t1]{This work has been carried out at the Energy4Climate Interdisciplinary Center (E4C) of IP Paris, which is in part supported by 3rd Programme d'Investissements d'Avenir [ANR\char45 18\char45 EUR\char45 0006\char45 02], and by the Foundation of Ecole polytechnique (Chaire “D\'{e}carboner l'\'{e}conomie” financed by BNP Paribas). Part of this work was conducted while Julian Gutierrez held a postdoctoral position at CREST, ENSAE, Institut Polytechnique de Paris.}

\author[nyuad]{Julian Gutierrez}
\ead{jg8960@nyu.edu, julian.gutierrezpin@gmail.com}

\author[crest]{Redouane Silvente}
\ead{redsilvente@gmail.com} 

\address[nyuad]{New York University Abu Dhabi, Saadiyat Island, PO Box 129188, Abu Dhabi, United Arab Emirates}

\address[crest]{CREST, ENSAE, Institut Polytechnique de Paris, 5 Avenue Henry Le Chatelier
91120 Palaiseau, France}

\cortext[cor1]{Corresponding author: Julian Gutierrez (julian.gutierrezpin@gmail.com)}

\begin{abstract}
We propose two methodologies for modelling aggregated supply and demand curves in the EPEX SPOT Day\char45 Ahead market, emphasizing generative models as a way to recover distributional variability. The first is a low\char45 dimensional parametric representation that yields deterministic point forecasts; the second is a high\char45 dimensional order\char45 level representation that samples from a conditional distribution of plausible curves. Both model the full curve structure, enabling the analysis of price sensitivity, volume sensitivity, and price impact.

The parametric representation uses plateau levels, elastic\char45 region boundaries, and polynomial coefficients, forecast with eXtreme Gradient Boosting. The main contribution is the generative representation, which uses price arrivals and volume\char45 increment marks and is implemented with conditional Denoising Diffusion Probabilistic Models.

Using French EPEX data from 2021 to 2024, we evaluate both approaches through curve reconstruction and a price\char45 maker storage optimization problem. The parametric implementation provides a deterministic reference, while the diffusion\char45 based implementation produces distributions of plausible curves and achieves higher realized profits and smaller gaps to an oracle benchmark in the storage application.
\end{abstract}

\end{frontmatter}

\tableofcontents

\section{Introduction}
\begin{figure}[H]
    \centering
    \includegraphics[width=0.6\textwidth]{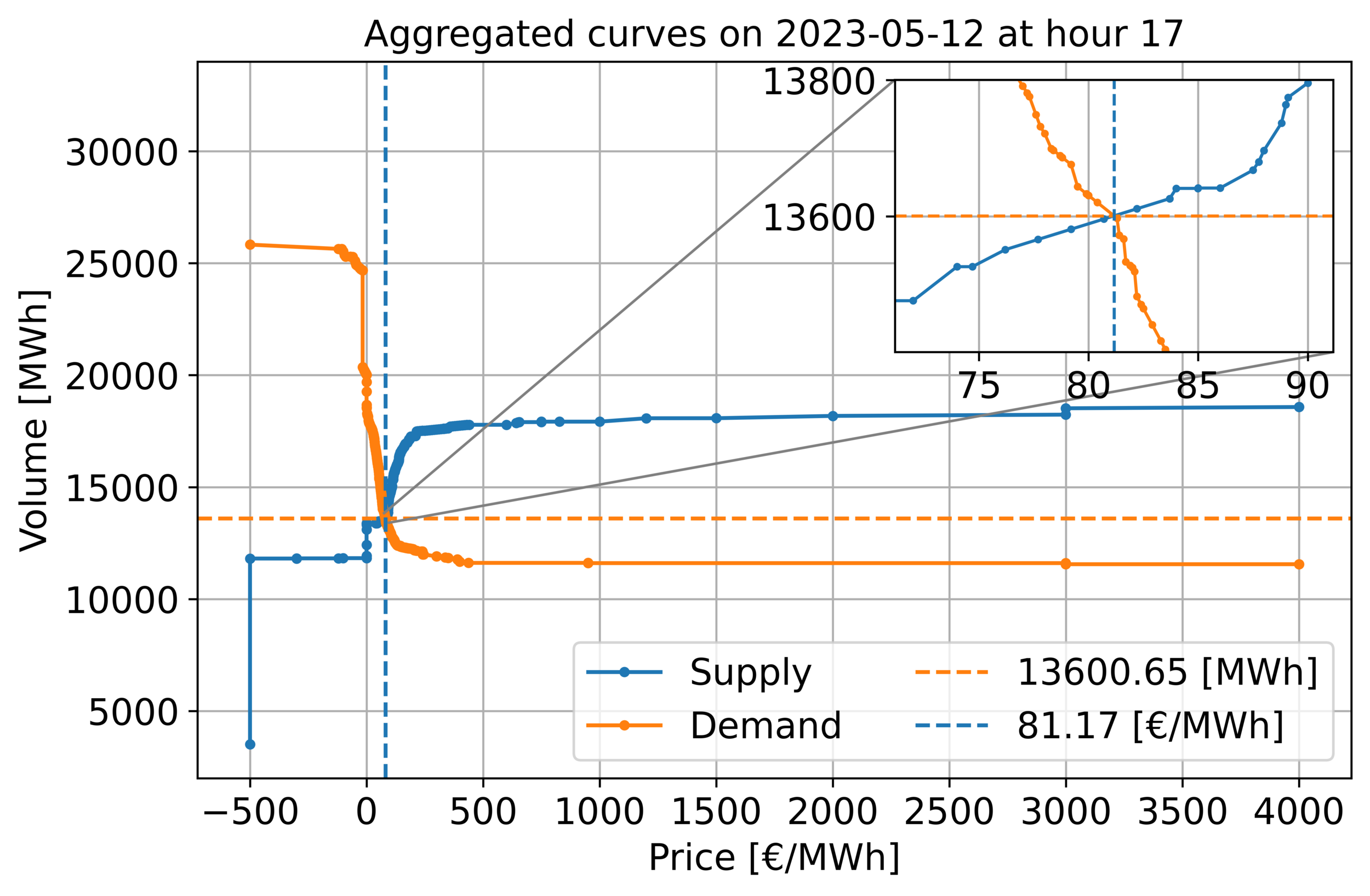}
    \caption{Aggregated bids and offers on 2023\char45 05\char45 12 at 17:00. The intersection between the demand and supply curves in the elastic region corresponds to the day\char45 ahead electricity price, 81.17 \EUR/MWh, and the traded volume, 13\, 600.65 MWh.}
    \label{fig:ex_merit}
\end{figure}

This paper studies the modelling of aggregated supply and demand curves in the EPEX SPOT Day\char45 Ahead market. In the Single Day\char45 Ahead Coupling (SDAC), market participants submit buy and sell orders that are cleared by the EUPHEMIA algorithm \citep{euphemia2023}, which maximizes social welfare subject to network and interconnection constraints. Within each bidding zone, a uniform hourly price is determined, although prices may differ across zones when constraints bind. The resulting market\char45 clearing price can be represented as the intersection of effective supply and demand curves constructed from accepted orders. Figure \ref{fig:ex_merit} illustrates this representation. Section \ref{Sec:EPEX data and curve structure} describes the empirical structure of the EPEX curves used in this work.

This curve\char45 based representation is important because prices alone do not describe how the market would react to additional volume. Full supply and demand curves contain information on price sensitivity, volume sensitivity, and price impact. This is particularly relevant for agents whose decisions are large enough to affect prices. For example, in a price\char45 taker storage problem, arbitrage consists of charging during low\char45 price hours and discharging during high\char45 price hours. In contrast, a price\char45 maker storage operator must account for the fact that large charge or discharge decisions shift the market equilibrium. This requires anticipating the shape of supply and demand curves across several delivery hours, not only forecasting their clearing prices. In Section \ref{Sec:Application Optimal strategy of a price maker storage agent}, we use this price\char45 maker storage formulation to evaluate the methodologies we propose.

Existing curve\char45 forecasting approaches include parametric representations \citep{ciarreta2023forecasting,soloviova2020efficient,sinha2025demand}. These methods reduce each curve to a finite set of parameters and typically produce a single predicted curve for given inputs. In this paper, we first develop a low\char45 dimensional parametric representation based on plateau levels, elastic\char45 region boundaries, and polynomial coefficients. This encoding is introduced in Section \ref{Sec:Parametric encoding}. For its implementation, we use eXtreme Gradient Boosting to forecast the curve parameters, as described in Section \ref{Sec:eXtreme Gradient Boosting (XGBoost)}. This parametric model provides a deterministic reference and highlights the type of information captured by point forecasts of curves.

The main contribution of the paper is a generative methodology for modelling the conditional distribution of supply and demand curves. Instead of reducing each curve to a small number of parameters, we represent curves at the order level through price arrivals and volume\char45 increment marks. This high\char45 dimensional representation is introduced in Section \ref{Sec:Generative encoding}. It interprets the order structure as a marked point process and models a block of eight delivery hours jointly, capturing cross\char45 hour dependence; the same construction can be extended directly to the full 24\char45 hour cycle. For its implementation, we use conditional Denoising Diffusion Probabilistic Models (DDPMs), presented in Section \ref{Sec:Denoising Diffusion Probabilistic Models (DDPMs)}. Generation proceeds in two stages: first, price levels are sampled from a learned intensity; second, conditional on these prices and on exogenous variables such as weather and fuel prices, the corresponding volume increments are generated simultaneously for the block of hours.

The parametric model has two roles. Methodologically, it provides a compact and interpretable encoding of full supply and demand curves. Empirically, it also serves as a deterministic reference against which the distributional nature of the generative approach can be interpreted. The two approaches have different outputs and objectives; therefore, the comparison should not be read as a direct model\char45 to\char45 model benchmark. The empirical implementation and curve\char45 reconstruction results for both approaches are reported in Section \ref{Sec:Implementation and results}. In Section \ref{Sec:Implied price computation}, we also compute the market\char45 clearing prices implied by the intersections of predicted supply and demand curves. This provides a complementary diagnostic, but price forecasting is not the primary objective of either methodology, nor the main focus of this paper.

The proposed framework is finally evaluated from an application perspective through a price\char45 maker storage optimization problem in Section \ref{Sec:Application Optimal strategy of a price maker storage agent}. The storage application is motivated by the increasing relevance of spot\char45 market arbitrage for larger, longer\char45 duration batteries. In this setting, price\char45 maker formulations account for the feedback between storage actions and market prices, including spread\char45 compression effects documented in the literature \citep{dumitrescu2024price}. The storage application therefore illustrates the value of distributional curve forecasts for decisions under price impact.

This paper addresses the lack of a order\char45 level probabilistic forecasting framework for day\char45 ahead supply and demand curves that preserves bid\char45 level structure and can be evaluated directly in a price\char45 impact decision problem. It makes three main contributions.

First, we introduce a compact parametric encoding of aggregated day\char45 ahead supply and demand curves. Building on existing curve\char45 parametrization ideas, the proposed representation combines plateau levels, elastic\char45 region boundary prices, and Chebyshev coefficients for the elastic segment. This yields an interpretable representation of each curve and a deterministic model for full\char45 curve forecasting.

Second, we introduce an order\char45 level generative representation of day\char45 ahead curves based on price arrivals and multi\char45 hour volume\char45 increment marks, implemented with conditional Denoising Diffusion Probabilistic Models. This approach produces conditional distributions of plausible supply and demand curves rather than a single deterministic forecast.

Third, we evaluate both approaches through a price\char45 maker storage optimization problem, using storage profits and gaps to an oracle benchmark as decision\char45 based measures of forecast quality under price impact. This provides an application\char45 driven evaluation of curve forecasts beyond curve\char45 reconstruction and implied\char45 price errors. Empirically, the DDPM\char45 based strategy achieves higher out\char45 of\char45 sample storage profits and smaller oracle gaps, suggesting that accurately representing the elastic region can be more valuable for price\char45 maker decisions than minimizing global curve\char45 reconstruction error.

The paper is organized as follows. Section \ref{Sec:EPEX data and curve structure} describes the empirical structure of EPEX curves and highlights the main components of curve encoding. Section \ref{Sec:Curve encoding} introduces the two curve encodings. Section \ref{Sec:Learning methods} describes the learning tools used to implement them. Section \ref{Sec:Implementation and results} presents and statistically evaluates the resulting curve forecasts. Section \ref{Sec:Application Optimal strategy of a price maker storage agent} then evaluates the forecasts in the price\char45 maker storage problem, where local curve geometry around the clearing volume determines realized decision performance. Section \ref{Sec:Implied price computation} uses implied clearing prices as a complementary diagnostic. Section \ref{Sec:ConclusionsAndFuture} presents the conclusions and discusses future research directions.

\paragraph{\bf Literature review}

\begin{table}[htbp]
\centering
{\footnotesize
\begin{tabular}{p{1.8cm} p{2.7cm} p{0.6cm} p{0.5cm} p{4.2cm} p{3.2cm}}
\hline
\textbf{Reference} & \textbf{Forecast type} & 
\textbf{Order\char45 level} & \textbf{24h joint} & 
\textbf{Market \& Calib. \& Test} & \textbf{Performance criteria} \\
\hline
\cite{aneiros2013functional} & 
Semi\char45 functional Partial Linear model for curves  &
No & No &
Spain. \& 2008\char45 2009 \& 2009 & $L^1$ error between curves
\\
\cite{canale2016constrained} & FAR model for curves (Gas) &
No & NA &
Italy \& 12\char45 2011 to 12\char45 2012 \& in\char45 sample & RMSE and MAE between curves
\\
\cite{shah2020forecasting} & FAR model for curves &
No & No &
Italy \& 2014 \& 01\char45 01\char45 2015 to 30\char45 04\char45 2015  & MAE, MAPE, RMSE between curves
\\
\cite{ghelasi2024hierarchical} & Hierarchical TS for curves &
No & No &
Germany \& 2017\char45 2018 \& 01\char45 01\char45 2019 to 30\char45 06\char45 2019 & MAE, RMSE
\\
\cite{ciarreta2023forecasting} & Linear and logistic TS for curves & 
No & No & 
Spain and Portugal \& 2015\char45 2016 \& 2017\char45 2017 & MAE, RMSE, DM test, and Theil's U statistic
\\
\cite{soloviova2020efficient} & Radial Basis Function interpolation for curves & 
No & No &
Italy \& 2017 \& in\char45 sample & CL
\\
\cite{sinha2025demand} & ML and TS for curves &
No & No &
Italy \& 01\char45 01\char45 2018 to 19\char45 01\char45 2019 \& 05\char45 07\char45 2019 to 31\char45 12\char45 2019 & Mean absolute error between curves
\\
\cite{koechlin2026dayaheadelectricitypriceforecasting} & PCA and TS for curves & 
No & No & 
Italy \& 2023\char45 2024 \& 01\char45 07\char45 2024 to 31\char45 12\char45 2024 & Average squared correlation between curves and MAE, RMSE over CL
\\
\cite{yildirim2022supply} & HMM model for distribution of curves & 
No & No & 
Turkey \& 01\char45 01\char45 2020 to 19\char45 05\char45 2020 \& 20\char45 05\char45 2020 to 09\char45 06\char45 2020 & Mean absolute percentage error between curves
\\
\cite{foronda2023prediction} & Histogram\char45 Based Gradient Boosting for curves & 
No & Yes & 
Spain \& 2014\char45 2018 \& 2019\char45 2021 & MAE, RMSE
\\
\cite{li2025predicting} & Functional regression, PCA and Random Forest for curves & 
No & Yes & 
Spain \& 2016\char45 2020 \& 2020 & Absolute error of curves and CL
\\
\cite{Guo2021} & ML with PCA for curves &
No & Yes & 
United States \& 80\% data from 01\char45 01\char45 2015 to 30\char45 06\char45 2019 \& 10 \% & MSE, MAPE
\\
\cite{kulakov2020x} &  Econometric TS model for distribution of curves &
Partial & Yes & 
Germany \& 01\char45 01\char45 2016 to 01\char45 01\char45 2017 \& 2017 & MAE, RMSE, DM test
\\
\cite{mitridati2017bayesian} & HMM model for distribution of curves &
Yes & Yes &
NA (Simulated data) & Prediction intervals and curve forecasts
\\
\cite{ziel2016electricity} & Econometric TS model for distribution of curves &
Yes & Yes &
Germany and Austria \& 01\char45 10\char45 2012 to 19\char45 04\char45 2015 \& in\char45 sample & Prediction intervals, and CL
\\
\hline 
\end{tabular}
}
\caption{Summary of representative works forecasting electricity supply and/or demand curves, including forecast type, level of order representation, joint modelling across 24 hours, markets, sample periods, and performance criteria. RMSE: Root mean squared error. MAE: Mean absolute error. MAPE: Mean absolute percentage error. DM: Diebold\char45 Mariano. NA: Not applicable. FAR: Functional autoregressive. TS: time\char45 series. ML: machine learning. HMM: Hidden Markov model. PCA: Principal component analysis. CL: clearing price.}
\label{tab:literature}
\end{table}

Forecasting supply and demand curves serves at least two purposes. First, it allows for the determination of the market\char45 clearing price as the intersection of these curves. Second, it provides information on the sensitivity of market participants beyond the realized clearing price, as this information is inherently encoded in the full shape of the supply and demand curves.

To clarify the modelling choices available for forecasting electricity markets, we distinguish between two standard approaches based on how supply and demand curves are represented and calibrated.

A first approach assumes an explicit representation of the supply and demand curves and focuses exclusively on their intersection to model the clearing price. In this setting, the curves are specified through parametric forms, and their algebraic intersection yields a parametric representation of the price, which is then calibrated against observed market prices. In this line, \cite{barlow2002diffusion} is among the early contributions, using stochastic models, while \cite{buzoianu2005dynamic} proposes a nonlinear state\char45 space framework with latent, time\char45 varying curves and particle filtering for one\char45 step\char45 ahead predictions of prices and traded volumes. We do not pursue this direction further, as our objective is not to construct reduced\char45 form models for prices based solely on curve intersections.

A second approach consists of modelling the supply and demand curves directly, with calibration performed at the curve level. Although the clearing price may still be recovered ex\char45 post as their intersection, the primary output of this approach is a model for the curves themselves, fitted to curve data rather than to prices alone. This is the perspective adopted in this work. The literature in this direction focuses on constructing flexible representations of the curves and on capturing their temporal dynamics and uncertainty; we review the most relevant contributions below and summarize them in Table \ref{tab:literature}.

\cite{aneiros2013functional} models residual demand as a functional time series, comparing a functional non\char45 parametric autoregression with a semi\char45 functional partial linear model using wind and load forecasts. \cite{canale2016constrained} forecast gas offer and demand curves using a constrained (penalized) functional autoregression.

A significant difficulty is that curves are not defined on a common price\char45 volume grid across hours, since each hourly curve is supported on its own set of price\char45 volume points. Consequently, curves from different hours are not directly comparable. One common approach is to impose a fixed grid shared across all hours, thereby providing a common support on which hourly curves can be represented and analyzed jointly. The influential paper \cite{ziel2016electricity} adopts this approach within the X\char45 model, an econometric framework that aggregates bids by price class. \cite{shah2020forecasting} and subsequent extensions of the X\char45 model by \cite{kulakov2020x} incorporate exogenous variables such as temperature and wind. \cite{haben2021probabilistic} tests different models in the Great Britain market. 

Some papers focus on supply prediction while assuming demand is inelastic (e.g., \cite{mitridati2017bayesian,kulakov2020x}). In most cases, the goal is price forecasting instead of an accurate representation of the entire curve. \cite{ghelasi2024hierarchical} extends the X\char45 Model by combining class\char45 wise LASSO regressions with hierarchical reconciliation to ensure consistency between marginal and cumulative curves, improving curve accuracy while remaining an econometric approach. More recently, \cite{koechlin2026dayaheadelectricitypriceforecasting} represents curves as functional objects, reducing their dimensionality using functional principal component analysis, and models their temporal dynamics with regularized autoregressive and vector autoregressive models with exogenous variables. 

Beyond econometric methods, machine learning has also been applied to curve forecasting. For instance, \cite{Guo2021} uses an LSTM to forecast aggregated supply curves. However, they forecast only the aggregated supply curve, using an approach that requires extensive preprocessing. \cite{yildirim2022supply} formulates the evolution of curves as a hidden Markov model, where latent variables govern how external factors such as weather and fuel prices affect each resource’s supply. \cite{foronda2023prediction} focuses on supply curves and proposes a two\char45 step curve\char45 based forecasting approach that maps all curves onto a common fixed\char45 price grid, casting the problem as multivariate prediction. Using lagged curves, calendar effects, market variables, fuel prices, and weather indicators, the full supply curve is forecast using machine\char45 learning methods, with Histogram\char45 Based Gradient Boosting yielding the best performance. 

Another approach is to change the representation of curves. For instance, \cite{li2025predicting} forecasts both curves functionally, using about $50$ coefficients per curve on a fixed price grid and a post\char45 processing step that replaces each prediction by its closest historical curve to restore monotonicity. Several methods represent each curve with a finite set of parameters. In \cite{ciarreta2023forecasting}, the elastic part is approximated with logistic or linear functions. In \cite{soloviova2020efficient}, curves are represented via integrals of radial basis functions. In the recent paper \cite{sinha2025demand}, curves are split into three zones defined by fixed prices, with the central zone capturing elasticity.


We focus on models that capture the inherent randomness of the curves and enable the recovery of their distribution, rather than producing a single point prediction. In this line, the DDPMs provide a powerful alternative, with strong performance in probabilistic energy time\char45 series (load, photovoltaic (PV), wind) and scenario generation. For example, \cite{capel2023denoising} generates full daily load, PV, and wind series conditioned on weather forecasts, outperforming Generative Adversarial Networks (GANs), Variational Autoencoders (VAEs), and Normalizing Flows (NFs) when evaluated with both statistical scores and market\char45 based measures such as day\char45 ahead bidding and storage scheduling. \cite{li2024diffcharge} adapts diffusion models with self\char45 attention to generate electric vehicle (EV) charging demand, validated through statistical distances and bidding simulations. \cite{lin2024energydiff} demonstrates scalability to high\char45 resolution time series (up to minute\char45 level) and introduces marginal calibration via optimal transport. \cite{zhang2024generating} proposes a physics\char45 informed diffusion model integrating PV system performance for net load generation. 
These works highlight the potential of DDPM\char45 based tools for realistic scenario generation in electricity markets, especially for supply\char45 demand modelling. This justifies our choice of DDPMs as the core modelling framework. To the best of our knowledge, this is the first work to model day\char45 ahead auction supply and demand curves at the order level using DDPMs. 


Most studies on storage in the spot market assume a price\char45 taker setting (\cite{mercier2023value,aazami_optimal_2022,collet2017optimal,dumitrescu2024price}); see \cite{machlev_review_2020} for reviews and \cite{biggins2022going} for a comparison between price\char45 taker and price\char45 maker strategies. We instead consider the less common case of a price\char45 maker storage agent in the day\char45 ahead market. Similar analyses appear in \cite{ding2017optimal,shafiee2016economic,barbry2019robust}. In these works, supply and demand curves are typically assumed known or reconstructed from historical data, without being embedded in an explicit modelling framework. 

For our illustrative analysis, we build on Supply Function Equilibrium (SFE) models (\cite{klemperer_supply_1989,anderson_existence_2013,baldick_theory_2004,holmberg_supply_nodate,holmberg_unique_2008}). In this setting, the storage agent optimizes block orders across hours based on price signals.


\section{EPEX data and curve structure}
\label{Sec:EPEX data and curve structure}

\begin{figure}[H]
    \centering
    \includegraphics[width=\textwidth]{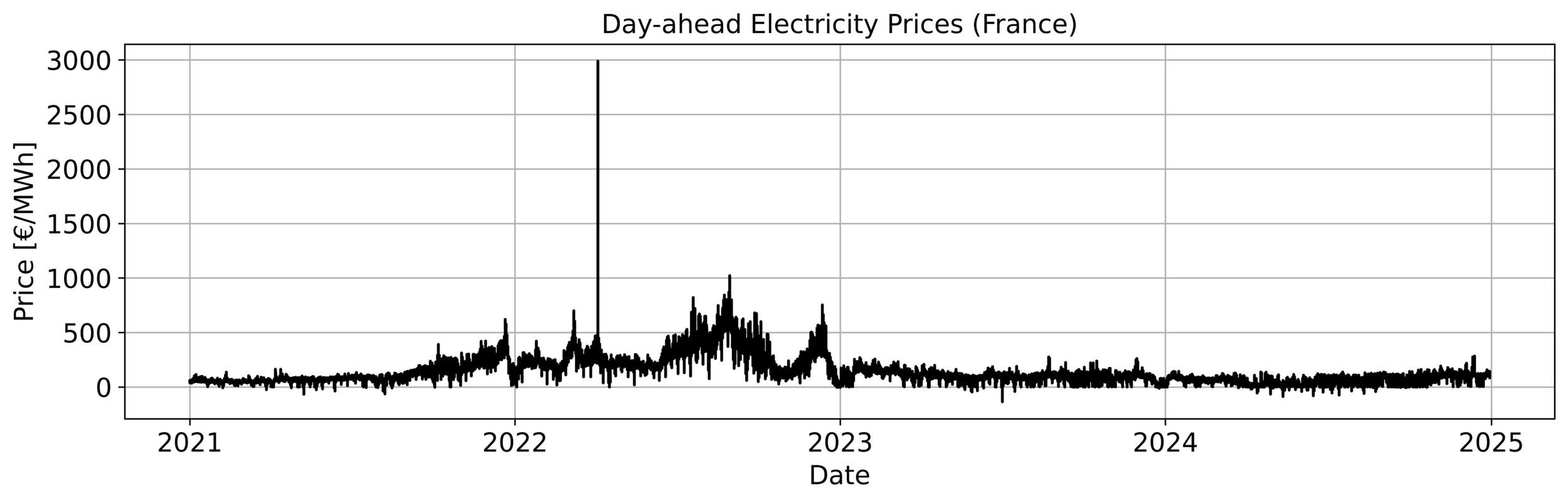}
    \caption{Evolution of the day\char45 ahead electricity price over time between 2021 and 2024}
    \label{fig:Prices}
\end{figure}

We use aggregated EPEX SPOT demand and supply curves for the French Day\char45 Ahead market, covering the period from 2021\char45 01\char45 01 to 2024\char45 12\char45 31. Figure \ref{fig:Prices} reports that extremely high prices are rare and concentrated in the upper tail. Thus, we restrict each curve to the price range from $-300$ \EUR/MWh to $3\,000$ \EUR/MWh. This range was chosen to contain every spot price during this period\footnote{In practice, $3\,000$ \EUR/MWh served as the upper bound for day\char45 ahead prices until 2022\char45 04\char45 04, after France recorded its highest peak price and the limit was raised to $4\,000$ \EUR/MWh (\cite{cre2022})}. For clarity of exposition, we restrict the dataset and curve forecast modelling to eight representative delivery hours:
\begin{align}
\label{eq:Selected hours}
H:=\{5,\quad 6,\quad 7,\quad 8,\quad 18,\quad 19,\quad 20,\quad 21\}.
\end{align}
These hours cover the morning and evening ramp periods, when demand changes rapidly, and price formation is more sensitive to variations in the demand and supply curves. The same construction can be extended directly to all 24 delivery hours; we use this subset to simplify the analysis and presentation.

For each delivery hour, supply and demand curves are defined on distinct price\char45 volume grids but exhibit a similar structure: two inelastic regimes separated by an elastic region, where the market price is determined. As Figure \ref{fig:ex_merit} shows, the demand curve starts with a plateau representing the maximum quantity demanded, followed by a decreasing elastic segment in which demand adjusts to price through arbitrage and demand\char45 response mechanisms. It ends with a second plateau, corresponding to the minimum quantity demanded regardless of price.

As shown in Figure \ref{fig:ex_merit}, each hourly aggregated curve is represented by an ordered set of price\char45 volume pairs, with prices increasing along the curve. The monotonicity of the associated volumes depends on the market side: supply volumes are non\char45 decreasing in price, whereas demand volumes are non\char45 increasing in price. Each curve consists of two plateau regions separated by an elastic segment, whose position and shape are described by its boundary prices and local slopes. Curves are observed as ordered price\char45 volume pairs, rather than on a fixed common grid, so both the number and the location of price points vary across hours. The volume changes between consecutive prices identify the additional supply, or demand, associated with each price level. When several hours are considered jointly, a given price level may be associated with volume changes in multiple hours, revealing cross\char45 hour structure in the block of curves. These structural features will be used in the following sections to motivate the proposed encoding approaches.

Having described the empirical plateau\char45 elastic\char45 plateau structure of the curves, we now introduce two representations designed to capture this structure at different levels of detail. A parametric encoding compresses each curve into a small number of interpretable quantities, while a generative encoding retains order\char45 level information and produces distributions over possible curves.

\section{Curve encoding}
\label{Sec:Curve encoding}

We now introduce two complementary encodings of the EPEX supply and demand curves, based on the structural features described in Section \ref{Sec:EPEX data and curve structure}. The first encoding is parametric: it exploits the plateau\char45 elastic\char45 plateau shape of aggregated curves to summarize each curve by a small number of interpretable parameters. This encoding is applied separately to each delivery hour and produces deterministic curve forecasts. The second encoding is generative: it represents curves at the order level through price arrivals and volume increments, and simultaneously encodes the hours in the block $H$ defined in \eqref{eq:Selected hours}. This representation allows the model to generate multiple curve scenarios conditional on the observed market covariates.

\subsection{Parametric encoding}
\label{Sec:Parametric encoding}

In this section, we consider parametric curve\char45 representation models that produce a single forecasted curve. We propose a parametric representation inspired by the representative works of \cite{ciarreta2023forecasting,soloviova2020efficient,sinha2025demand}. We provide a detailed comparison with these works at the end of this section. The approach is presented for demand curves, but it extends directly to supply curves. Results for both demand and supply are reported in Section \ref{Sec:Implementation and results}. Figure \ref{fig:diagram_Parametric} summarizes the proposed encoding pipeline, whose individual steps are detailed in the remainder of this subsection.

\begin{figure}[H]
	\centering
    \resizebox{1\textwidth}{!}{%
\begin{tikzpicture}[
	box/.style={draw, rectangle, rounded corners, align=center,
		minimum width=0.5cm, minimum height=1.2cm,
		inner sep=2pt, outer sep=0pt},
	inputBox/.style={draw, rectangle, dashed, inner sep=2pt},
	arrow/.style={-{Latex[length=3mm]}, thick},
	node distance=4mm and 8mm
	]
	
	\node[box] (observation) at (0, 0.75){
		Data\\
		$(P^h_n, D^h_n)_{n=0}^{N^h}$
	};

	\node[box, right=0.5cm of observation] (1Augment) {
		1. Curve\\ 
        augmentation\\
		$(P^h_i, D^h_i)_{i=1}^{M'}$\\ $M'\geq 200$.
	};

    \draw[arrow] (observation.east) -- (1Augment.west);

    \node[box, right=0.8cm of 1Augment, yshift=0.7cm] (2PlateauU) {
		2. First plateau\\
        detection \\
        $(p_{\mathrm{start}},U)$
	};

    \node[box, right=0.6cm of 1Augment, yshift=-0.7cm] (3PlateauL) {
		3. Second plateau\\
        detection \\
        $(p_{\mathrm{end}},L)$
	};

    \node[inputBox, fit=(2PlateauU) (3PlateauL)] (23) {};

    \draw[arrow] (1Augment.east) -- (23.west);

    \node[box, right=0.5cm of 3PlateauL, yshift=0.7 cm] (4PolyApprox) {
		4. Polynomial\\
        approx. of \\
        elastic region\\ $c_0,\ldots,c_3$.
	};

    \draw[arrow] (23.east) -- (4PolyApprox.west);

    \node[box, right=0.5cm of 4PolyApprox] (5MonoEnforce) {
		5. Monotonicity\\ enforcement
	};

    \draw[arrow] (4PolyApprox.east) -- (5MonoEnforce.west);

    \node[box, right=0.5cm of 5MonoEnforce] (finalencoding) {
		Parametric\\ encoding\\
        $\left(p_{\mathrm{start}},U,p_{\mathrm{end}},\right.$\\
        $\left.L,c_0,\ldots,c_3\right)$
	};

    \draw[arrow] (5MonoEnforce.east) -- (finalencoding.west);
	
\end{tikzpicture}}
	\caption{Parametric encoding}
	\label{fig:diagram_Parametric}
\end{figure}

Since the demand and supply curves are observed on different $(\text{price}, \text{volume})$ grids across delivery hours, we use their inelastic and elastic regimes to construct a reduced, interpretable parametrization. Rather than interpolating all curves on a fine common price grid, which would lead to a high\char45 dimensional representation, we reduce the forecasting problem to a small set of parameters: the values associated with the inelastic regions and a polynomial-based approximation of the elastic region.

A direct polynomial fit can be unstable for aggregated curves because the number of observed points varies across hours, and large gaps may occur between consecutive price levels with nonzero volumes. To mitigate this issue, we augment each curve with regularly spaced candidate price points over its price range. Existing points are preserved, while missing candidate points are filled by linear interpolation. This increases point density, especially in plateau regions, yielding between $200$ and $N+200$ points, where $N$ is the number of original points.

\begin{figure}[H]
  \centering
  \begin{subfigure}{0.495\linewidth}
    \centering
    \includegraphics[width=\linewidth]{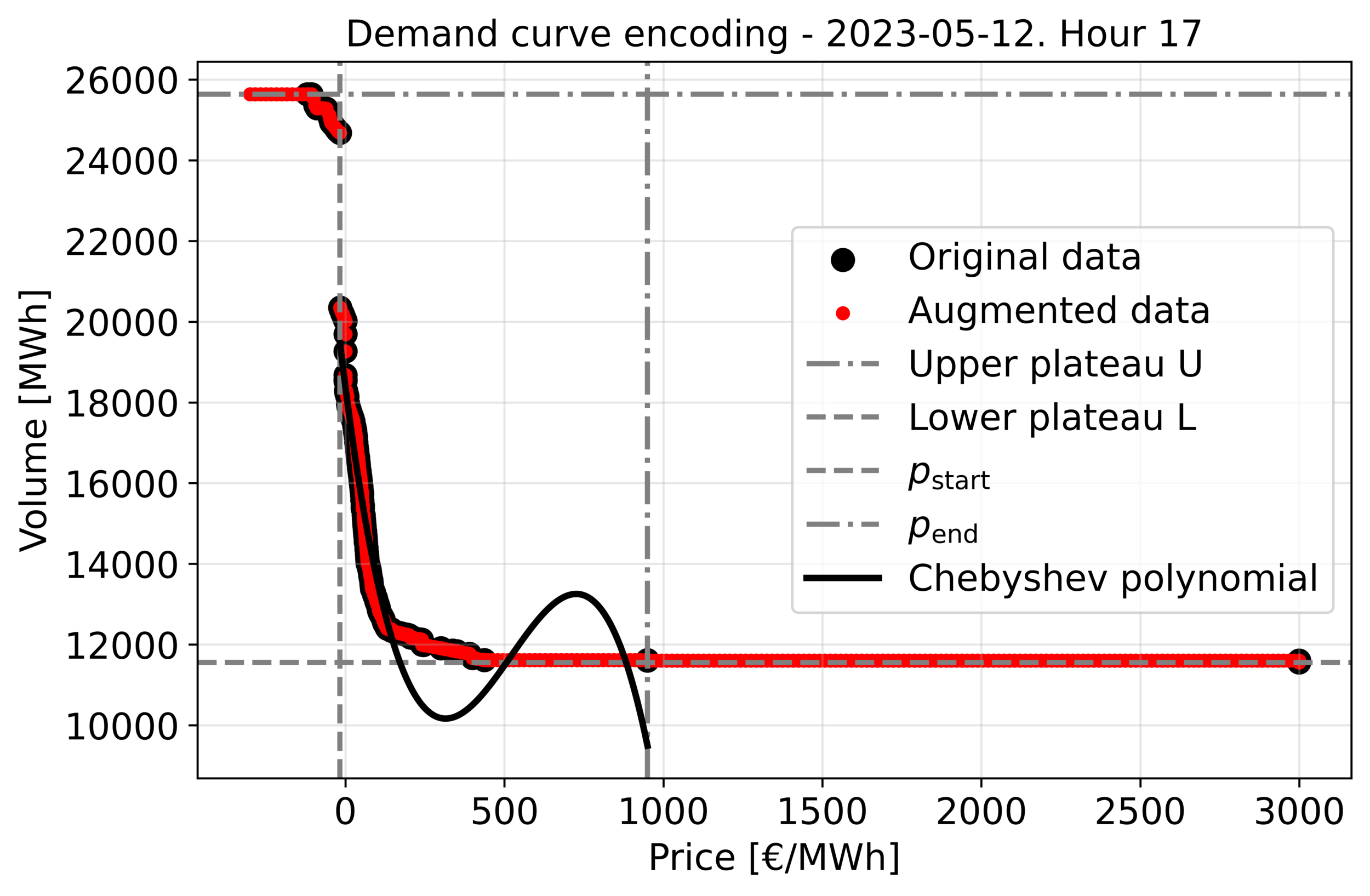}
    \caption{Steps 1 to 4}
    \label{fig:Steps 1 to 4}
  \end{subfigure}\hfill
  \begin{subfigure}{0.495\linewidth}
    \centering
    \includegraphics[width=\linewidth]{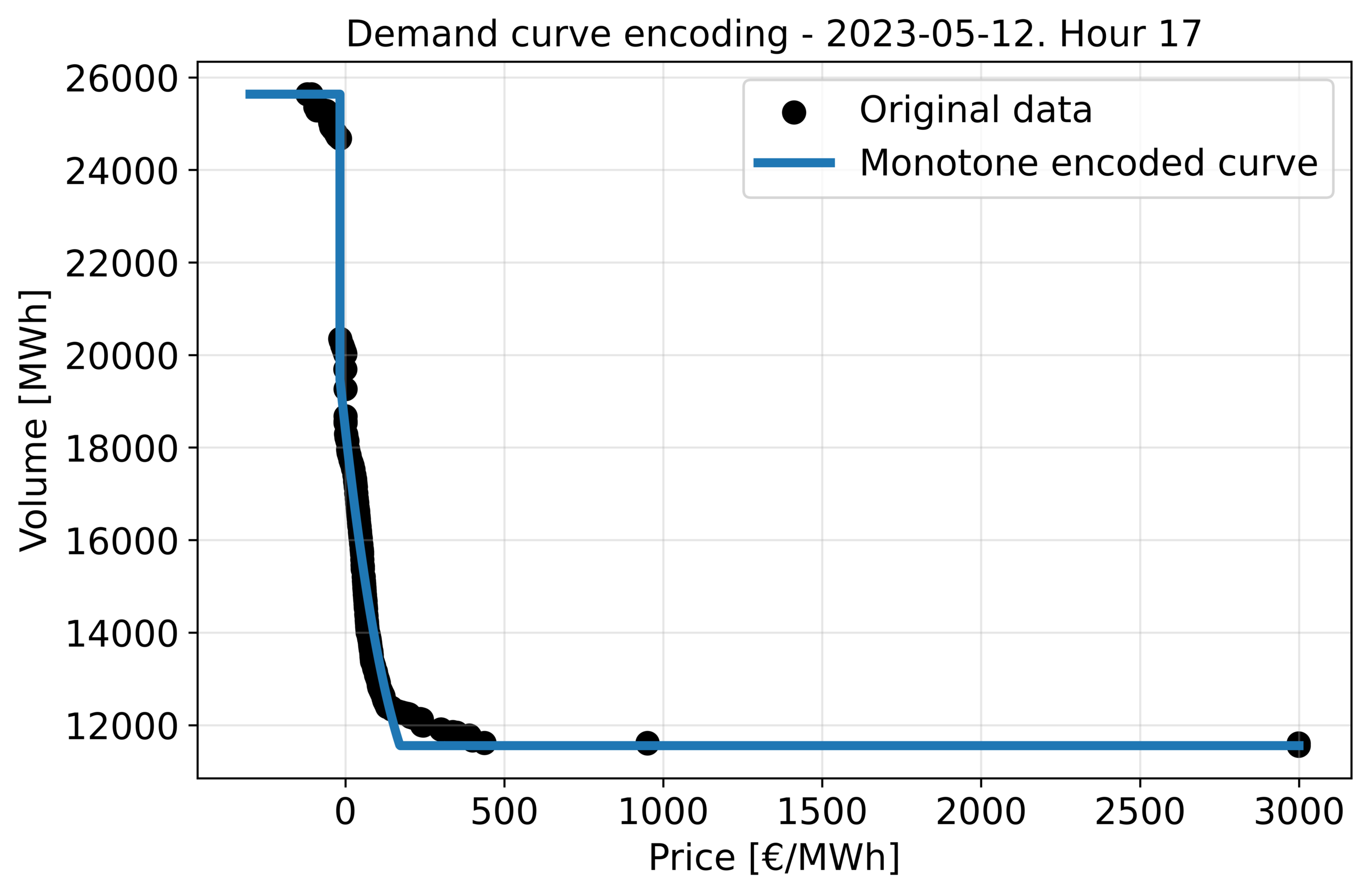}
    \caption{Step 5}
    \label{fig:Step 5}
  \end{subfigure}
  \caption{Illustration of the encoding steps applied to the aggregated demand curve for 2023\char45 05\char45 12, delivery hour 16.}
  \label{fig:approximation}
\end{figure}

The proposed approach, illustrated in Figure \ref{fig:approximation}, consists of five steps:
\begin{enumerate}

\item We augment each aggregated curve with $200$ regularly spaced candidate price points. 

\item We detect the first plateau, yielding two parameters: $U$, the demand value on this plateau, and $p_{\text{start}}$, the price at which the plateau ends. (Figure \ref{fig:Steps 1 to 4})

\item We detect the second plateau, yielding two parameters: $L$, the minimum demand required to keep the system running, and $p_{\text{end}}$, the price at which this plateau begins. (Figure \ref{fig:Steps 1 to 4})

\item We rescale prices on the elastic segment between $p_{\text{start}}$ and $p_{\text{end}}$ to $[-1,1]$ and fit a degree\char45 3 polynomial, represented in the Chebyshev basis for numerical stability, resulting in four coefficients $(c_0,\dots,c_3)$. (Figure \ref{fig:Steps 1 to 4})

\item We enforce monotonicity by replacing the fitted values with their left\char45 to\char45 right cumulative minima.\footnote{Algorithmically, this is implemented using NumPy's \texttt{accumulate} function applied to the minimum function.} For supply curves, we analogously use cumulative maxima. (Figure \ref{fig:Step 5})
\end{enumerate}

Our encoding approach reduces curve representation to eight parameters:
\begin{align}
\label{eq:Eight parameters}
p_{\text{start}},\;U,\;p_{\text{end}},\;L,\; c_0,\;c_1,\;c_2,\;c_3.
\end{align}
Alternative specifications for the elastic region, including higher\char45 degree Chebyshev polynomials of degrees 5 and 7, achieved performance similar to the degree\char45 3 specification, whereas sigmoid functions were less accurate. Monotone I\char45 splines with four basis functions and degree 3 also produced results close to those of the degree\char45 3 polynomial. Since forecasting accuracy is mainly driven by the correct prediction of the plateau regions, which are common to all specifications, we retain the degree\char45 3 Chebyshev approximation for its accuracy, simplicity, and computational efficiency.

The elastic segment is detected using Algorithm \ref{algo_elastic}. We first compute the discrete slopes between consecutive price\char45 volume points. The elastic segment is then defined as the interval between the first and last prices whose associated absolute slope exceeds the $90 \%$ quantile of all absolute slopes. This threshold was selected after empirical trials with quantiles ranging from $70\%$ to $95\%$. 

{
\renewcommand{\baselinestretch}{1.15}\selectfont
\begin{algorithm}[H]
\caption{Detection of the elastic segment}
\label{algo_elastic}
\begin{algorithmic}[1]
\REQUIRE Sorted price\char45 volume observations $(P^h_n,D^h_n)_{n=0}^{N^h}$; percentile level $pct$
\ENSURE Boundary prices $p_{\text{start}}$ and $p_{\text{end}}$ of the elastic segment
\STATE Compute the discrete slopes: $s_n \leftarrow \frac{D^h_{n+1}-D^h_n}{P^h_{n+1}-P^h_n}, 
    \qquad n=0,\dots,N^h-1$.
\STATE Compute the slope threshold: $\text{thr} \leftarrow \operatorname{percentile}(|s_0|,\dots,|s_{N^h-1}|;\,pct)$.

\STATE Identify the intervals with sufficiently large slope: $I \leftarrow \{\, n \in \{0,\dots,N^h-1\} : |s_n| > \text{thr} \,\}$.

\IF{$I=\emptyset$}
    \RETURN $p_{\text{start}} \leftarrow P^h_0$, \quad $p_{\text{end}} \leftarrow P^h_{N^h}$
\ELSE
    \STATE Let $n_{\min} \leftarrow \min I$ and $n_{\max} \leftarrow \max I$
    \RETURN $p_{\text{start}} \leftarrow P^h_{n_{\min}}$, \quad $p_{\text{end}} \leftarrow P^h_{n_{\max}+1}$
\ENDIF
\end{algorithmic}
\end{algorithm}
}

On the elastic segment, we fit a Chebyshev polynomial and enforce monotonicity.\footnote{Computed with NumPy's \texttt{Chebyshev} function; monotonicity is enforced using \texttt{np.minimum.accumulate}.} Although this post\char45 processing does not guarantee continuity at $p_{\text{start}}$ and $p_{\text{end}}$, the resulting kinks are negligible at the scale of the full curve.

To measure the approximation error introduced by the proposed encoding, we compare the reconstructed curves with the original data points. For a curve with $N$ observations, we compute the Mean Absolute Error (MAE) and its normalized version as
\begin{equation}
\label{MAE_def}
\text{MAE}=\frac{1}{N}\sum_{n=1}^{N} |D_n-\hat{D}_n|, \qquad \text{nMAE}=\frac{\text{MAE}}{(U+L)/2},
\end{equation}
where $U$ and $L$ denote the plateau levels, $D_n$ is the observed value, and $\hat{D}_n$ the encoded approximation. Table \ref{tab:encoding_error_train} reports the resulting median errors over the full dataset.

\begin{table}[H]
\centering
\resizebox{0.8\textwidth}{!}{
\begin{tabular}{|c|c|cccccccc|}
\hline
\multirow{2}{*}{Curve}
& \multirow{2}{*}{Metric}
& \multicolumn{8}{c|}{Hour} \\
\cline{3-10}
& & 5 & 6 & 7 & 8 & 18 & 19 & 20 & 21 \\
\hline

\multirow{2}{*}{Demand}
& MAE [MWh]
& 681.47 & 644.30 & 656.21 & 807.07 & 747.05 & 678.98 & 691.35 & 811.32 \\

& nMAE (\%)
& 4.06 & 3.73 & 3.78 & 4.51 & 3.85 & 3.48 & 3.58 & 4.42 \\
\hline

\multirow{2}{*}{Supply}
& MAE [MWh]
& 773.92 & 682.65 & 695.86 & 800.23 & 919.12 & 773.59 & 726.71 & 781.55 \\

& nMAE (\%)
& 5.48 & 5.01 & 4.98 & 5.56 & 5.82 & 5.04 & 5.06 & 5.46 \\
\hline
\end{tabular}}
\caption{Median approximation error by selected delivery hour.}
\label{tab:encoding_error_train}
\end{table}

We close this section by positioning the proposed parametrization relative to three representative approaches of \cite{ciarreta2023forecasting,soloviova2020efficient,sinha2025demand} that motivate different aspects of our construction.

\cite{ciarreta2023forecasting} focuses on day\char45 ahead price forecasting rather than curve reconstruction. Their approach represents curves as price functions of quantity and uses linear or logistic parametrizations over an interior region. This is related to our parametrization, since their boundary quantities, such as $Q^d_{\min,t}$ and $Q^d_{\max,t}$, delimit the fitted region, while their coefficients \(a_{0,t},a_{1,t}\), or \(a_{0,t},\dots,a_{3,t}\), describe its shape. In contrast, we represent curves as volume functions of price and encode the full curve through plateau levels, boundary prices, and Chebyshev coefficients for the elastic segment.

Our approach is also related to the curve\char45 compression method of \cite{soloviova2020efficient}, which approximates curves by mesh\char45 free interpolation with integrated Gaussian radial basis functions. Both methods provide low\char45 dimensional functional encodings, but their parameters are basis\char45 function coefficients controlling scale, location, and steepness, whereas ours are directly tied to interpretable curve features.

The closest structural comparison is with \cite{sinha2025demand}, which also reduces dimensionality by isolating the informative part of the curve. They identify two boundary points, $A$ and $B$, forecast their coordinates, and encode the intermediate segment $C$ through a monotonic autoencoder. In our framework, $A$ and $B$ correspond to the boundary points of the elastic segment $(U,p_{\mathrm{start}})$ and $(L, p_{\mathrm{end}})$, while $C$ is represented explicitly by Chebyshev coefficients. Thus, their method is more flexible but latent, whereas ours is more restrictive but explicit and interpretable.

\subsection{Generative encoding}
\label{Sec:Generative encoding}

In this section, we develop a generative representation of supply and demand curves by modelling the order\char45 level structure from which they are built. The model learns the conditional distribution of orders, generates synthetic order\char45 level realizations, and aggregates them to reconstruct curves. It also accounts for the dependence of order prices and volumes on covariates such as weather and fuel prices. We present the approach for supply curves, but it extends directly to demand curves once the corresponding sign convention for demand\char45 side volume increments is fixed. Results for both supply and demand are reported in Section \ref{Sec:Implementation and results}. Figure \ref{fig:diagram_generative} shows the proposed encoding pipeline, with individual steps detailed in the remainder of this subsection.

\begin{figure}[H]
	\centering
    \resizebox{1\textwidth}{!}{%
\begin{tikzpicture}[
	box/.style={draw, rectangle, rounded corners, align=center,
		minimum width=0.5cm, minimum height=1.2cm,
		inner sep=2pt, outer sep=0pt},
	inputBox/.style={draw, rectangle, dashed, inner sep=2pt},
	arrow/.style={-{Latex[length=3mm]}, thick},
	node distance=4mm and 8mm
	]
	
	\node[box] (observation) at (0, 0.75){
		Data\\
		$((P^h_n, S^h_n)_{n=0}^{N^h})_{h\in H}$
	};

	\node[box, right=0.5cm of observation] (1Deltas) {
		Computation of \\
        increments \eqref{eq:Deltas supply}\\
        $((\Delta S^h_n)_{n=1}^{N^h})_{h\in H}$
	};

    \draw[arrow] (observation.east) -- (1Deltas.west);

    \node[box, right=0.5cm of 1Deltas] (CommonPrices) {
		Collection of\\
        unified prices (arrivals) \eqref{eq:common price support}\\
        $(P_m)_{m=1}^{M}$
	};

    \draw[arrow] (1Deltas.east) -- (CommonPrices.west);

    \node[box, right=0.5cm of CommonPrices] (8Volumes) {
		8\char45 dimensional \\
        increments (marks) \eqref{eq:Deltas}\\
        $(\Delta S_m)_{m=1}^{M}$
	};

    \draw[arrow] (CommonPrices.east) -- (8Volumes.west);

    \node[box, right=0.5cm of 8Volumes] (Finalencoding) {
		Generative \\
        encoding\\
        $((P_m,\Delta S_m))_{m=1}^{M}$
	};

    \draw[arrow] (8Volumes.east) -- (Finalencoding.west);

\end{tikzpicture}}
	\caption{Generative encoding}
	\label{fig:diagram_generative}
\end{figure}

We model the supply curve as a marked Cox process (see, e.g., \cite{jacobsen2006point}): arrival prices are paired with volume increments across the selected hours \eqref{eq:Selected hours}, with a stochastic price intensity and volume marks.

\begin{figure}[H]
   \centering
    \begin{subfigure}[b]{0.495\textwidth}
        \includegraphics[width=1\textwidth]{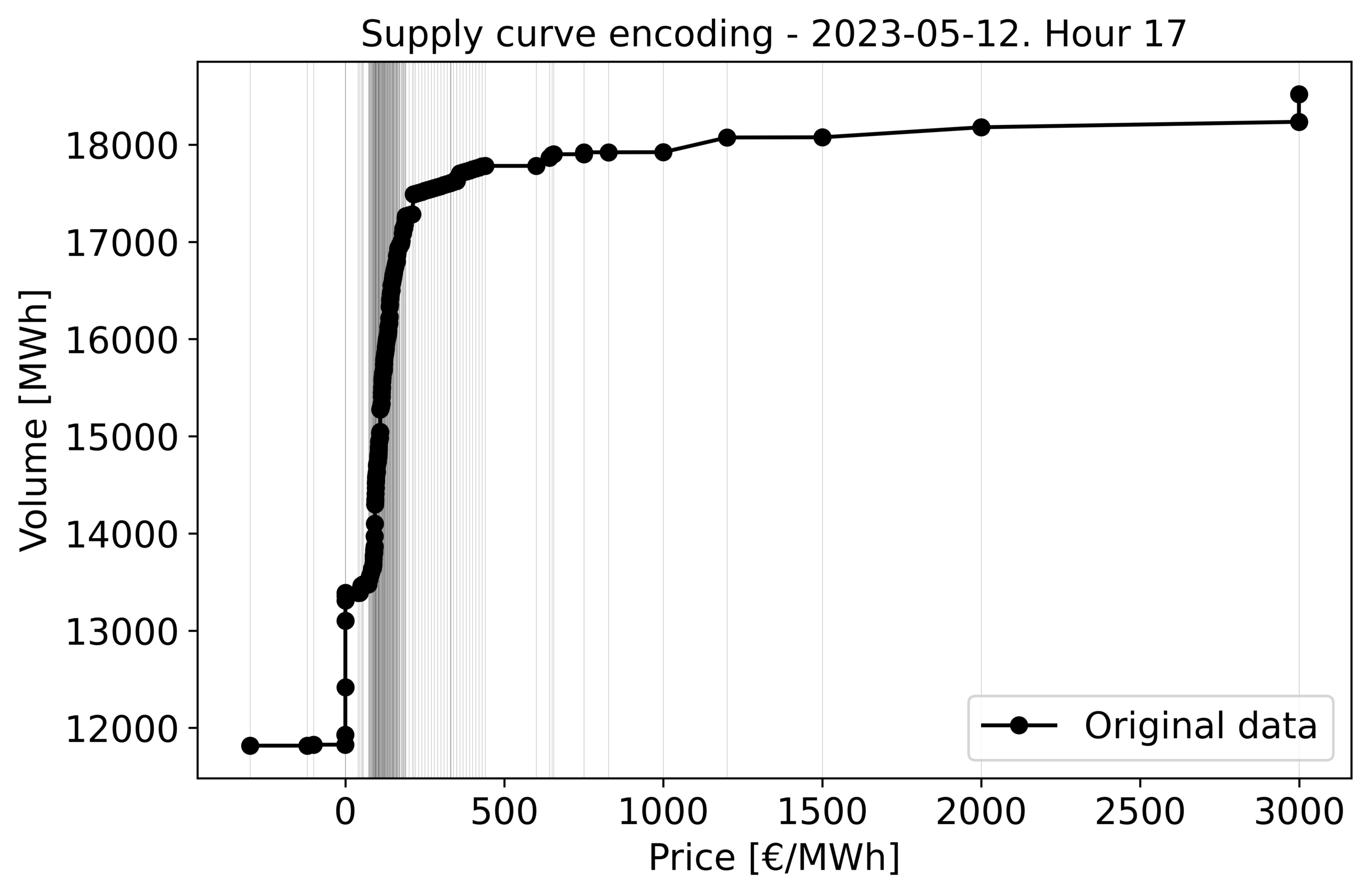}
    \caption{Supply curve}
    \label{fig:Supply curve}
    \end{subfigure}
    \hfill
    \begin{subfigure}[b]{0.495\textwidth}
        \includegraphics[width=1\textwidth]{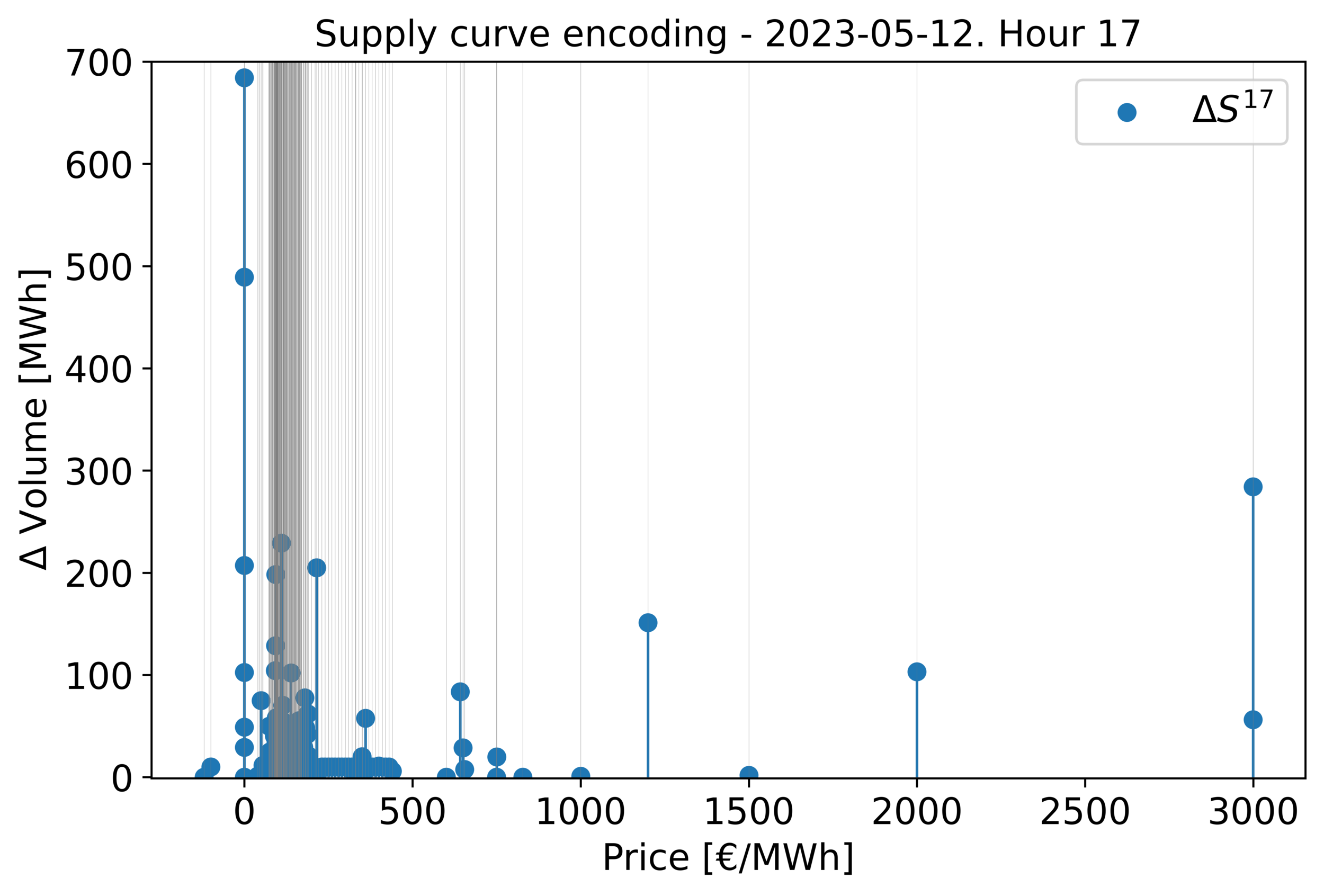}
    \caption{Marked Cox process representation}
    \label{fig:Marked Cox process representation}
    \end{subfigure}  

    \caption{Supply curve and marked point process $(P^{17}_n, \Delta S^{17}_n)_{n=1}^{206}$ for hour 17 on 2023\char45 05\char45 12}
    \label{fig:MeritOrder_Example}
\end{figure}

Fix a day, and for each selected hour $h \in H$ (see \eqref{eq:Selected hours}), let $(P^h_n, S^h_n)_{n=0}^{N^h}$ denote the price\char45 volume points defining the supply curve, as shown in Figure \ref{fig:Supply curve}. We set\footnote{If no volume is observed at these boundary prices, we impute it by interpolation.} $P^h_0=-300$ and $P^h_{N^h}=3\,000$. We compute volume increments along the supply curve as
\begin{align}
\label{eq:Deltas supply}
\Delta S^h_n = S^h_n - S^h_{n-1}, \quad n=1,\ldots,N^h.
\end{align} 
Each increment $\Delta S^h_n$ gives the additional volume offered at price $P^h_n$. We then represent the order\char45 level supply data by the marked points $(P^h_n,\Delta S^h_n)_{n=1}^{N^h}$, as shown in Figure \ref{fig:Marked Cox process representation}. 

\begin{figure}[H]
    \centering
        \includegraphics[width=1\textwidth]{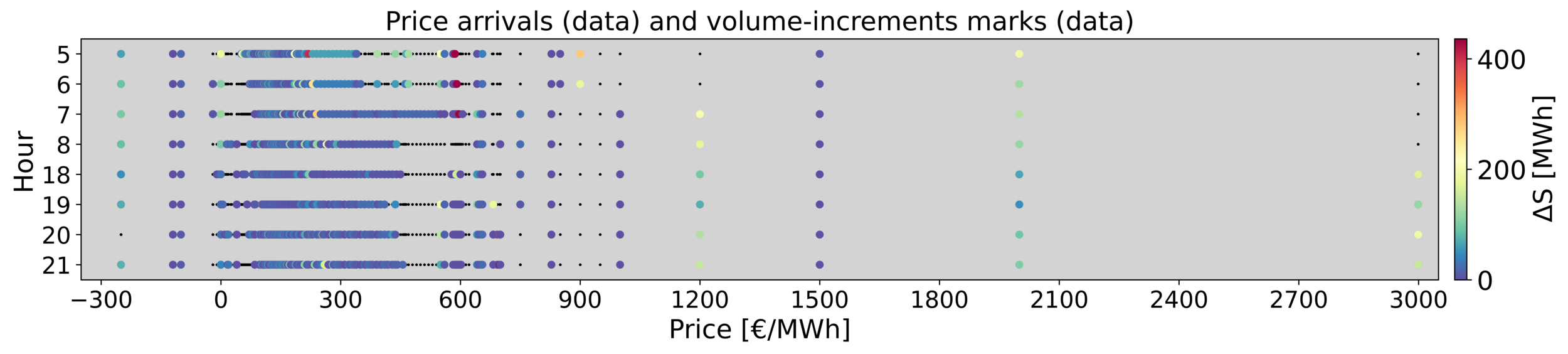}
        \caption{$(P_m, \Delta S_m)_{m=1}^{530}$ on 2023\char45 05\char45 12. Black dots represent zero entries of $\Delta S$ defined in \eqref{eq:Deltas}.}
        \label{fig:Ex_DV}
\end{figure}

Instead of fitting one model per hour, we model the block simultaneously by collecting all distinct price values across all hours 
\begin{align}
\label{eq:common price support}   
\bigcup\limits_{h \in H} \{P_n^h:\, n=1,\ldots,N^h\},
\end{align}
from which we construct an increasing ordered sequence $(P_m)_{m=1}^{M}$. This sequence provides a unified price grid at the day level and offers a common support on which all hourly curves can be aligned. For each $P_m$, we attach the $8$\char45 dimensional vector of volume increments (mark)
\begin{equation}
\label{eq:Deltas}
\Delta S_m := \left( \Delta S_m^h \right)_{h\in H}, \quad \Delta S^h_m := \begin{cases}
     \Delta S^h_n, & \text{if } P_m = P^h_n \text{ for some }n \in \{1,\ldots,N^h\},
     \\
     0, & \text{otherwise.}
\end{cases} \quad \quad h\in H
\end{equation}
Figure \ref{fig:Ex_DV} illustrates the pairs $(P_m, \Delta S_m)_{m=1}^{M}$. This representation captures complex supply offers\footnote{We do not model linked block orders or minimum income condition orders, as these impose recursive cross\char45 mark constraints beyond the present scope.} such as block orders\footnote{Consisting of all\char45 or\char45 nothing bids spanning consecutive hours at one price, yielding multiple non\char45 zero entries in $\Delta S_m$} and flexible hourly block orders\footnote{Consisting of one\char45 hour blocks placed over an admissible hour set}. 

To complete the encoding, we model $\left(P_m, \Delta S_m\right)_{m=1}^{M}$ as a realization of a marked Cox process $(\mathbf{P}, \mathbf{\Delta S})$. The price arrivals $\mathbf{P}$ follow a Cox process with stochastic intensity $\boldsymbol{\lambda}$, conditional on features $\mathbf{X}^P$. The marks $\mathbf{\Delta S}$ are $8$\char45 dimensional volume increments, modelled conditionally on the arrival $\mathbf{P}$ and features $\mathbf{X}^S$. We write
\begin{align}
\label{eq:Dependence} 
& \boldsymbol{\lambda}
\sim p^{\lambda}(\,\cdot \mid \mathbf{X}^P), \qquad 
\mathbf{\Delta S} \sim p^{S}(\,\cdot \mid \mathbf{P},\mathbf{X}^S),
\end{align}
where \(p^{\lambda}\) and \(p^{S}\) denote the conditional distributions. 

We normalize prices from $[-300,3\,000]$ \EUR/MWh to $[0,1]$ and, by abuse of notation, still denote the normalized prices by $\mathbf{P}$. The intensity $\boldsymbol{\lambda}$ is represented on a 30\char45 point grid: 5 equally spaced points on each outer interval $[0,\,0.06)$ and $(0.22,\,1]$, and 20 equally spaced points on the central interval $[0.06,\,0.22]$. These intervals correspond to the price ranges $[-300,-102]$, $[-102,426]$, and $[426,\,3\,000]$ \EUR/MWh. The grid allocation was selected after testing several configurations, balancing curve accuracy and computational cost. 

\begin{figure}[H]
    \centering
    \begin{subfigure}[b]{0.5\textwidth}      \includegraphics[width=1\textwidth]{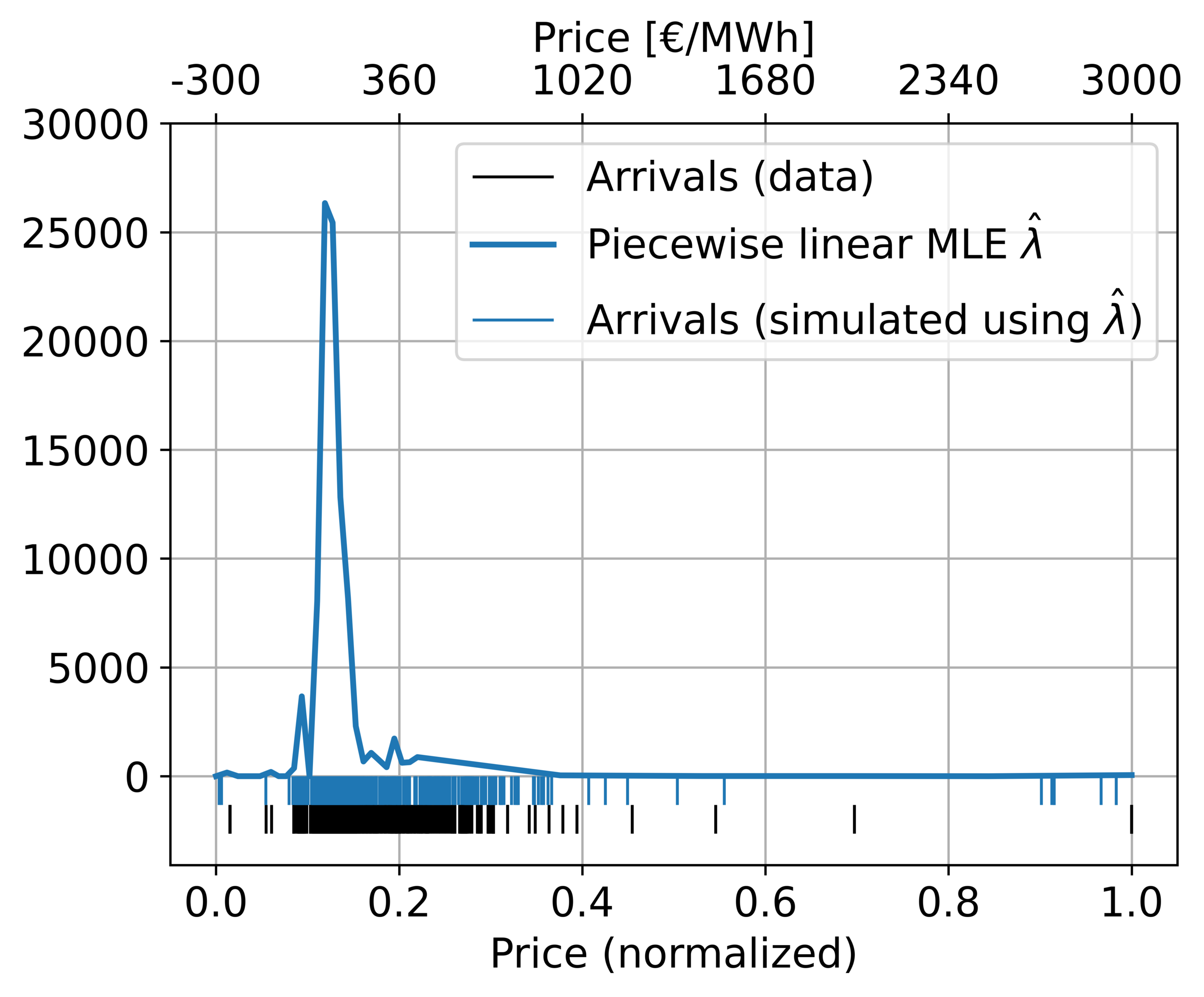}
        \caption{Arrivals (data), surrogate $\hat{\lambda}$ and generated arrivals}
        \label{fig:Ex_Intensity}
    \end{subfigure}
    \begin{subfigure}[b]{0.38\textwidth}   \includegraphics[width=1\textwidth]{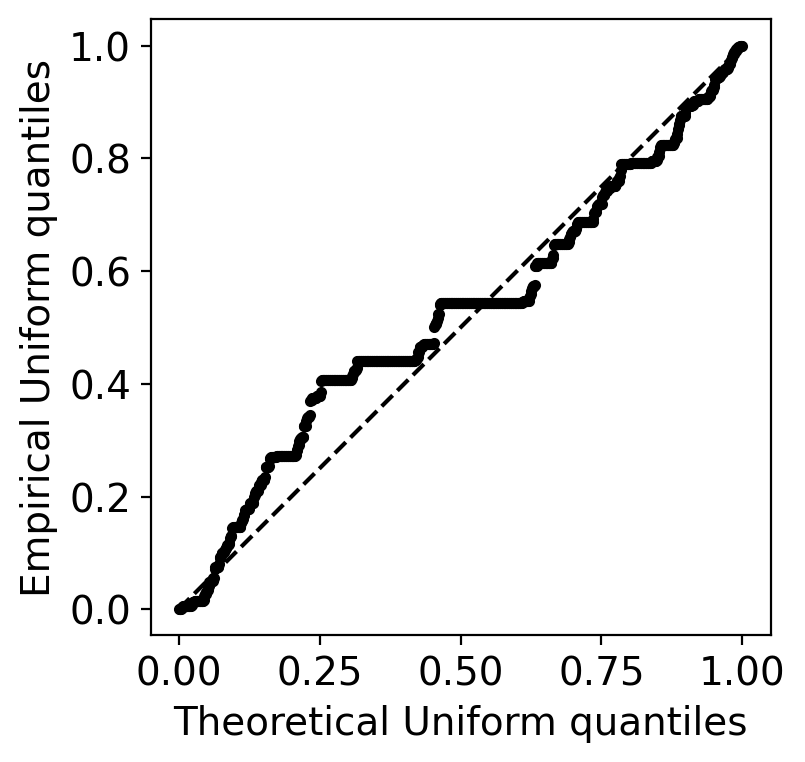}
        \caption{Q\char45 Q plot of transformed interarrival times under $\hat{\lambda}$ vs. U$[0,1]$ distribution.}
        \label{fig:Ex_Intensity_fit}
    \end{subfigure} 

    \caption{MLE estimation of surrogate $\hat{\lambda}$ on 2023\char45 05\char45 12}
\end{figure}

For each day, we estimate a piecewise\char45 linear intensity $\hat{\lambda}$ from the price arrivals $(P_m)_{m=1}^M$ by maximum likelihood and use it as a surrogate realization of $\boldsymbol{\lambda}$. Polynomial and Gaussian\char45 kernel parametrizations were unstable in DDPM training, while piecewise\char45 linear surrogates were robust. Figure \ref{fig:Ex_Intensity} displays the real price arrivals (data), the parametrization $\hat{\lambda}$, and a sample of generated price arrivals obtained from $\hat{\lambda}$ by standard thinning. To assess the fit, we apply the time\char45 rescaling theorem. We compute the compensator of $\hat{\lambda}$ at observed arrival times. The increments of the compensator between consecutive observed arrivals should be i.i.d. Exp$(1)$. Applying the exponential cumulative distribution function to these gaps yields values that should be i.i.d. $\mathrm{Unif}(0,1)$. Under a correct specification, the points should align along the 45\char45 degree line (Figure \ref{fig:Ex_Intensity_fit}).

From the EPEX training data, we estimate $1\,216$ surrogate intensities $\hat{\lambda}$ for supply and the same number for demand, one per day in each case. For the marks $\mathbf{\Delta S}$, we obtain $841\,501$ observations in dimension $8$. Pairing each mark with its corresponding price gives the $9$\char45 dimensional observations of $(\mathbf{P},\mathbf{\Delta S})$.

Thus, each curve is encoded in two parts: the price\char45 arrival structure, summarized by the surrogate intensity $\hat{\lambda}$, and the $8$\char45 dimensional volume\char45 increment marks $\mathbf{\Delta S}$, modelled conditionally on the arrivals.

\section{Learning methods}
\label{Sec:Learning methods}

\subsection{eXtreme Gradient Boosting (XGBoost)}
\label{Sec:eXtreme Gradient Boosting (XGBoost)}

XGBoost is a gradient\char45 boosted ensemble of regression trees. Given a training sample
\((x_i,y_i)_{i=1}^n\), with \(x_i\in\mathbb{R}^m\) and \(y_i\in\mathbb{R}\), the
prediction is represented as
\[
\hat y_i=\sum_{k=1}^{K} f_k(x_i),
\qquad f_k\in\mathcal{F},
\]
where \(\mathcal{F}\) is the class of regression trees. Each tree maps the input
features to a terminal leaf and assigns a constant score to that leaf. The model is
learned sequentially: at boosting iteration \(t\), a new tree is added to the
current ensemble according to
\[
\hat y_i^{(t)}
=
\hat y_i^{(t-1)}
+
\eta f_t(x_i),
\]
where \(\eta\) is the learning rate. The tree \(f_t\) is chosen by minimizing a regularized objective of the form
\[
\mathcal{L}^{(t)}
=
\sum_{i=1}^{n}
\ell_{\tau}\!\left(
y_i,\hat y_i^{(t-1)}+f_t(x_i)
\right)
+
\Omega(f_t).
\]
Here, \(\ell_{\tau}\) is the quantile loss, with \(\tau\) determining the targeted
conditional quantile, and \(\Omega\) penalizes tree complexity. In particular,
\(\gamma\) penalizes additional leaves, while \(\lambda\) and
\(\alpha_{\mathrm{reg}}\) control the \(L^2\)- and \(L^1\)-regularization of the
leaf scores.

XGBoost optimizes this objective using a second\char45 order approximation of the loss,
based on the first\char45 and second\char45 order derivatives $g_i
=
\partial_{\hat y}
\ell_{\tau}(y_i,\hat y_i^{(t-1)})$, $h_i
=
\partial_{\hat y}^{2}
\ell_{\tau}(y_i,\hat y_i^{(t-1)})$. Thus, each split and each leaf score are selected using both gradient and curvature
information. Remaining hyperparameters control tree construction: \(d_{\max}\) limits the tree depth, \(H_{\min}\) sets
the minimum child weight, and \(\rho_{\mathrm{row}}\) and \(\rho_{\mathrm{col}}\)
define the row\char45 and feature\char45 sampling rates. For further details on XGBoost, see \cite{chen2016xgboost}.

\subsection{Denoising Diffusion Probabilistic Models (DDPMs)}
\label{Sec:Denoising Diffusion Probabilistic Models (DDPMs)}

Following \cite{ho2020denoising}, DDPMs are generative models based on a diffusion mechanism consisting of two stages: a forward Markovian corruption process and a reverse denoising process. They are also closely related to score\char45 based generative models through the SDE formulation of \cite{song2021scorebased}.

Let $x_0 \sim q_{\mathrm{data}}$ be a data sample in $\mathbb{R}^d$. The forward DDPM process is the Markov chain
\[
q(x_t\mid x_{t-1})=
\mathcal{N}\left(
\sqrt{1-\beta_t}x_{t-1}, \, 
\beta_t I
\right),
\]
where $(\beta_t)_{t=1}^T$ is a variance schedule and $I$ the $d$\char45 dimensional identity matrix. Equivalently,
\[
q(x_t\mid x_0)=
\mathcal{N}\left(
\sqrt{\bar\alpha_t}x_0, \, 
(1-\bar\alpha_t)I
\right),
\quad \text{or} \quad 
x_t=
\sqrt{\bar\alpha_t}x_0
+
\sqrt{1-\bar\alpha_t}\,\epsilon,
\]
where $\bar\alpha_t=\prod_{s=1}^t(1-\beta_s)$ and $\epsilon\sim\mathcal{N}(0,I)$. For large $t$, this process transforms data samples into approximately Gaussian noise. Generation reverses this process. Since the true reverse transition is unknown, it is approximated by
\[
p_\theta(x_{t-1}\mid x_t)
=
\mathcal{N}\left(
\mu_\theta(x_t,t),\, 
\beta_t I
\right).
\]
Following \cite{ho2020denoising}, the mean can be parametrized through a neural network $\epsilon_\theta$ that predicts the added noise, trained with
\[
L(\theta)
:=
\mathbb{E}_{t,x_0,\epsilon}
\left[
\left\|
\epsilon
-
\epsilon_\theta\!\left(
\sqrt{\bar\alpha_t}x_0
+
\sqrt{1-\bar\alpha_t}\,\epsilon,
t
\right)
\right\|^2
\right],
\]
where $t$ is uniformly sampled and $\epsilon \sim \mathcal{N}(0,I)$. After training, samples are generated by drawing $x_T\sim\mathcal{N}(0,I)$ and iteratively applying the learned reverse transitions.

To include exogenous information, we use a conditional version with features $x^{\mathrm{cond}}$. The reverse transition becomes
\[
p_\theta(x_{t-1}\mid x_t,{x}^{\mathrm{cond}})
=
\mathcal{N}\left(
\mu_\theta(x_t,t,{x}^{\text{cond}}), \,
\beta_t I
\right),
\]
where the denoising neural network predicts $\epsilon_\theta\!\left(
\sqrt{\bar\alpha_t}x_0
+
\sqrt{1-\bar\alpha_t}\,\epsilon,
t,
{x}^{\mathrm{cond}}
\right).$  The reverse process generates samples from the conditional distribution $\mathcal{L}\!\left(X_0 \mid \boldsymbol{X}^{\mathrm{cond}}=x^{\mathrm{cond}}\right)$.

As illustrated in Figure \ref{fig:diagram_DDPM_labels}, during training the model receives the noisy sample $x_t=\sqrt{\bar\alpha_t}\,x_0+\sqrt{1-\bar\alpha_t}\,\epsilon$ together with $x^{\mathrm{cond}}$, and learns to predict the noise
$\epsilon$ added to $x_0$. 

After training, Algorithm \ref{alg:supply_gen} generates new samples by starting from Gaussian noise $x_T$ and recursively denoising it through the conditional reverse process. The final sample approximates a draw from $\mathcal{L}\!\left(X_0 \mid X^{\mathrm{cond}}=x^{\mathrm{cond}}\right)$, so the generated curve is consistent with the prescribed exogenous features.

\begin{figure}[H]
	\centering
    \resizebox{1\textwidth}{!}{%
\begin{tikzpicture}[
	box/.style={draw, rectangle, rounded corners, align=center,
		minimum width=0.5cm, minimum height=1.2cm,
		inner sep=2pt, outer sep=0pt},
	inputBox/.style={draw, rectangle, dashed, inner sep=2pt},
	arrow/.style={-{Latex[length=3mm]}, thick},
	node distance=4mm and 8mm
	]
	
	
	\node[box] (observation) at (0, 0.75){
		Data\\
		$(x_0,\,{x}^{\mathrm{cond}})
		\in \Rr^d \times \Rr^l$
	};
	
	\node[box, below=0.8cm of observation] (noise) {
		Noise\\
		$\epsilon \sim \mathcal{N}(0,I)$\\
		Diffusion step\\
		$t \sim \mathrm{Uniform}\{1,\ldots,T\}$
	};
	
	\node[inputBox, fit=(observation) (noise)] (input) {};
	
	
	\node[box, right=2cm of observation] (context) {
		Conditioning features\\
		${x}^{\mathrm{cond}}$
	};
	
	\node[box, below=of context] (noised) {
		Noisy sample\\
		$x_t
		=
		\sqrt{\bar\alpha_t}\,x_0
		+
		\sqrt{1-\bar\alpha_t}\,\epsilon$\\
		$t$
	};
	
	\node[inputBox, fit=(context) (noised)] (input2) {};
	
	
	\node[box, right=0.5cm of input2] (nn) {
		Neural network\\
		Input:
		$(x_t,{x}^{\mathrm{cond}},t)$\\
		Output:
		$\epsilon_\theta(x_t,t,{x}^{\mathrm{cond}})
		\in \Rr^d$
	};
	
	\node[box, below=0.5cm of nn] (nnpred) {
		Loss computation\\
		$\left\|
		\epsilon
		-
		\epsilon_\theta(x_t,t,{x}^{\mathrm{cond}})
		\right\|^2$
	};
	
	\draw[arrow] (input.east)++(0,-0.35cm) -- (noised.west);
	\draw[arrow, dashed] (observation.east) -- (context.west);
	
	\draw[arrow] (input2.east) -- (nn.west);
	
	\draw[arrow] (nn.south) -- (nnpred.north);
	
\end{tikzpicture}
}
	\caption{Loss computation of the conditional DDPM}
	\label{fig:diagram_DDPM_labels}
\end{figure}

\begin{algorithm}[H]
	\caption{Conditional DDPM sampling}
	\label{alg:supply_gen}
	\begin{algorithmic}[1]
		\REQUIRE Conditioning features ${x}^{\mathrm{cond}} \in \Rr^l$, denoising steps $T$, variance schedule $(\beta_t)_{t=1}^T$, trained $\epsilon_{\theta}$
		\STATE $\alpha_t = 1-\beta_t$, $\bar\alpha_t = \prod_{s=1}^t \alpha_s$
		\STATE Sample $x_T \sim \mathcal{N}(0,I)$
		\FOR{$t = T$ {\bf down to} $1$}
			\STATE Predict noise: $\epsilon=\epsilon_\theta(x_t,t,{x}^{\mathrm{cond}})$
			\STATE Compute the reverse mean: 
            $\mu
			=
			\frac{1}{\sqrt{\alpha_t}}
			\left(
			x_t
			-
			\frac{\beta_t}{\sqrt{1-\bar\alpha_t}}
			\epsilon
			\right)$
			\IF{$t>1$}
				\STATE Sample $z \sim \mathcal{N}(0,I)$
				\STATE $x_{t-1}
				=
                \mu
				+
				\sqrt{\beta_t}\,z$
			\ELSE
				\STATE $x_0
				=
                \mu$
			\ENDIF
		\ENDFOR
		\RETURN $x_0$
	\end{algorithmic}
\end{algorithm}

We use the following parameters in Algorithm \ref{alg:supply_gen}. The number of diffusion steps is $T=501$, and the noise schedule is linear, with variance coefficients increasing from $\beta_1=10^{-4}$ to $\beta_T=0.02$. 

\section{Forecast implementation and curve\char45 level evaluation}
\label{Sec:Implementation and results}

Within the 2021\char45 01\char45 01 to 2024\char45 12\char45 31 dataset described in Section \ref{Sec:EPEX data and curve structure}, we train on observations up to 2024\char45 04\char45 30, and test on the remaining period, from 2024\char45 05\char45 01 to 2024\char45 12\char45 31.

\subsection{Features for the parametric model}
\label{Sec:Features for the parametric model}

We forecast the eight curve parameters in \eqref{eq:Eight parameters} using lagged curve information, cross\char45 hour dependence, order descriptors, and external covariates. The lagged parameter features treat each parameter as a time series observed up to the day before prediction, while the cross\char45 hour features account for intra\char45 day dependence among selected delivery hours. Order descriptors summarize the original EPEX curves as point clouds of price\char45 volume pairs.

\begin{figure}[H]
    \centering
    \includegraphics[width=0.8\textwidth]{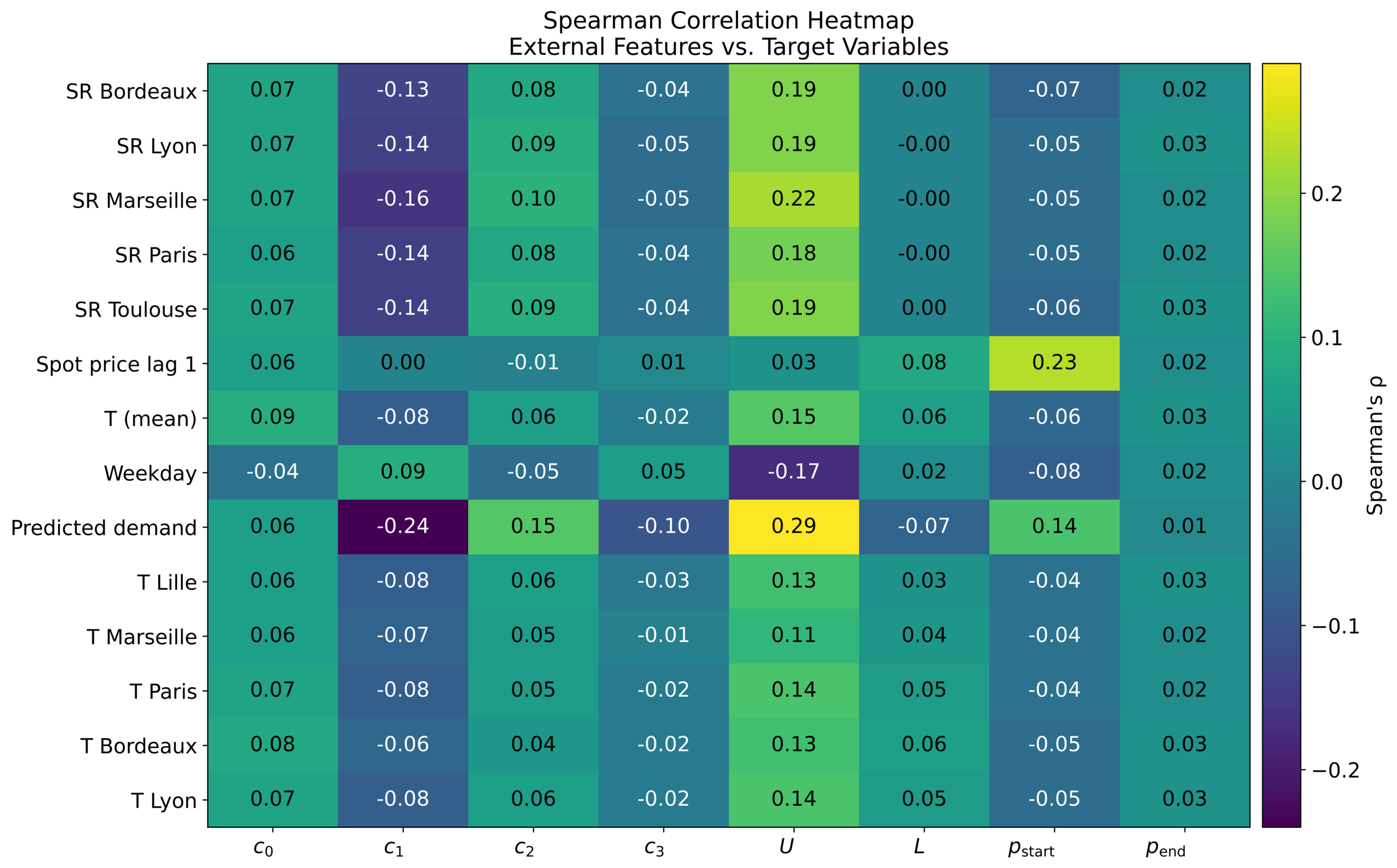}
    \caption{Spearman correlation heatmap between the target coefficients and the external features for the demand forecast. SR: Solar irradiance. T: Temperature.}
    \label{fig:corr_ext}
\end{figure}

The external covariates include calendar and event indicators, demand forecasts\footnote{Day\char45 ahead electricity demand forecasts from ENTSO\char45 E: \url{https://transparency.entsoe.eu/}.}, weather variables\footnote{\href{https://cds.climate.copernicus.eu/datasets/reanalysis-era5-single-levels}{Copernicus ERA5}. These data serve as a public proxy for operational weather forecasts.}, and the previous day's electricity price.\footnote{Day\char45 ahead electricity price from EPEX SPOT.} External data are adjusted for daylight\char45 saving time changes. Figure \ref{fig:corr_ext} reports Spearman correlations between some external features and predicted parameters; we retain features with $|\rho|\geq 0.05$. Table \ref{tab:parametric_features_summary} summarizes the final feature groups and their counts. The exact feature composition differs between demand and supply, mainly in the lagged parameter structure and weather variables.

\begin{table}[H]
\centering
\resizebox{\textwidth}{!}{
\begin{tabular}{|c|p{9.5cm}|c|c|}
\hline
Feature group & Variable & Count (D) & Count (S) \\
\hline

Calendar and curve\char45 shape
&
Year, month, hour, event indicators (national event, holiday, night, peak, weekend, weekday, winter holiday), convexity ratio, order counts, count ratios, counts before the elastic segment, and kurtosis.
& 15 & 15 \\
\hline

Weather
&
Temperature for Lille, Lyon, Marseille, Bordeaux, Paris, and Toulouse; solar irradiance, except for Lille; mean temperature for France. In addition, for the supply, relative humidity, except for Lille; wind speed, except for Lille and Bordeaux; 
& 12 & 21 \\
\hline

Day\char45 ahead
&
Spot price and predicted demand.
& 2 & 2 \\
\hline

Parameters \eqref{eq:Eight parameters}
&
Demand: lags 1 to 14 and 30; rolling standard deviations over 12, 24, and 168 days, except for $L$ and $c_1$. Supply: lags 1, 2, 3, 7, 14, and 30.
& 138 & 48 \\
\hline

Cross\char45 hour
&
Demand: Lag 1 values of $U$ and $L$ for hours \eqref{eq:Selected hours}. Supply: Lag 1 values of $U$, $L$, $p_{\mathrm{start}}$, and $p_{\mathrm{end}}$ for hours \eqref{eq:Selected hours}. 
& 16 & 32 \\
\hline

Order descriptors
&
Minimum, mean, and maximum order volumes at lags 1, 7, 14, and 30 days. Previous maximum\char45 order price, size, value, and volume.
& 17 & 17 \\
\hline

\textbf{Total}
&
&
\textbf{200} & \textbf{135} \\
\hline

\end{tabular}}
\caption{Summary of the external features used in the demand (D) and supply (S) parametric models.}
\label{tab:parametric_features_summary}
\end{table}

\subsection{Implementation of the parametric model and results}

Using the features defined in Section \ref{Sec:Features for the parametric model},
we estimate one XGBoost model for each target parameter and each hour. For both
demand and supply, the targets are the parameters \eqref{eq:Eight parameters}. The model is fast enough to allow
an exhaustive hyperparameter search; the selected values are reported in
Table \ref{tab:hyperparams_xgb}, separately for demand and supply. Estimating all eight parameters takes less than one minute on commodity hardware,\footnote{Performed on a computer with an AMD Ryzen 7 5800X3D CPU and an Nvidia RTX 3070 GPU.} and predicting the curve for a single hour takes less than half a second. 

\begin{table}[H]
\centering
\resizebox{\textwidth}{!}{
\begin{tabular}{|c|cc|cc|cc|cc|cc|cc|cc|cc|cc|}
\hline
\multirow{2}{*}{Target} 
& \multicolumn{2}{c|}{$\tau$} 
& \multicolumn{2}{c|}{$\eta$} 
& \multicolumn{2}{c|}{$d_{\max}$} 
& \multicolumn{2}{c|}{$H_{\min}$} 
& \multicolumn{2}{c|}{$\rho_{\mathrm{row}}$} 
& \multicolumn{2}{c|}{$\rho_{\mathrm{col}}$} 
& \multicolumn{2}{c|}{$\lambda$} 
& \multicolumn{2}{c|}{$\alpha_{\mathrm{reg}}$} 
& \multicolumn{2}{c|}{$\gamma$} \\
\cline{2-19}
& D & S & D & S & D & S & D & S & D & S & D & S & D & S & D & S & D & S \\
\hline
$c_0$ 
& 0.5 & 0.7 & 0.030 & 0.030 & 3 & 3 & 3 & 6  & 0.6 & 0.6 & 1.0 & 1.0 & 1  & 0.5  & 1.0 & 0.8 & 0.3 & 0.0 \\

$c_1$ 
& 0.5 & 0.5 & 0.005 & 0.010 & 3 & 3 & 7 & 8  & 0.4 & 0.5 & 0.8 & 0.8 & 10 & 8.7  & 1.0 & 0.8 & 0.5 & 0.9 \\

$c_2$ 
& 0.5 & 0.3 & 0.030 & 0.020 & 5 & 6 & 5 & 15 & 0.6 & 0.5 & 0.6 & 0.6 & 10 & 15.1 & 1.0 & 1.3 & 0.3 & 0.4 \\

$c_3$ 
& 0.5 & 0.7 & 0.010 & 0.010 & 3 & 5 & 1 & 2  & 0.7 & 0.7 & 0.8 & 0.7 & 1  & 1.3  & 0.0 & 0.0 & 0.0 & 0.0 \\

$U$ 
& 0.5 & 0.5 & 0.030 & 0.050 & 5 & 3 & 1 & 2  & 0.7 & 0.5 & 1.0 & 1.0 & 1  & 1.5  & 0.0 & 0.0 & 0.0 & 0.5 \\

$L$ 
& 0.8 & 0.5 & 0.030 & 0.040 & 5 & 3 & 1 & 6  & 0.7 & 0.8 & 1.0 & 1.0 & 1  & 1.0  & 0.0 & 0.0 & 0.0 & 0.0 \\

$p_{\text{start}}$ 
& 0.4 & 0.7 & 0.010 & 0.010 & 3 & 2 & 1 & 6  & 0.7 & 0.6 & 0.8 & 0.7 & 1  & 1.4  & 0.0 & 0.0 & 0.0 & 0.0 \\

$p_{\text{end}}$ 
& 0.5 & 0.5 & 0.010 & 0.010 & 3 & 5 & 1 & 10 & 0.7 & 0.5 & 0.8 & 0.7 & 1  & 0.4  & 0.0 & 0.0 & 0.0 & 0.2 \\
\hline
\end{tabular}}
\caption{Chosen hyperparameters for each XGBoost model for demand (D) and supply (S).
\(\tau\): quantile level. \(\eta\): learning rate. 
\(d_{\max}\): maximum depth. \(H_{\min}\) : minimum child weight. \(\rho_{\mathrm{row}}\): row\char45 sampling rate. \(\rho_{\mathrm{col}}\): feature\char45 sampling rate. \(\lambda\): \(L^2\) regularization. \(\alpha_{\mathrm{reg}}\): \(L^1\) regularization. \(\gamma\): split penalty.}
\label{tab:hyperparams_xgb}
\end{table}

After estimating the XGBoost models for the parameters in \eqref{eq:Eight parameters}, curve forecasts are obtained through the reconstruction procedure summarized in Figure \ref{fig:diagram_parametric_forecast}. For each day $d$ and hour $h$, the external features $X^{d,h}$ are passed through the corresponding XGBoost models to forecast the eight curve parameters. These predicted parameters are then used to reconstruct the elastic segment and plateau levels, after which monotonicity is enforced to obtain the final curve forecast.

\begin{figure}[H]
	\centering
    \resizebox{1\textwidth}{!}{%
\begin{tikzpicture}[
	box/.style={draw, rectangle, rounded corners, align=center,
		minimum width=0.5cm, minimum height=1.2cm,
		inner sep=2pt, outer sep=0pt},
	inputBox/.style={draw, rectangle, dashed, inner sep=2pt},
	arrow/.style={-{Latex[length=3mm]}, thick},
	node distance=4mm and 8mm
	]
	\node[box] (ExternalFeat) at (0, 0.75){
		External features $X^{d,h}$\\
        for day $d$ and hour $h$\\
		Table \ref{tab:parametric_features_summary}
	};
	\node[box, right=2.5cm of ExternalFeat] (XGB) {
        
        Forecast\\
        $(\hat{c}_0,\ldots,\hat{c}_3,\hat{U},\hat{L},\hat{p}_{\mathrm{start}},\hat{p}_{\mathrm{end}})$
	};
    \draw[arrow] 
(ExternalFeat) -- 
node[midway, above]{XGBoost} 
node[midway, below]{models for  \eqref{eq:Eight parameters}} 
(XGB);
    \node[box, right=0.8cm of XGB] (MonoE) {
		Monotonicity\\ enforcement
	};
    \draw[arrow] (XGB.east) -- (MonoE.west);
    \node[box, right=0.8cm of MonoE] (FCurve) {
		Demand curve \\
        forecast $P\mapsto \hat{D}(P)$
	};
    \draw[arrow] (MonoE.east) -- (FCurve.west);
\end{tikzpicture}}
	\caption{Parametric curve forecasting pipeline}
\label{fig:diagram_parametric_forecast}
\end{figure}

We evaluate the predicted parameters on the test set by reconstructing the corresponding curve and comparing it with two references: the curve reconstructed from ground\char45 truth parameters \eqref{eq:Eight parameters} and the original EPEX curve data points. We report MAE and its normalized version, rather than both MAE and Root Mean Square Error (RMSE), to keep the presentation concise. MAE gives a direct measure of the average discrepancy, whereas RMSE places more weight on extreme errors. The errors are computed separately for each hour in \eqref{eq:Selected hours} and aggregated over the full test set.

\begin{figure}[H]
  \centering
  \begin{subfigure}{0.495\linewidth}
    \centering
    \includegraphics[width=\linewidth]{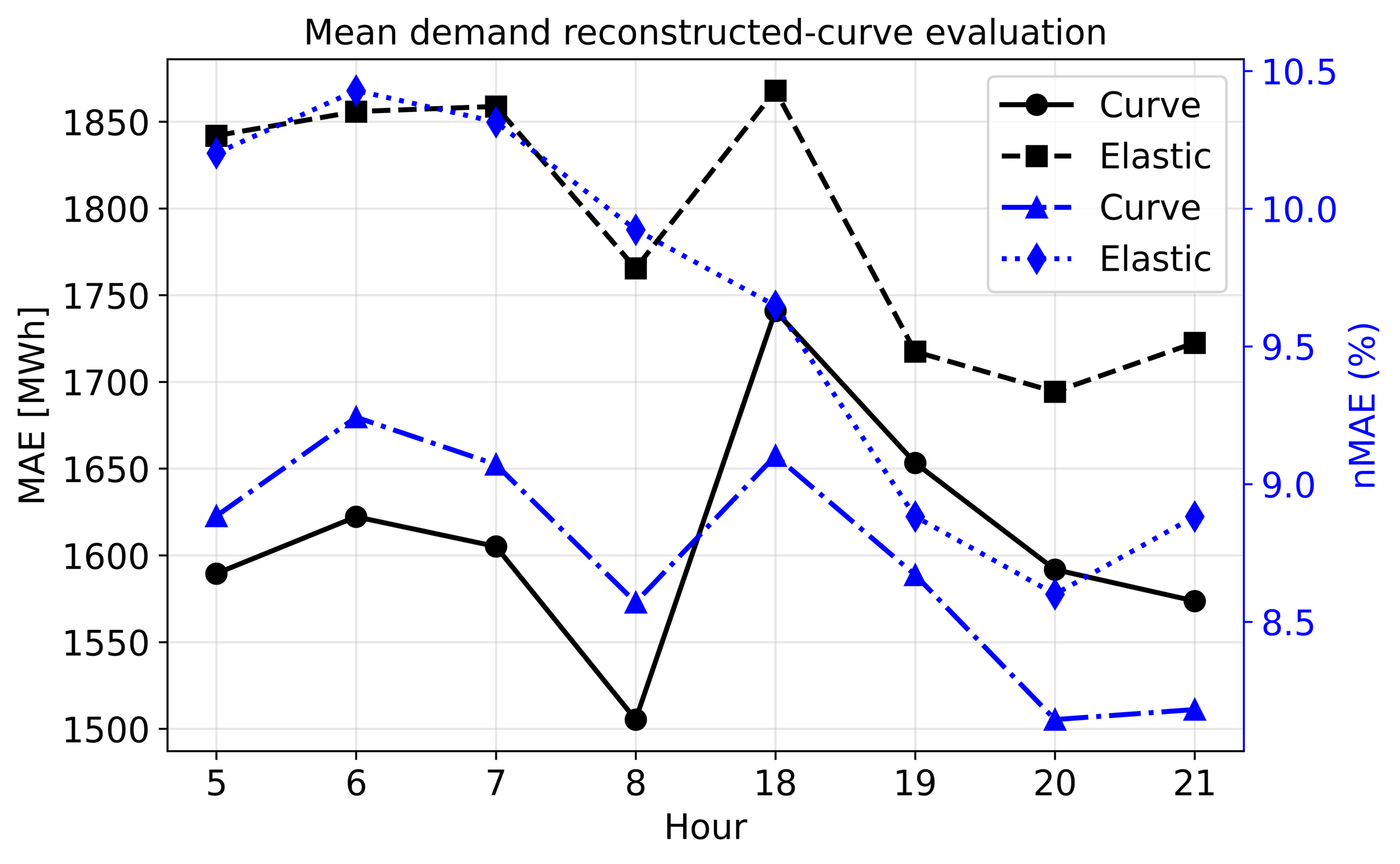}
    \caption{Demand}
    \label{fig:Mean MAE and nMAE demand}
  \end{subfigure}\hfill
  \begin{subfigure}{0.495\linewidth}
    \centering
    \includegraphics[width=\linewidth]{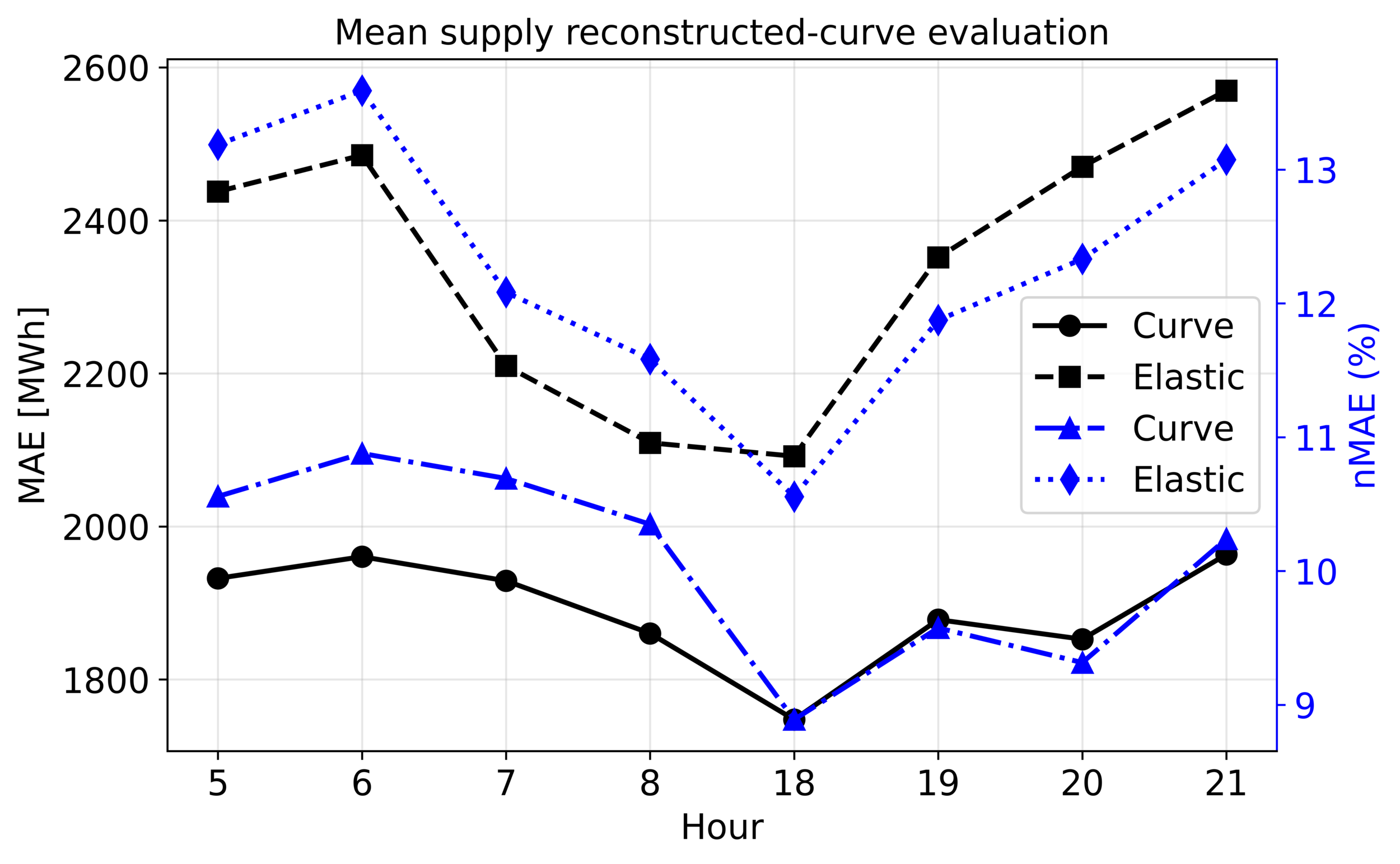}
    \caption{Supply}
    \label{fig:Mean MAE and nMAE demand supply}
  \end{subfigure}
  \caption{Mean MAE and nMAE between curves reconstructed from predicted and true parameters, reported for each hour. Demand results are shown on the left and supply results on the right.}
  \label{fig:First_evaluation}
\end{figure}

For the first evaluation, let $(p_{j})_{j=1}^{2\,000}$ denote a uniform price grid. For each test observation $i \in \{1,\dots,N_{\mathrm{test}}\}$, with $N_{\mathrm{test}}=245$, and hour $h$, we compute the MAE over the curve, over the elastic segment, and its normalized versions, as
\begin{equation}
\label{eq:MAES parametric}
\begin{alignedat}{2}
\mathrm{MAE}_{i,h}^{\mathrm{par}}
&=
\frac{1}{2\,000}
\sum_{j=1}^{2\,000}
\left|
D^h_i(p_{j})
-
\hat{D}^h_i(p_{j})
\right|,
\qquad
&
\mathrm{MAE}_{i,h,\mathrm{el}}^{\mathrm{par}}
&=
\frac{1}{|J^{h,\mathrm{el}}_i|}
\sum_{j\in J^{h,\mathrm{el}}_i}
\left|
D^h_i(p_{j})
-
\hat{D}^h_i(p_{j})
\right|,
\\[0.5em]
n\mathrm{MAE}_{i,h}^{\mathrm{par}}
&=
\frac{\mathrm{MAE}_{i,h}^{\mathrm{par}}}{(U^h_i + L^h_i)/2},
\qquad
&
n\mathrm{MAE}_{i,h,\mathrm{el}}^{\mathrm{par}}
&=
\frac{\mathrm{MAE}_{i,h,\mathrm{el}}^{\mathrm{par}}}{(U^h_i + L^h_i)/2},
\end{alignedat}
\end{equation}
where $D^h_i(p_j)$ and $\hat{D}^h_i(p_j)$ denote the ground\char45 truth and predicted reconstructed demand volumes at price $p_j$, respectively, $U^h_i$ and $L^h_i$ denote the upper and lower plateau levels, respectively, and $J^{h,\mathrm{el}}_i$ is the set of grid indices corresponding to the elastic region of test observation $i$ and hour $h$. We summarize these errors over the test set by reporting their empirical mean and quartiles. Figure \ref{fig:First_evaluation} reports the mean MAE and nMAE for both demand and supply. For demand, the error over the full curve is about $2\%$ lower than over the elastic region. For supply, the corresponding difference is about $4\%$. Thus, the approximation error is larger in the elastic region, indicating that this segment is more difficult to reconstruct accurately.

\begin{figure}[H]
  \centering
  \begin{subfigure}{0.495\linewidth}
    \centering
    \includegraphics[width=\linewidth]{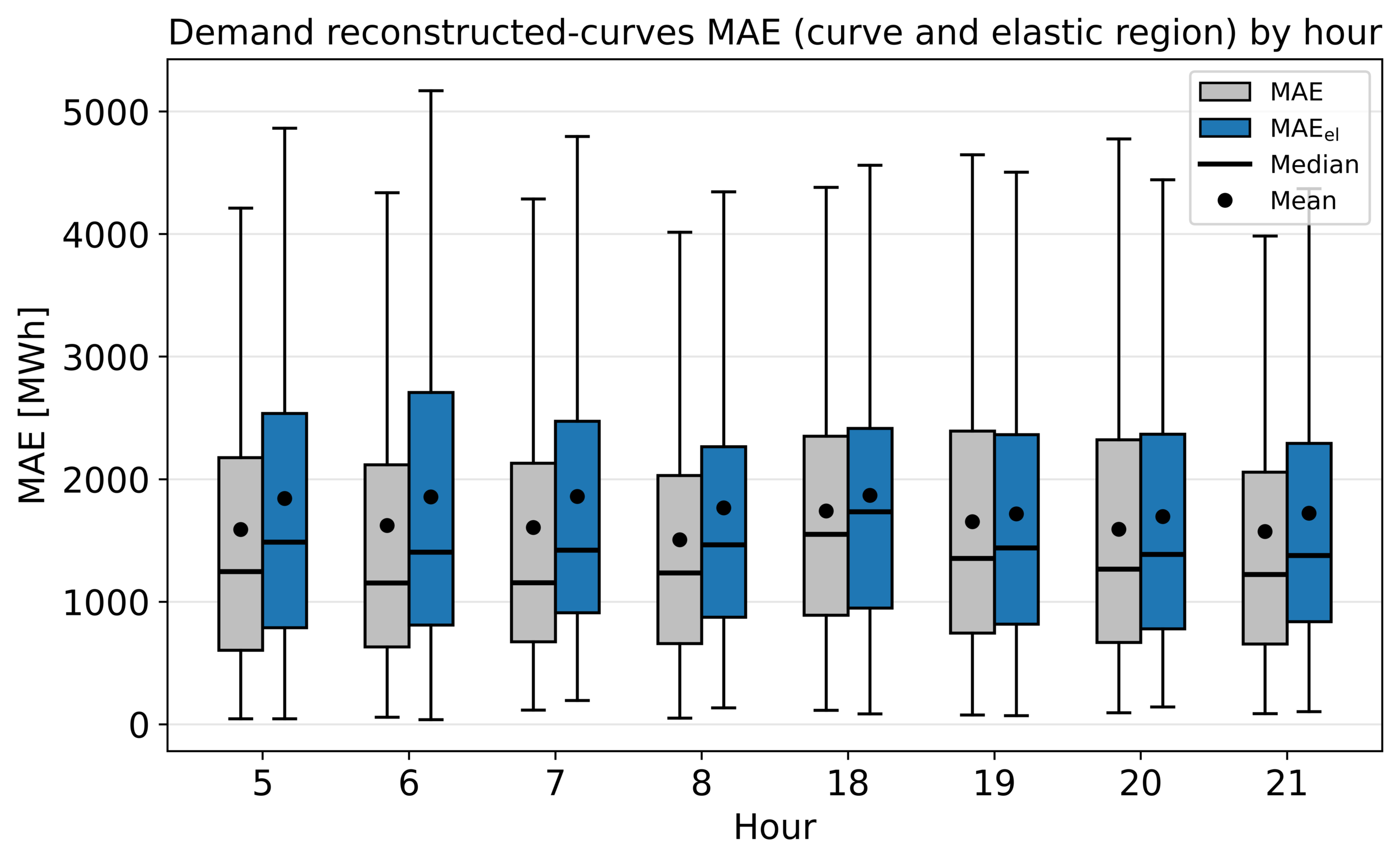}
    \caption{Demand}
    \label{fig:MAE boxplot demand}
  \end{subfigure}\hfill
  \begin{subfigure}{0.495\linewidth}
    \centering
    \includegraphics[width=\linewidth]{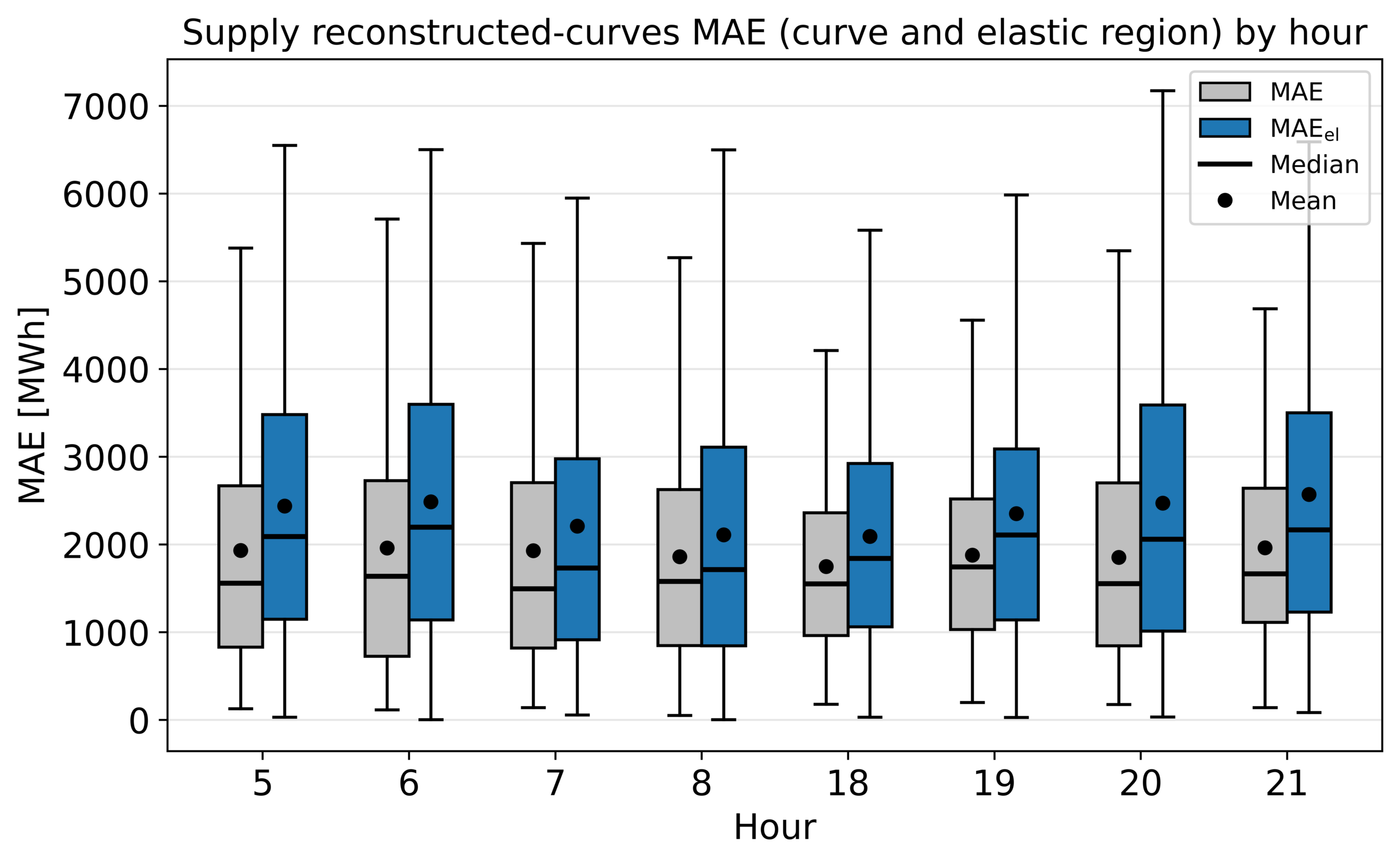}
    \caption{Supply}
    \label{fig:MAE boxplot supply}
  \end{subfigure}
  \caption{Distribution of MAE over the full curve and elastic region}
  \label{fig:First_evaluation_boxplot}
\end{figure}

For demand, hour 18 shows relatively high errors across the different regions, whereas hours 8, 20, and 21 show better performance across all regions. For supply, the performance gap between the full curve and the elastic region is similar across hours, although the errors are about $4\%$ higher than on the demand side. In this case, hours 7, 8, and 18 show better performance. Figure \ref{fig:First_evaluation_boxplot} displays boxplots of the MAE over the full curve and the elastic region for demand and supply. We observe no strong dependence of the error on the hour of the day, with generally better performance over the full curve, as observed above.

\begin{figure}[H]
  \centering
  \begin{subfigure}{0.495\linewidth}
    \centering
    \includegraphics[width=\linewidth]{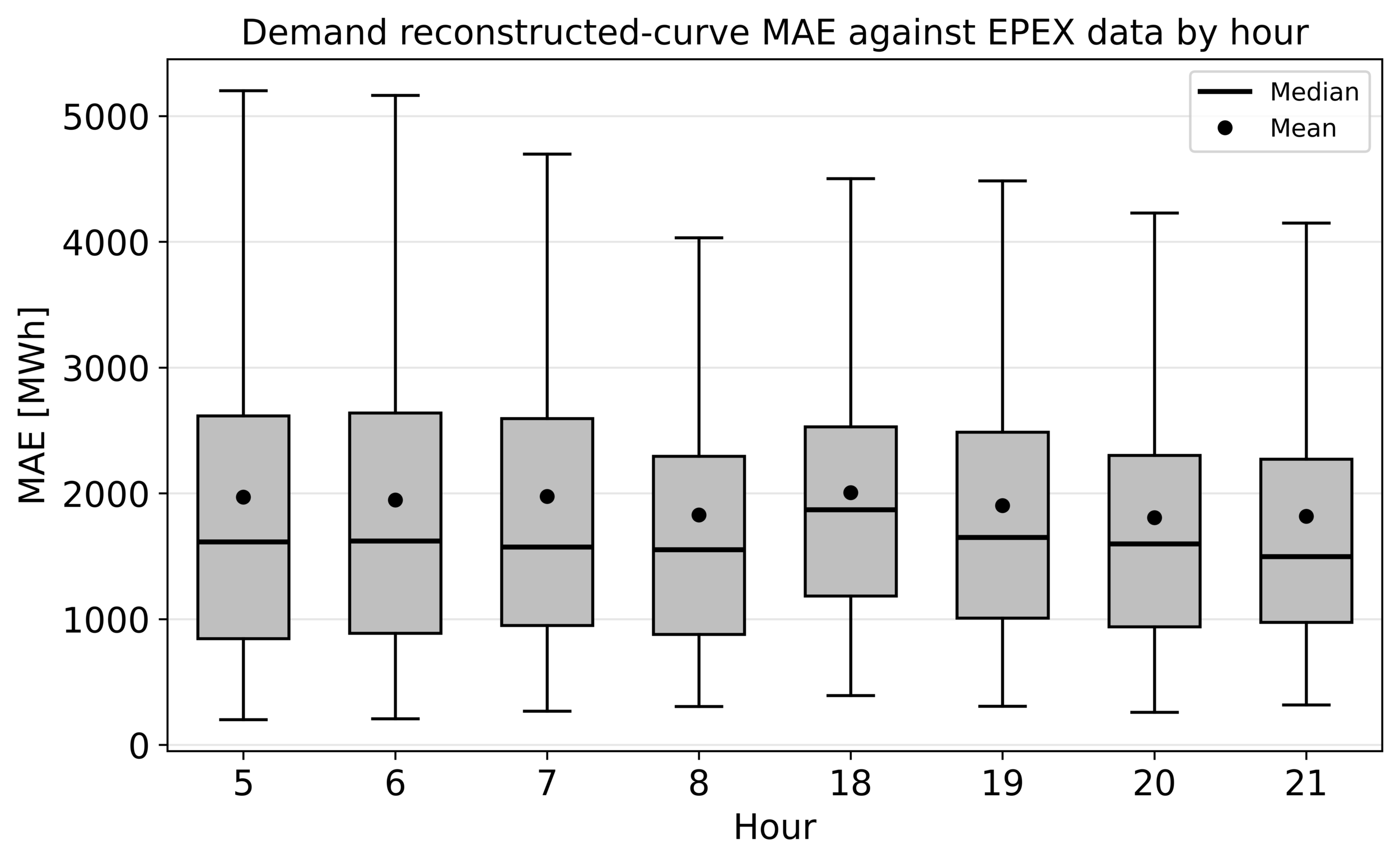}
    \caption{Demand}
    \label{fig:MAE boxplot demand EPEX}
  \end{subfigure}\hfill
  \begin{subfigure}{0.495\linewidth}
    \centering
    \includegraphics[width=\linewidth]{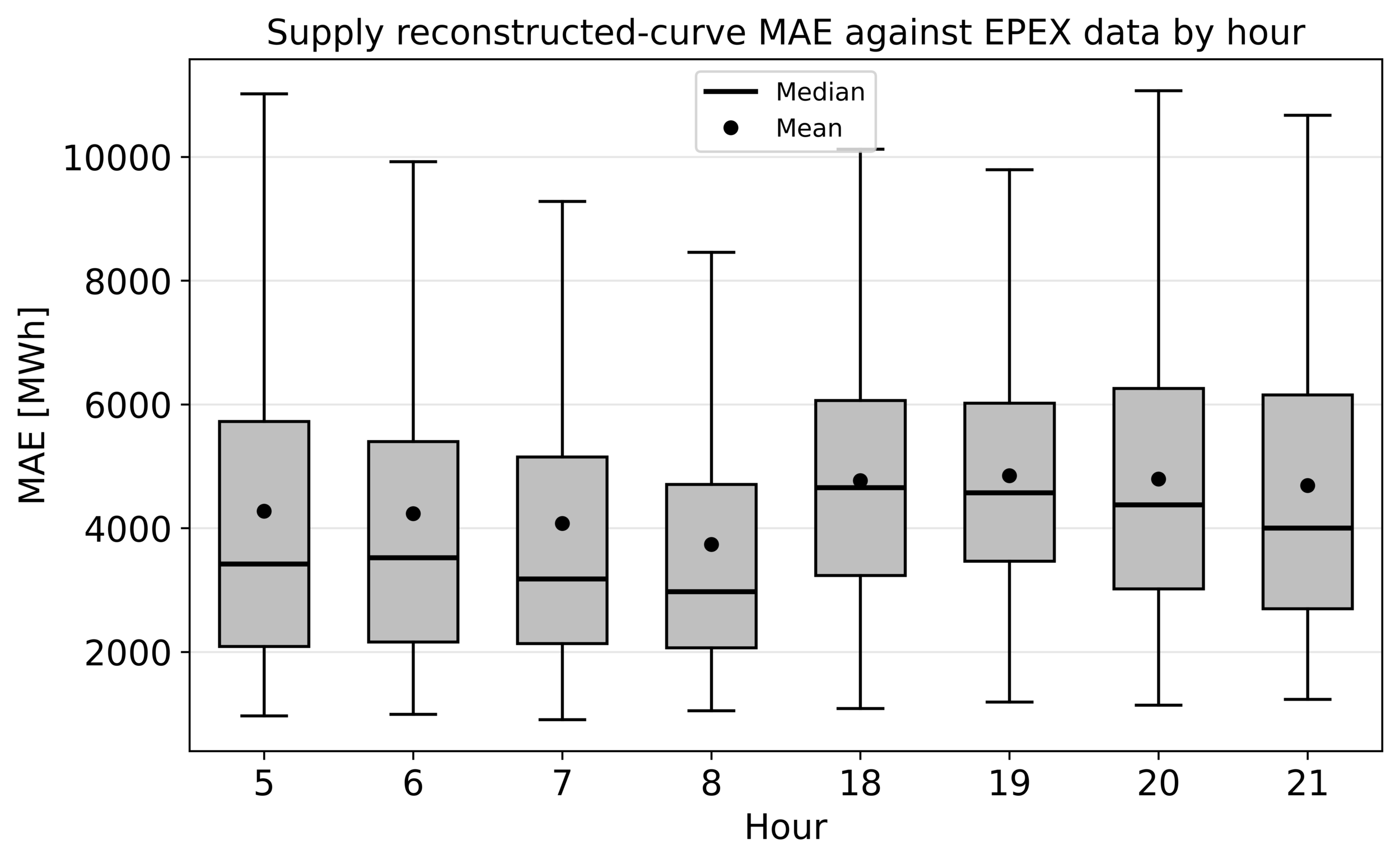}
    \caption{Supply}
    \label{fig:MAE boxplot supply EPEX}
  \end{subfigure}
  \caption{Distribution of MAE against EPEX data}
  \label{fig:Second_evaluation_MAE}
\end{figure}

For the second evaluation, we use the raw observed price\char45 volume pairs (EPEX data). Let 
\[
\{(p^h_{i,j},D^h_{i,j})\}_{j=1}^{n^h_i}
\]
denote the observed EPEX curve points for observation $i$ and hour $h$, where $n^h_i$ is the corresponding number of points. We evaluate the predicted reconstructed curve at the observed prices $p^h_{i,j}$ to obtain $\hat{D}^h_i(p^h_{i,j})$ and compute
\begin{equation}
\label{eq:MAES parametric raw}
\begin{alignedat}{2}
\mathrm{MAE}_{i,h}^{\mathrm{par}}
&=
\frac{1}{n^h_i}
\sum_{j=1}^{n^h_i}
\left|
D^h_{i,j}
-
\hat{D}^h_i(p^h_{i,j})
\right|.
\end{alignedat}
\end{equation}
As before, we summarize the errors over the test set using boxplots of the MAE, reported in Figure \ref{fig:Second_evaluation_MAE}. The distributions have similar shapes, although different magnitudes, for both demand and supply. The error distributions are narrower for hours 7, 8, 18, and 19, and the overall error performance does not vary substantially across hours.

\begin{figure}[H]
  \centering
  \begin{subfigure}{0.495\linewidth}
    \centering
    \includegraphics[width=\linewidth]{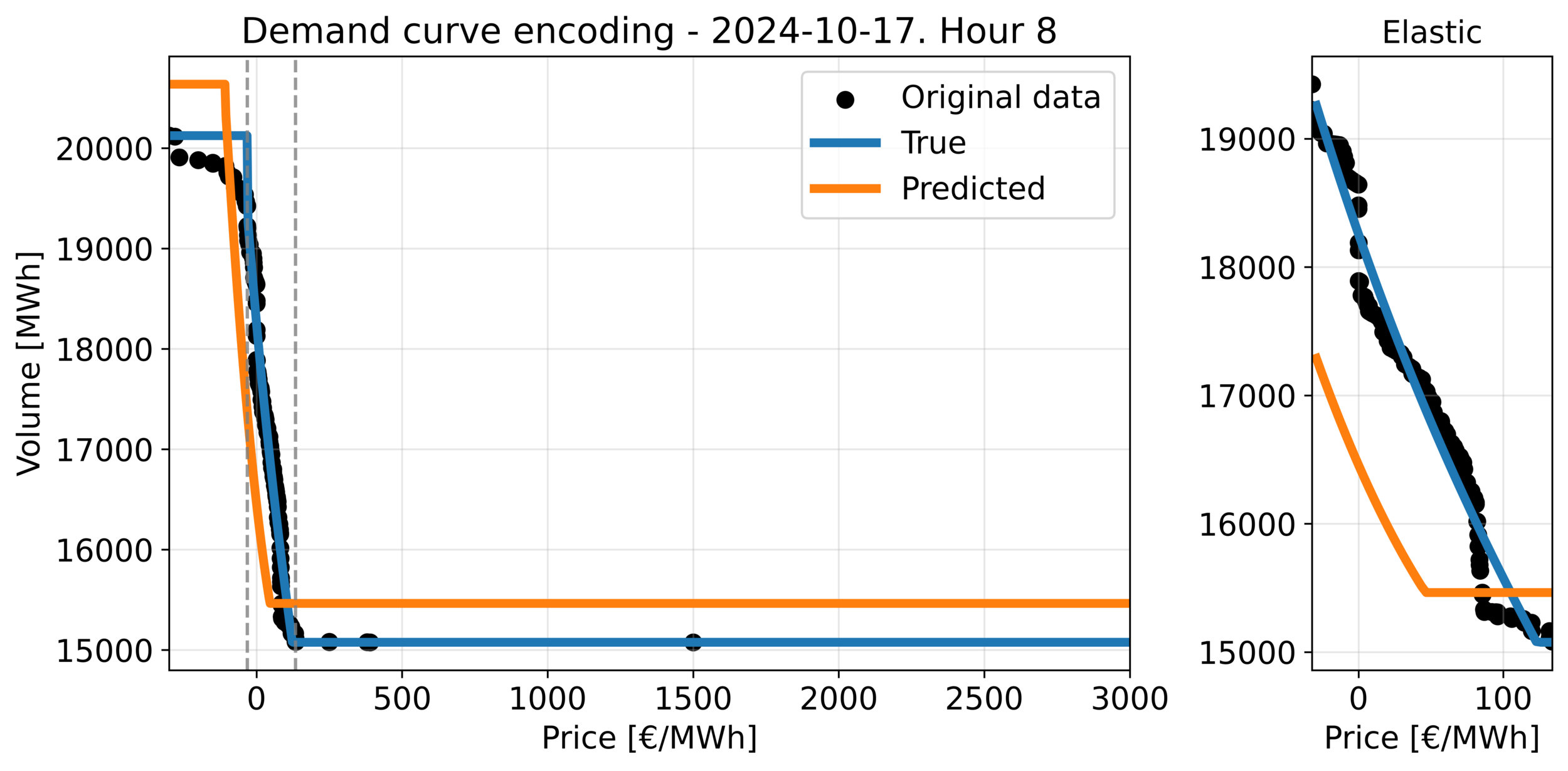}
    \caption{Demand, hour 8}
    \label{fig:MAE boxplot demand EPEX h6}
  \end{subfigure}\hfill
  \begin{subfigure}{0.495\linewidth}
    \centering
    \includegraphics[width=\linewidth]{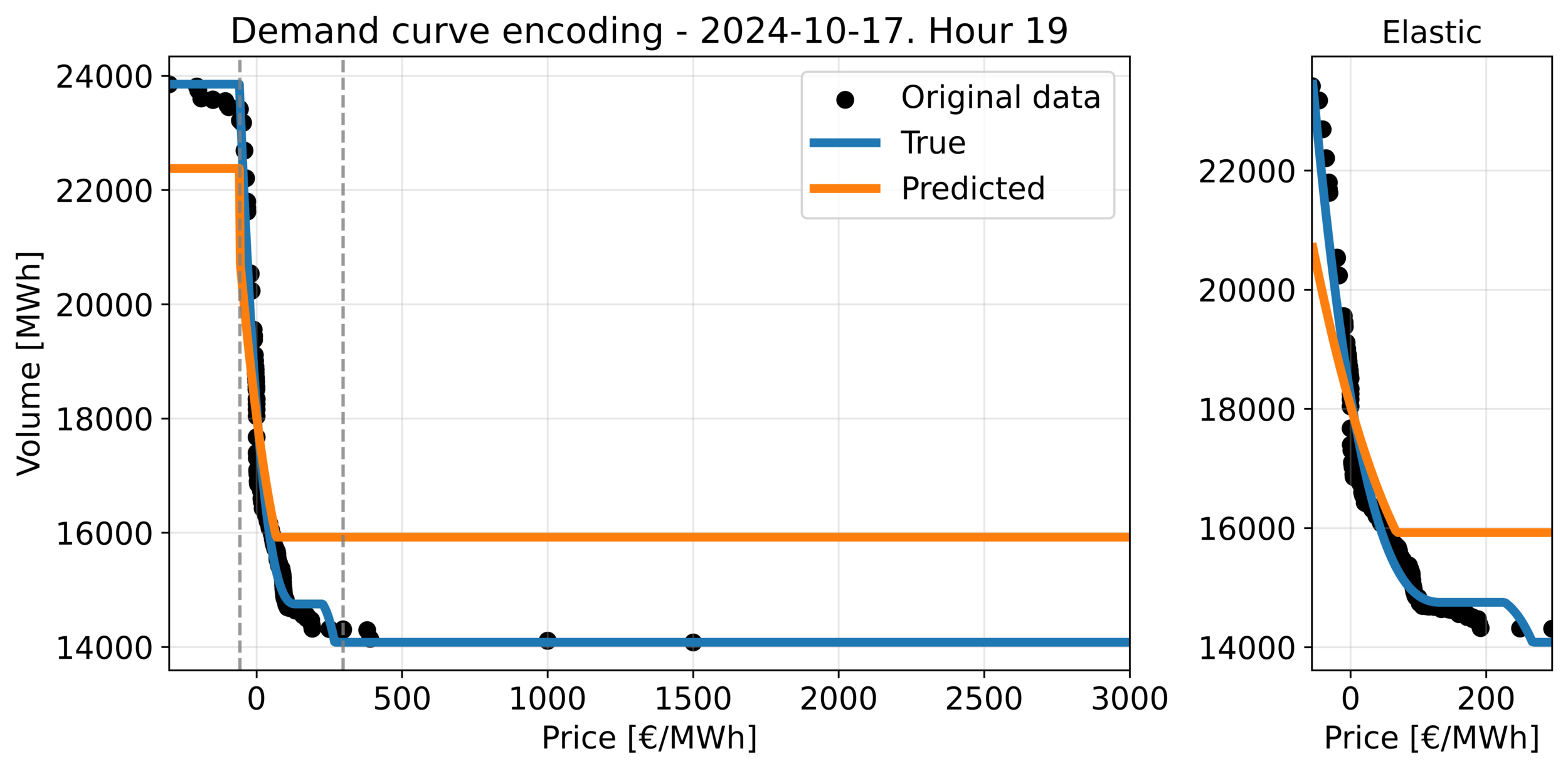}
    \caption{Demand, hour 19}
    \label{fig:MAE boxplot demand EPEX h19}
  \end{subfigure}

    \begin{subfigure}{0.495\linewidth}
    \centering
    \includegraphics[width=\linewidth]{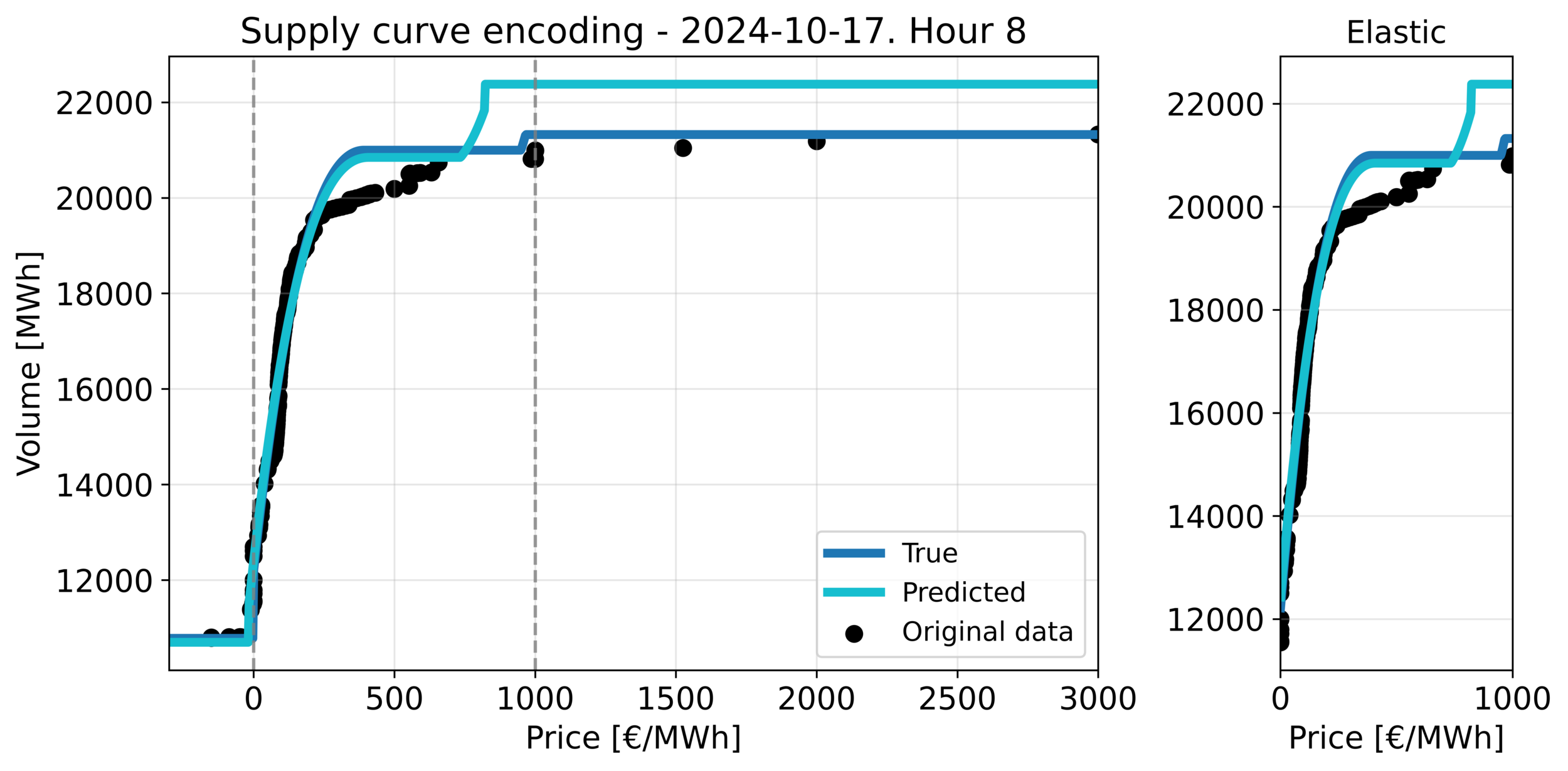}
    \caption{Supply, hour 8}
    \label{fig:MAE boxplot supply EPEX h6}
  \end{subfigure}\hfill
  \begin{subfigure}{0.495\linewidth}
    \centering
    \includegraphics[width=\linewidth]{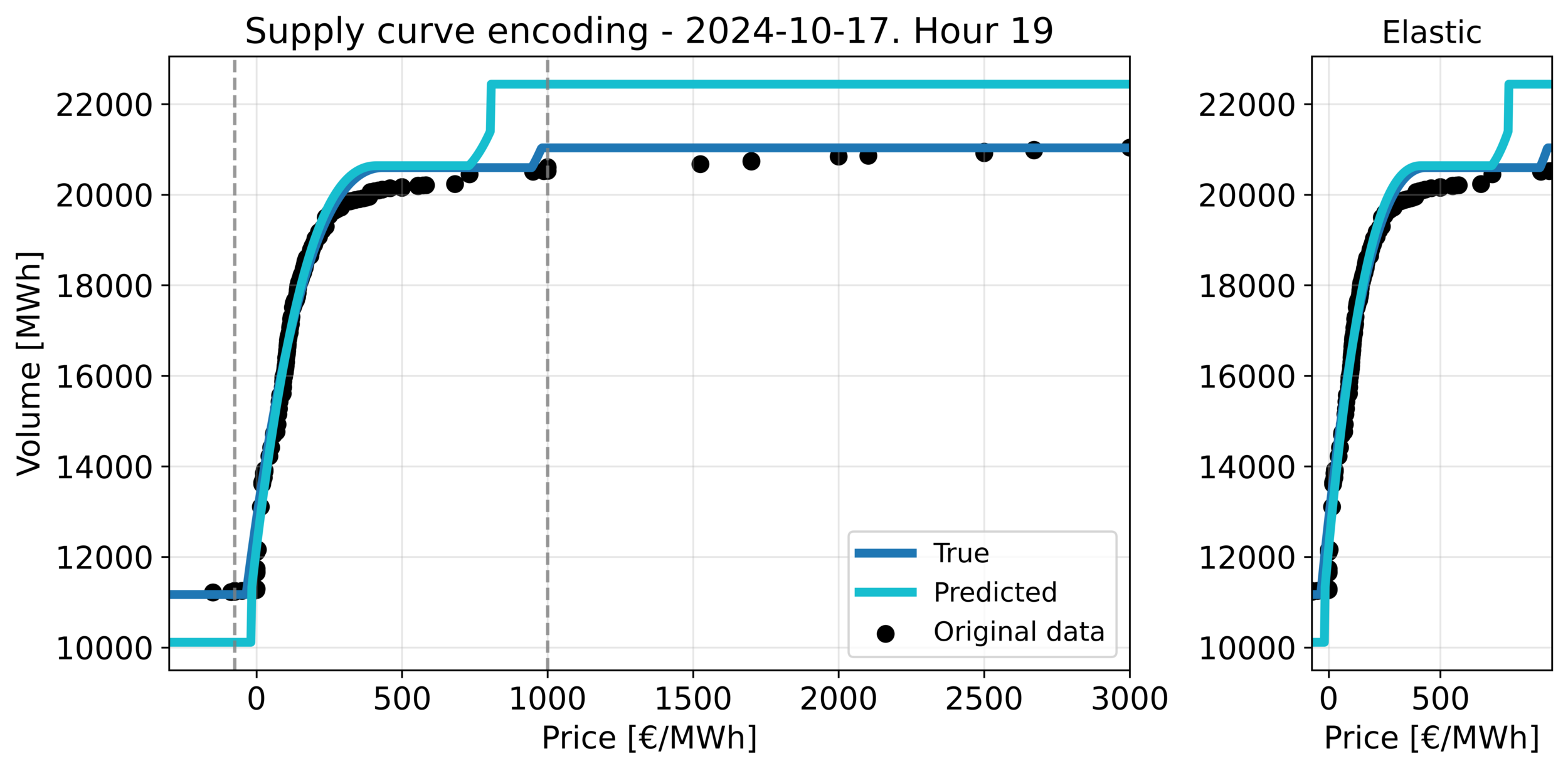}
    \caption{Supply, hour 19}
    \label{fig:MAE boxplot supply EPEX h19}
  \end{subfigure}
  
  \caption{Sample curves for 2024\char45 10 \char45 17, delivery hours 8 and 19.}
  \label{fig:Parametric Samples}
\end{figure}

We provide examples of predictions in Figure \ref{fig:Parametric Samples}. Discrepancies in the plateau regions can strongly affect the overall shape of the curve. However, the plateaus are generally well approximated, which explains the lower MAE over the full curve compared with the elastic segment. The figure illustrates that the elastic segment is the most challenging part for this method to capture.

A main advantage of the parametric encoding is its interpretability: the decomposition allows each coefficient to be understood directly. For instance, in the demand curves, an increase in $U$ indicates an increase in the system's energy absorption capacity. However, this representation does not retain information on the underlying order structure that generates each curve.

\subsection{Features for the generative model}
\label{Sec:Features for the generative model}

The price\char45 arrival features $\mathbf{X}^P$ and mark features $\mathbf{X}^S$ include Coal, Oil, and Gas prices, together with France\char45\footnote{Country\char45 level weather variables for France from the Copernicus ERA5 dataset, consisting of post\char45 processed area\char45 weighted spatial averages of the underlying gridded fields.} averaged minimum, maximum, and mean values of Air Temperature (AT), Global Horizontal Irradiance (GHI), and Wind Speed (WS). Table \ref{tab:summary_stats_transposed} reports the summary statistics. Fuel prices are computed as described in \ref{subsec:Fuel prices computation}.

\begin{table}[ht]
  \centering
  \large
  \setlength{\tabcolsep}{4pt} 
  \renewcommand{\arraystretch}{0.8}
  \scalebox{0.8}{
  \begin{tabular}{lccccccccccc}
    \hline   
      & \textbf{Coal}
      & \textbf{Oil}
      & \textbf{Gas}
      & \textbf{AT}
      & \textbf{AT}
      & \textbf{AT}
      & \textbf{GHI}
      & \textbf{GHI}
      & \textbf{WS}
      & \textbf{WS}
      & \textbf{WS} \\
    
      & 
      & 
      & 
      & \textbf{{mean}}
      & \textbf{{min}}
      & \textbf{{max}}
      & \textbf{{mean}}
      & \textbf{{max}}
      & \textbf{{mean}}
      & \textbf{{min}}
      & \textbf{{max}} \\      
      
      & {\scriptsize [\EUR/MWh]}
      & {\scriptsize [\EUR/MWh]}
      & {\scriptsize [\EUR/MWh]}
      & {\scriptsize [$^\circ$C]}
      & {\scriptsize [$^\circ$C]}
      & {\scriptsize [$^\circ$C]}
      & {\scriptsize [W/m$^2$]}
      & {\scriptsize [W/m$^2$]}
      & {\scriptsize [km/h]}
      & {\scriptsize [km/h]}
      & {\scriptsize [km/h]} \\
    \hline Min & 54.88 & 86.63 & 53.12 & -1.92 & -5.18 & 0.00 & 16.35 & 71.77 & 1.24 & 0.88 & 1.44 \\ Max & 252.85 & 244.33 & 844.76 & 27.52 & 21.16 & 35.89 & 338.48 & 907.87 & 7.25 & 6.38 & 8.50 \\ Mean & 133.52 & 167.98 & 182.29 & 12.26 & 8.49 & 16.30 & 151.98 & 484.16 & 3.14 & 2.45 & 3.87 \\ 
    \hline
  \end{tabular}
  }
  \caption{Summary statistics of fuel prices and weather features. AT: Air temperature. GHI: Global horizontal irradiance. WS: Wind speed. The minimum GHI is always $0.00$ and is therefore omitted from the table.}
  \label{tab:summary_stats_transposed}
\end{table}

\begin{figure}[H]
    \centering
    \includegraphics[width=\textwidth]{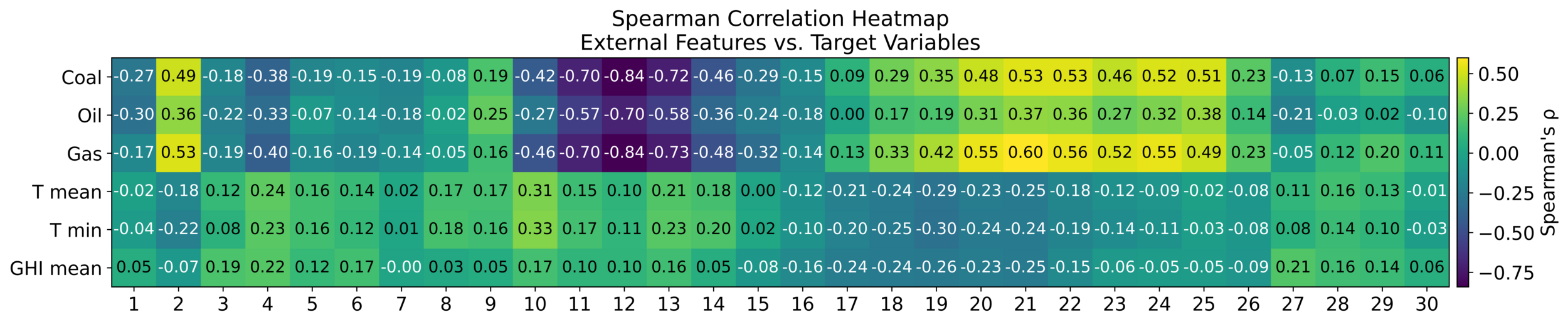}
    \caption{Spearman correlation heatmap between the fitted intensity values and selected features for the supply. GHI: Global solar irradiance. T: Temperature.}
    \label{fig:corr_ext_lambda}
\end{figure}

For the intensity $\boldsymbol{\lambda}$, we take the features $\mathbf{X}^P$ to include Coal, Oil, and Gas prices, together with minimum and mean temperature and mean global horizontal irradiance. To motivate this selection, Figure \ref{fig:corr_ext_lambda} reports the Spearman correlations between these features and the fitted intensity values over the 30\char45 point grid. For the demand, we add the minimum global horizontal irradiance to the previous features. 

\begin{figure}[H]
    \centering
    \includegraphics[width=0.7\textwidth]{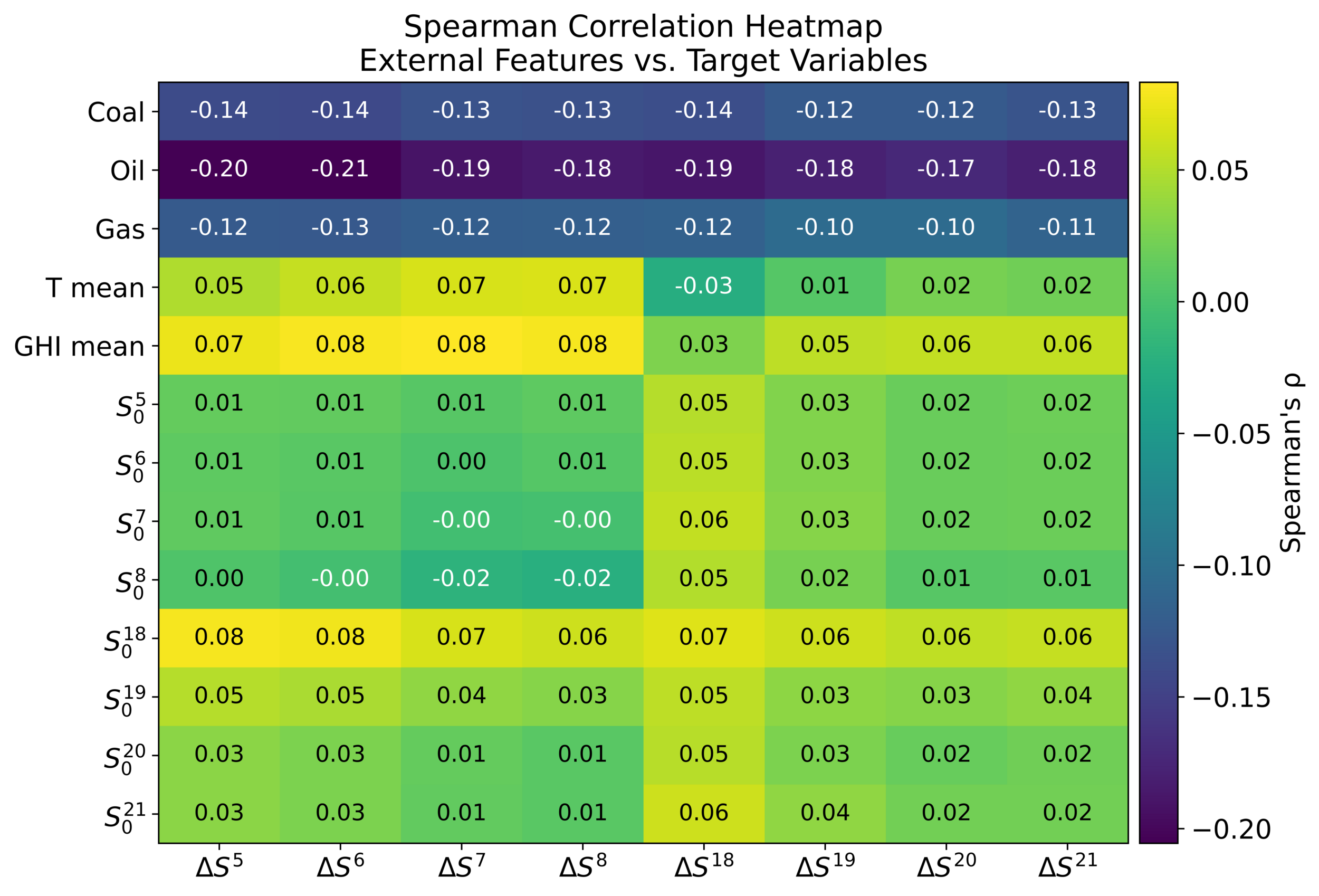}
    \caption{Spearman correlation heatmap between the marks and selected features for the supply. GHI: Global solar irradiance. T: Temperature. $S^h_0$: Initial volume at price $-300$ \EUR/MWh for hour $h$.}
    \label{fig:corr_ext_DV}
\end{figure}

For the marks $\boldsymbol{\Delta S}$, we take the features $\mathbf{X}^S$ to include Coal, Oil, and Gas prices, mean temperature and global horizontal irradiance, and the initial volume values of the curve\footnote{The volume values at price $-300$ \EUR/MWh.}. The selection is motivated by Figure \ref{fig:corr_ext_DV}, which reports the Spearman correlations between these features and the eight volume increments for the selected hours \eqref{eq:Selected hours}. For demand marks, we use the analogous initial demand values with the same atmospheric features.

\subsection{Implementation of the generative model and results}
\label{Sec:Implementation of the generative model and results}

Using the features defined in Section \ref{Sec:Features for the generative model}, we train separate DDPMs for the intensity $\boldsymbol{\lambda}$ and the marks $\mathbf{\Delta S}$ in \eqref{eq:Dependence}.

\subsubsection{Intensity model}

For the DDPM of the intensity $\boldsymbol{\lambda}$, $q_{\mathrm{data}}$ consists of $1\,216$ observations in dimension $30$, corresponding to the piecewise MLE surrogate estimates $\hat{\lambda}$; see Figure \ref{fig:Ex_Intensity}. Each observation is paired with the $6$ intensity features reported in the correlation table of Figure \ref{fig:corr_ext_lambda}. 
We train the model for $4\times 10^4$ epochs with a batch size of $256$. Training time was 10 minutes\footnote{Computations were performed in Google Colab Pro using a GPU\char45 accelerated runtime. 
Because Colab dynamically allocates resources, the exact CPU, RAM, and GPU model were session\char45 dependent; the reported runs used the GPU runtime available at execution time.}. 

\begin{figure}[H]
    \centering        
    \includegraphics[width=0.5\textwidth]{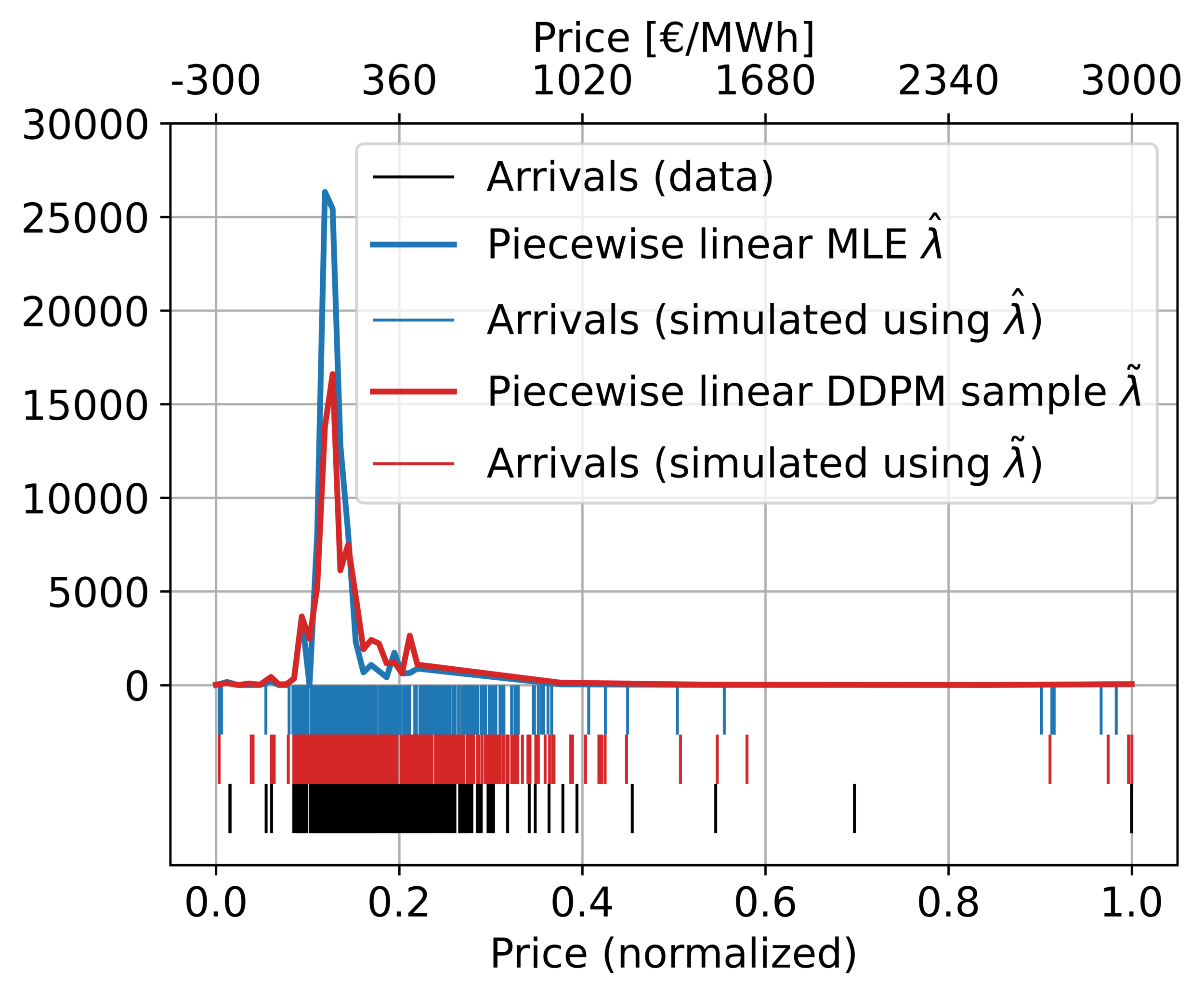}
    \caption{Arrivals (data), surrogates, and generated arrivals on 2023\char45 05\char45 12}
    \label{fig:Ex_lambda_param}
\end{figure}

Figure \ref{fig:Ex_lambda_param} compares the real surrogate $\hat{\lambda}$ of Figure \ref{fig:Ex_Intensity} with one generated $\tilde{\lambda}$, and displays further generated samples of arrival prices. We use the same modelling and training parameters for the demand side, yielding similar results.

\subsubsection{Marks model}

\begin{figure}[H]
    \centering
    \begin{subfigure}[t]{0.489\textwidth}
    \vspace{0pt}
        \includegraphics[width=\textwidth]{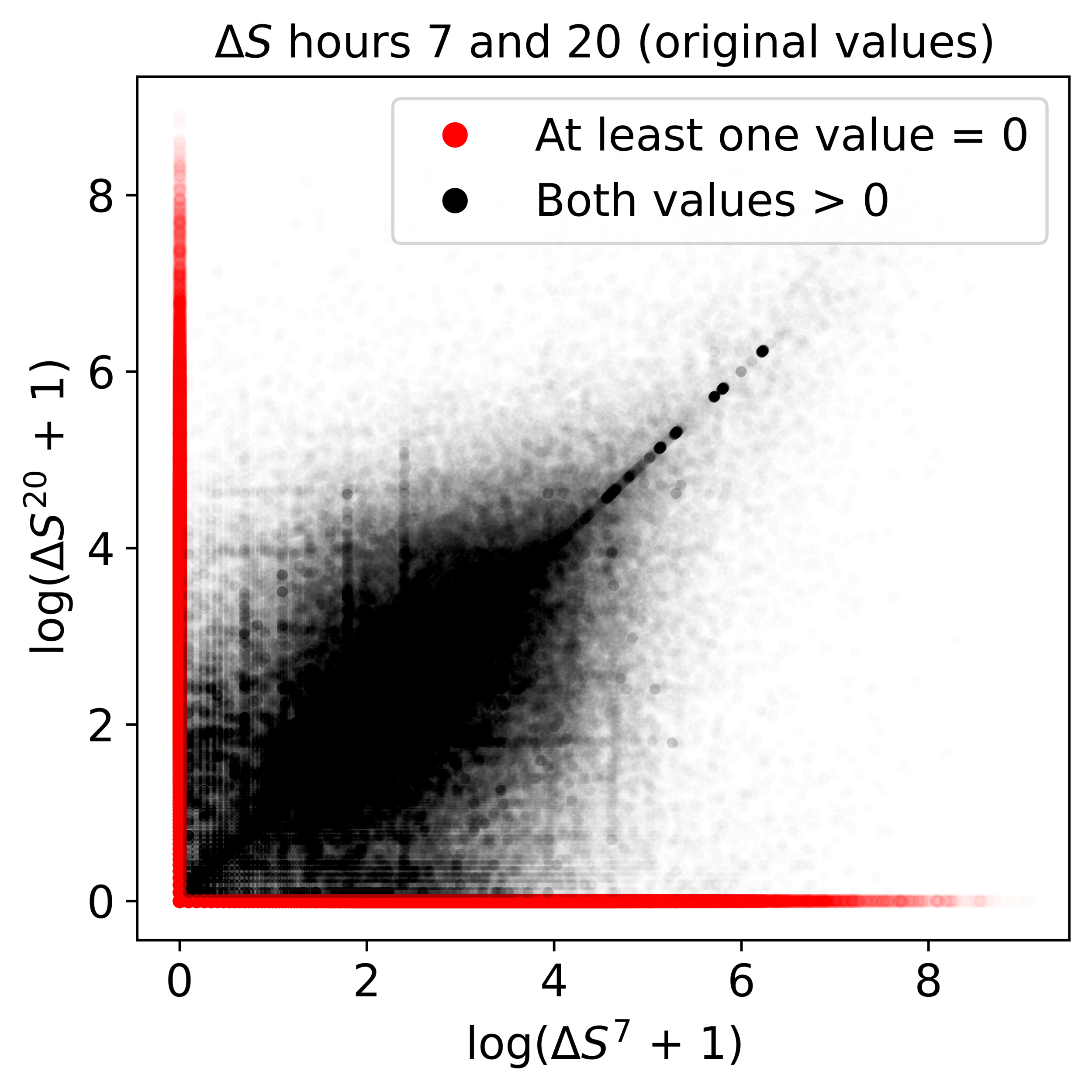}
    \caption{Pairs $(\Delta S^7,\Delta S^{20})$.}
    \label{fig:IMG_Original_DVZeros}
    \end{subfigure}
    \begin{subfigure}[t]{0.489\textwidth}
    \vspace{0pt}
        \includegraphics[width=\textwidth]{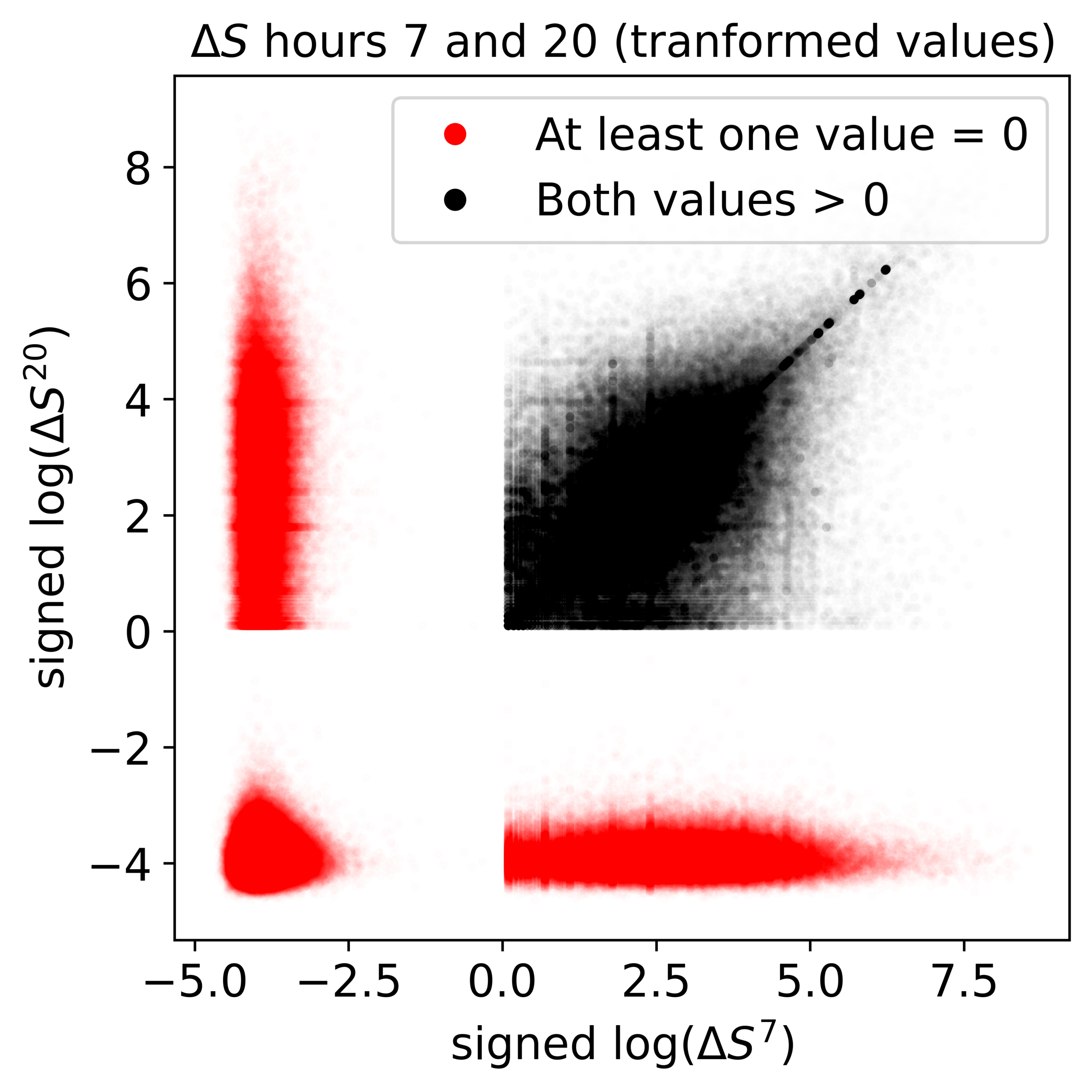}
    \caption{$\text{signed log}(x) = \text{sign}(x) \log(|x|+1)$.}    \label{fig:IMG_transformed_DVZeros}
    \end{subfigure}

    \caption{Zero\char45 valued volume increments and their randomization for hours 7 and 20.}
    \label{fig:Zeros_Randomization}
\end{figure}

Across the dataset, the zeros introduced in \eqref{eq:Deltas} create a large point mass at zero, which is difficult for diffusion models to learn directly. We therefore apply a data transformation, illustrated for hours 7 and 20. Figure \ref{fig:IMG_Original_DVZeros} shows the original zeros. We replace them with samples from small\char45 variance Gaussian distributions separated from the nonzero data. For supply, we use $\mathcal{N}(-50,12)$, obtaining the transformed data in Figure \ref{fig:IMG_transformed_DVZeros}; for demand, we use $\mathcal{N}(50,12)$. The DDPMs are then trained on the resulting signed\char45 log transformed data.

For the DDPM of the marks $\mathbf{\Delta S}$, $q_{\mathrm{data}}$ consists of $841\, 501$ observations in dimension $8$, corresponding to the vector of signed\char45 log volume increments in \eqref{eq:Deltas}; see Figures \ref{fig:Ex_DV} and \ref{fig:IMG_transformed_DVZeros}. Each observation is paired with its arrival price, and the $13$ mark features reported in the correlation table of Figure \ref{fig:corr_ext_DV}.

We train the model for $6\times 10^2$ epochs with a batch size of 256. Training time was 30 minutes. 

\begin{figure}[H]
    \centering
    \includegraphics[width=1\textwidth]{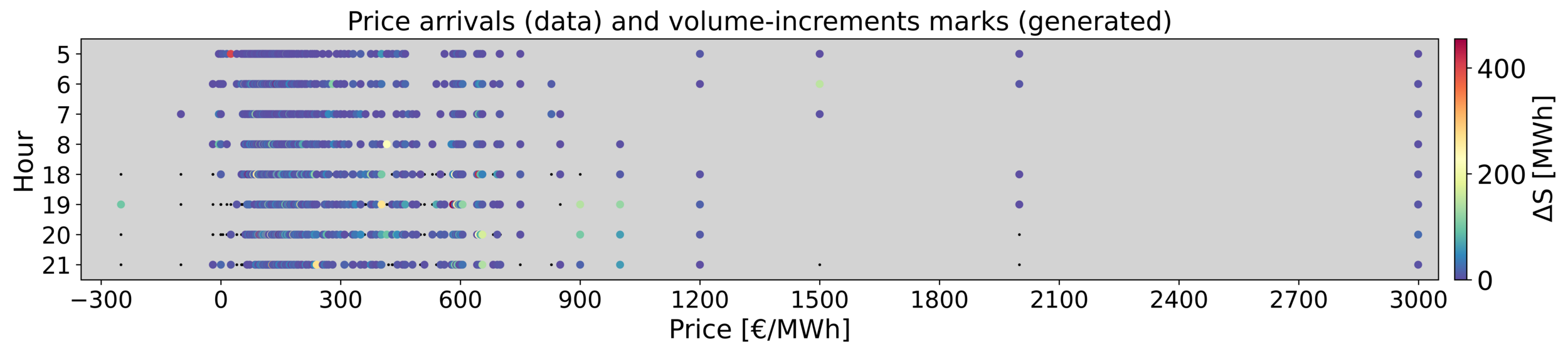}
    \caption{$(P_m, \Delta\tilde{S}_m)_{m=1}^{530}$ generated for 2023\char45 05\char45 12. Black dots represent 0\char45 values on $\Delta\tilde{S}$.}
    \label{fig:Ex_DV_Fake1DV}
\end{figure}

Figure \ref{fig:Ex_DV_Fake1DV} displays a generated sample of marks $\Delta\tilde{S}$ obtained conditional on the real price sequence for 2023\char45 05\char45 12 (shown in Figure \ref{fig:Ex_DV}). Although the generated offers do not reproduce the exact pattern observed in the real data, they preserve some cross\char45 hour structure, including block\char45 order patterns.

\subsubsection{Curve generation}

For each day $d$, the generative model produces the supply curves for all hours in $H$ jointly, using day\char45 level labels; see Figure \ref{fig:diagram_stages}:

\begin{enumerate}
\item The first DDPM generates an intensity sample $\tilde{\lambda}$ conditional on the exogenous features $\boldsymbol{X}_d^P$ for day $d$. The price arrivals $(\tilde{P}_{m})_{m=1}^{M}$ are then sampled by thinning (\cite{lewis1979simulation}). 

\item For each generated price $\tilde{P}_{m}$, the second DDPM samples its volume\char45 increment mark
\[
\Delta \tilde{S}_{m}
=
\left(\Delta \tilde{S}_{m}^h\right)_{h\in H}
\]
conditional on $\tilde{P}_{m}$ and the day\char45 level exogenous features $\boldsymbol{X}_d^S$. The cumulative supply volumes are then reconstructed recursively as
\begin{align}
\label{eq:Cummulatives}
\tilde{S}_m
=
\tilde{S}_{m-1}
+
\Delta \tilde{S}_m,
\quad
m=1,\ldots,M.
\end{align}
The collection of pairs $((\tilde{P}_{m},\tilde{S}_{m}^h)_{m=1}^{M})_{h\in H}$, defines the generated supply\char45 curve block for day $d$.
\end{enumerate}

When a generated mark $\Delta\tilde{S}_{m}^h$ contains negative components, we truncate them to zero. These values arise from the Gaussian randomization of zero volumes during training (Figure \ref{fig:Zeros_Randomization}) and are interpreted as the absence of an offer at the corresponding hour.

\begin{figure}[H]
	\centering
    \resizebox{1\textwidth}{!}{%
\begin{tikzpicture}[
	box/.style={draw, rectangle, rounded corners, align=center,
		minimum width=0.5cm, minimum height=1.2cm,
		inner sep=2pt, outer sep=0pt},
    box_text/.style={box,draw=none},
	inputBox/.style={draw, rectangle, dashed, inner sep=6pt},
	arrow/.style={-{Latex[length=3mm]}, thick},
	node distance=4mm and 8mm  
	]

	\node[box] (P_label) at (0, 0.75){Conditional feature for price \\ generation $\mathbf{X}^P_d$ for day $d$};

	\node[box, right=2cm of P_label] (Int_surr_gen) {
		Generated intensity \\ surrogate $\tilde{\lambda}$
	};
	
	\node[box, right=2cm of Int_surr_gen] (F_Prices) {Generated prices\\ $(\tilde{P}_m)_{m=1}^M$ 
	};
		
	\node[box, below = 1.2cm of F_Prices] (stage_2_input) {Conditional features for volume marks\\ generation for day $d$ $\{(\mathbf{X}^S_d, \tilde{P}_m):\, m=1,\ldots,M\}$
	};
    
    \node[box, left=2cm of stage_2_input] (F_vols) {
		Generated volume \\ marks
		$(\Delta \tilde{S}_m)_{m=1}^M$
        };
        
    \node[box, left=1.2cm of F_vols] (F_curve) {
		Generated supply \\ curve $(\tilde{P}_m,\tilde{S}_m)_{m=0}^M$
        };       
    \draw[arrow] 
(P_label) -- 
node[midway, above]{DDPM} 
node[midway, below]{for $\boldsymbol{\lambda}$} 
(Int_surr_gen);
    \draw[arrow] (Int_surr_gen) -- node[midway, below]{thinning} (F_Prices);
    \draw[arrow, dashed] (F_Prices) -- (stage_2_input);
    \draw[arrow] (stage_2_input) -- node[midway, above]{DDPM} node[midway, below]{for $\boldsymbol{\Delta S}$} (F_vols);
    \draw[arrow] (F_vols) -- node[midway, below]{\eqref{eq:Cummulatives}} (F_curve);

    \node[inputBox, fit=(P_label) (Int_surr_gen) (F_Prices)] (stage1) {};

    \node[box_text, above=-10pt of stage1] (stage1_name) {Stage 1};           

    \node[inputBox, fit=(stage_2_input) (F_vols) (F_curve) ] (stage2) {};
    
    \node[box_text, above=-10pt of stage2] (stage2_name) {Stage 2};        
	
\end{tikzpicture}}
	\caption{Two\char45 stage DDPM approach for generating supply curves}
	\label{fig:diagram_stages}
\end{figure}

Our approach generates volume increments. To focus on price\char45 dependent order\char45 level modelling, we fix the initial value in \eqref{eq:Cummulatives} from the data:
\[
\tilde{S}_0=(S^h_0)_{h\in H},
\]
since this value always corresponds to the common price level of $-300$ \EUR/MWh. This simplifies the approach but reduces generalisability, since the initial offer does not adapt to market conditions. A natural extension is to model $\tilde{S}_0$ using the first plateau values from the parametric encoding in Section \ref{Sec:Parametric encoding}. In preliminary experiments, this approach was more accurate than alternative stochastic parametrizations, such as fitting an Ornstein\char45\char45 Uhlenbeck process to historical data or learning $\tilde{S}_0$ with an additional generative model.

\begin{figure}[H]
    \centering
    \begin{subfigure}[b]{0.49\textwidth}
        \includegraphics[width=1\textwidth]{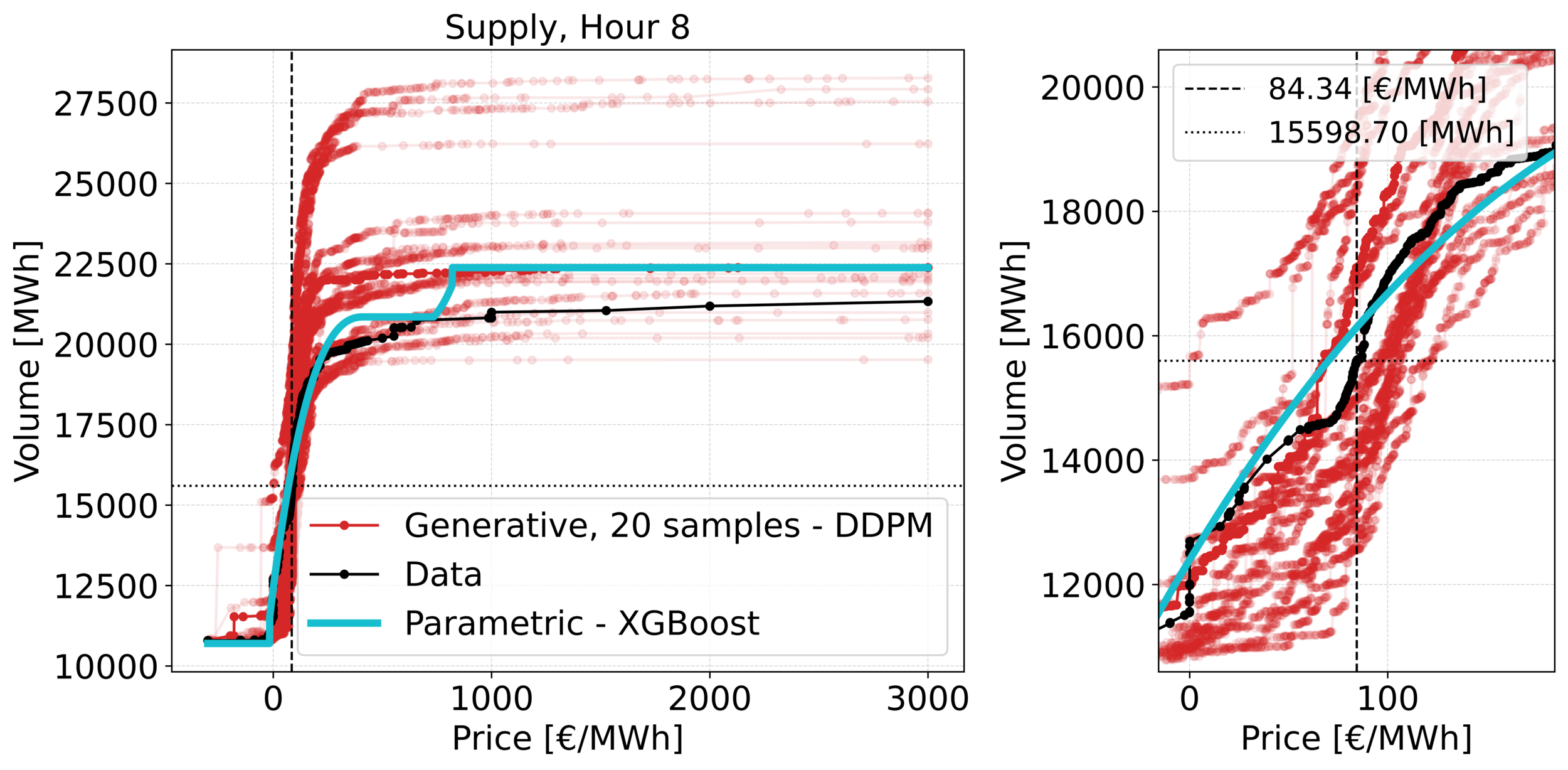}
    \end{subfigure}
    \hfill
    \begin{subfigure}[b]{0.49\textwidth}
        \includegraphics[width=1\textwidth]{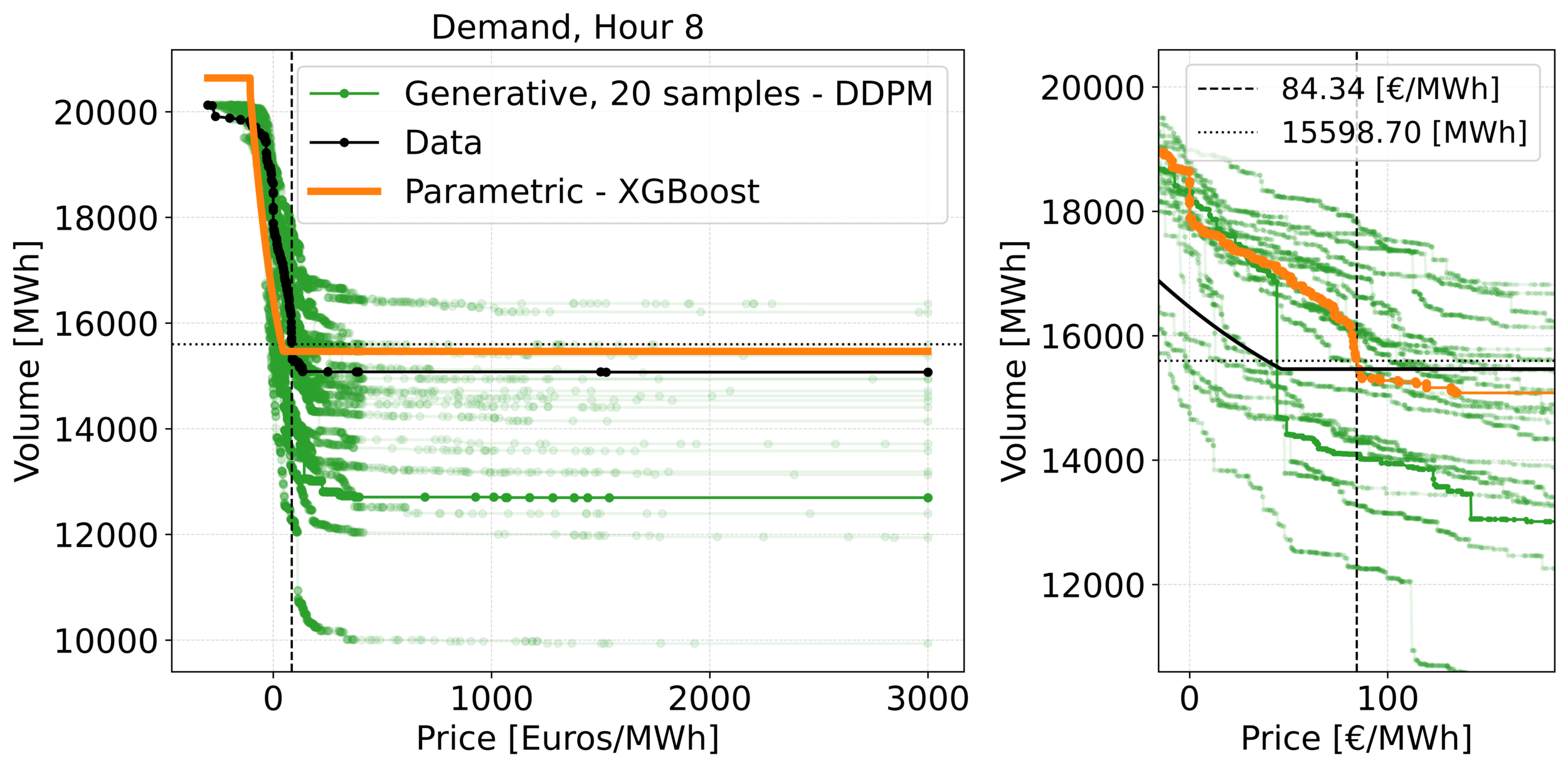}
    \end{subfigure} 

        \centering
    \begin{subfigure}[b]{0.49\textwidth}
        \includegraphics[width=1\textwidth]{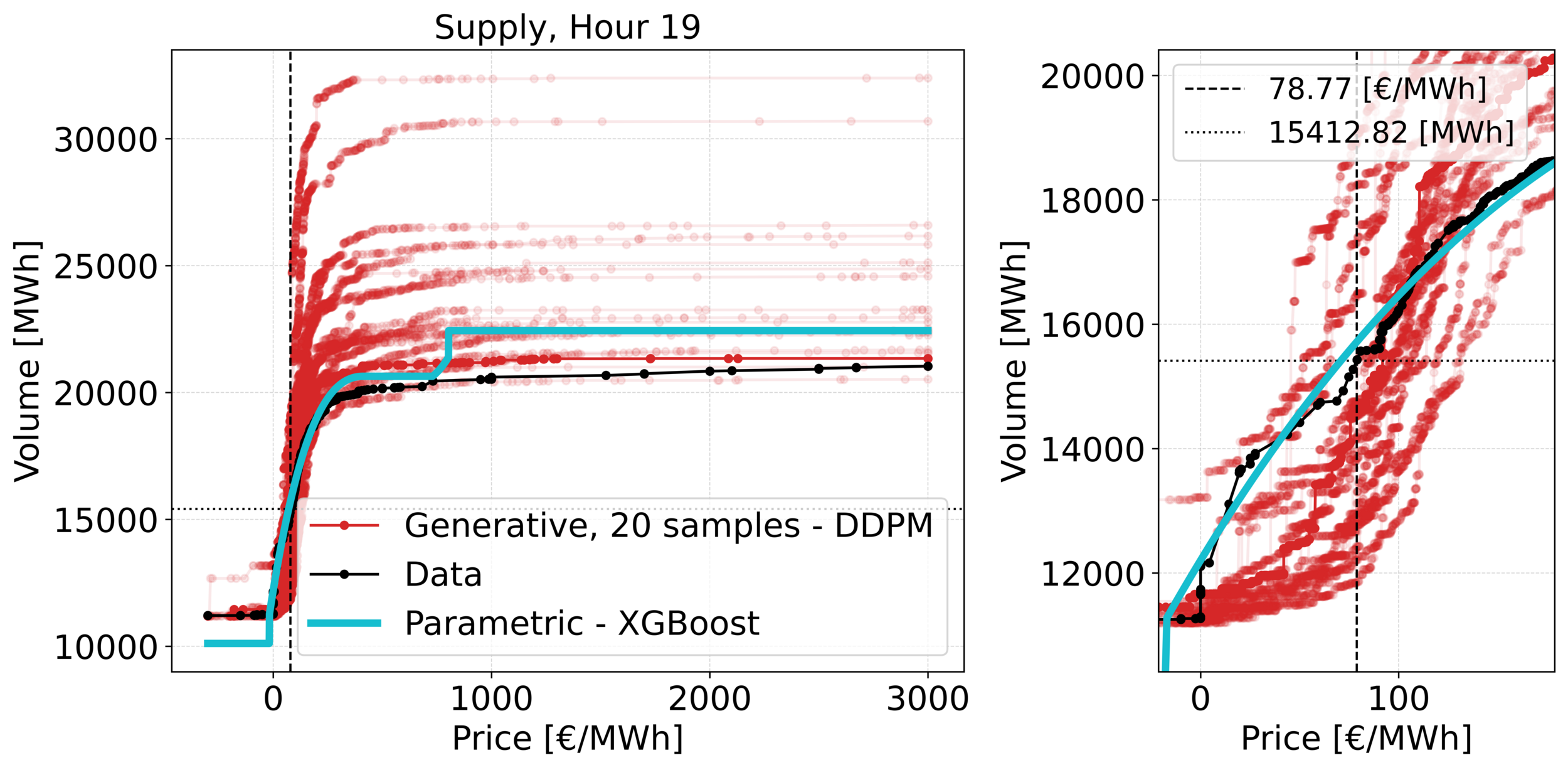}
    \end{subfigure}
    \hfill
    \begin{subfigure}[b]{0.49\textwidth}
        \includegraphics[width=1\textwidth]{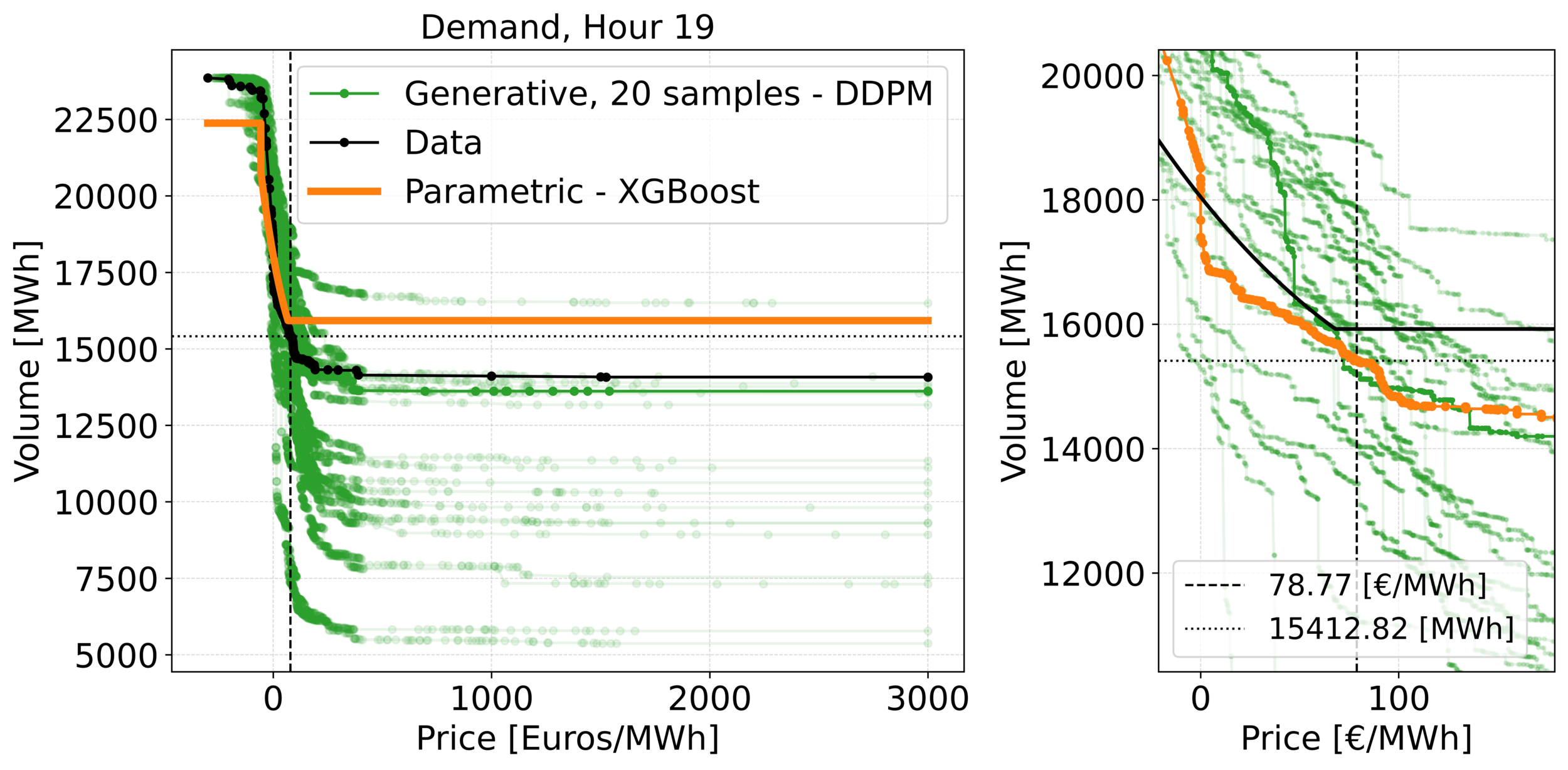}
    \end{subfigure} 

    \caption{Observed, parametric, and generative forecast curves for hours 8 and 19 on 2024\char45 10\char45 17. Supply curves are shown on the left and demand curves on the right, with zoomed views included.}
    \label{fig:Samples}
\end{figure}

Figure \ref{fig:Samples} compares the observed EPEX curve, the parametric prediction, and 200 samples from the generative model. The generated curves reproduce the transition between inelastic and elastic regimes, and the elastic region exhibits the sharp slope observed in the curves. At high price levels, the generated curves reach plateaus with varying magnitudes, sometimes below and sometimes above the observed value. In contrast, the parametric model recovers these plateau levels more accurately.

\begin{figure}[H]
    \centering
    \begin{subfigure}[b]{0.49\textwidth}
        \includegraphics[width=1\textwidth]{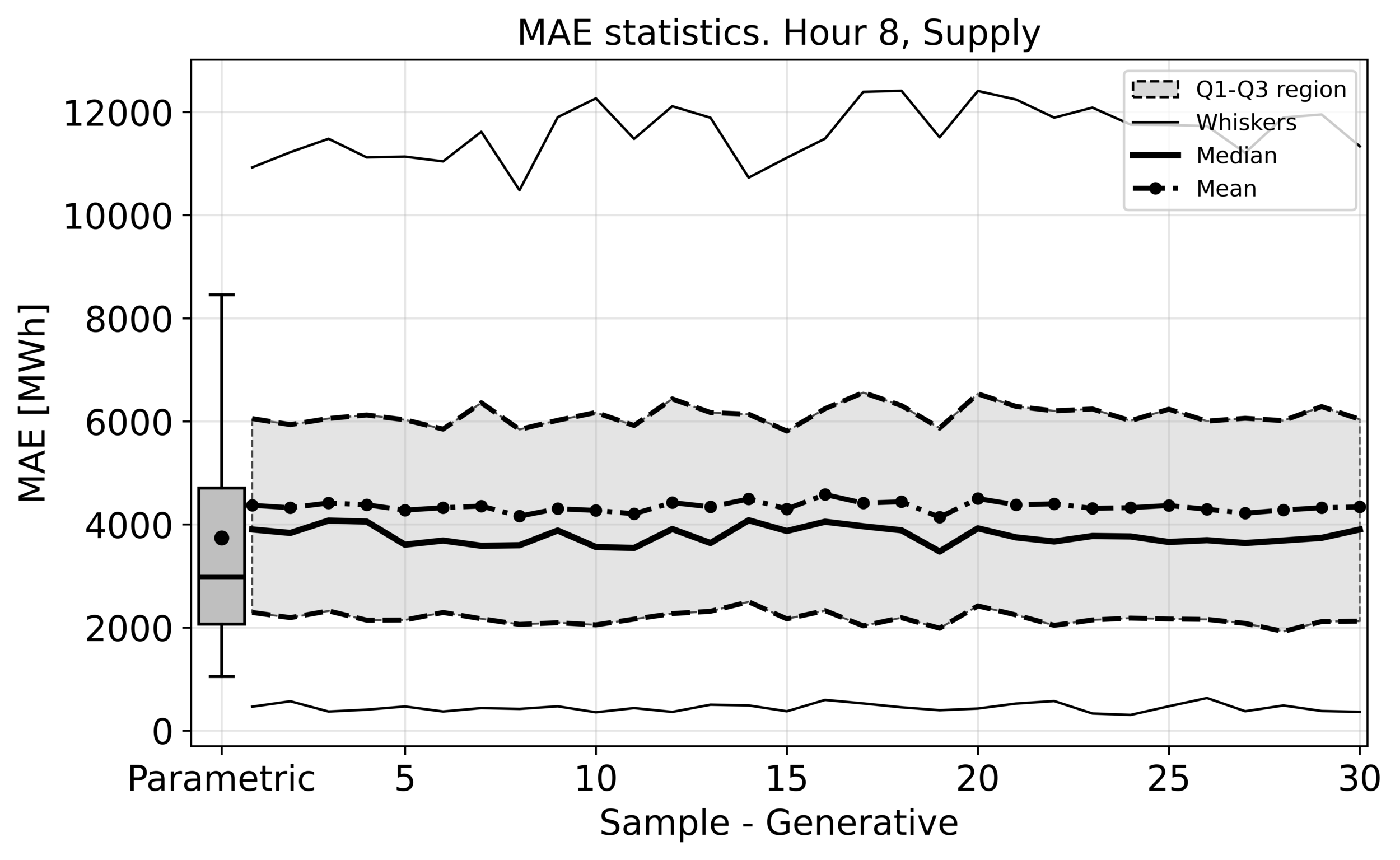}
    \end{subfigure}
    \hfill
    \begin{subfigure}[b]{0.49\textwidth}
        \includegraphics[width=1\textwidth]{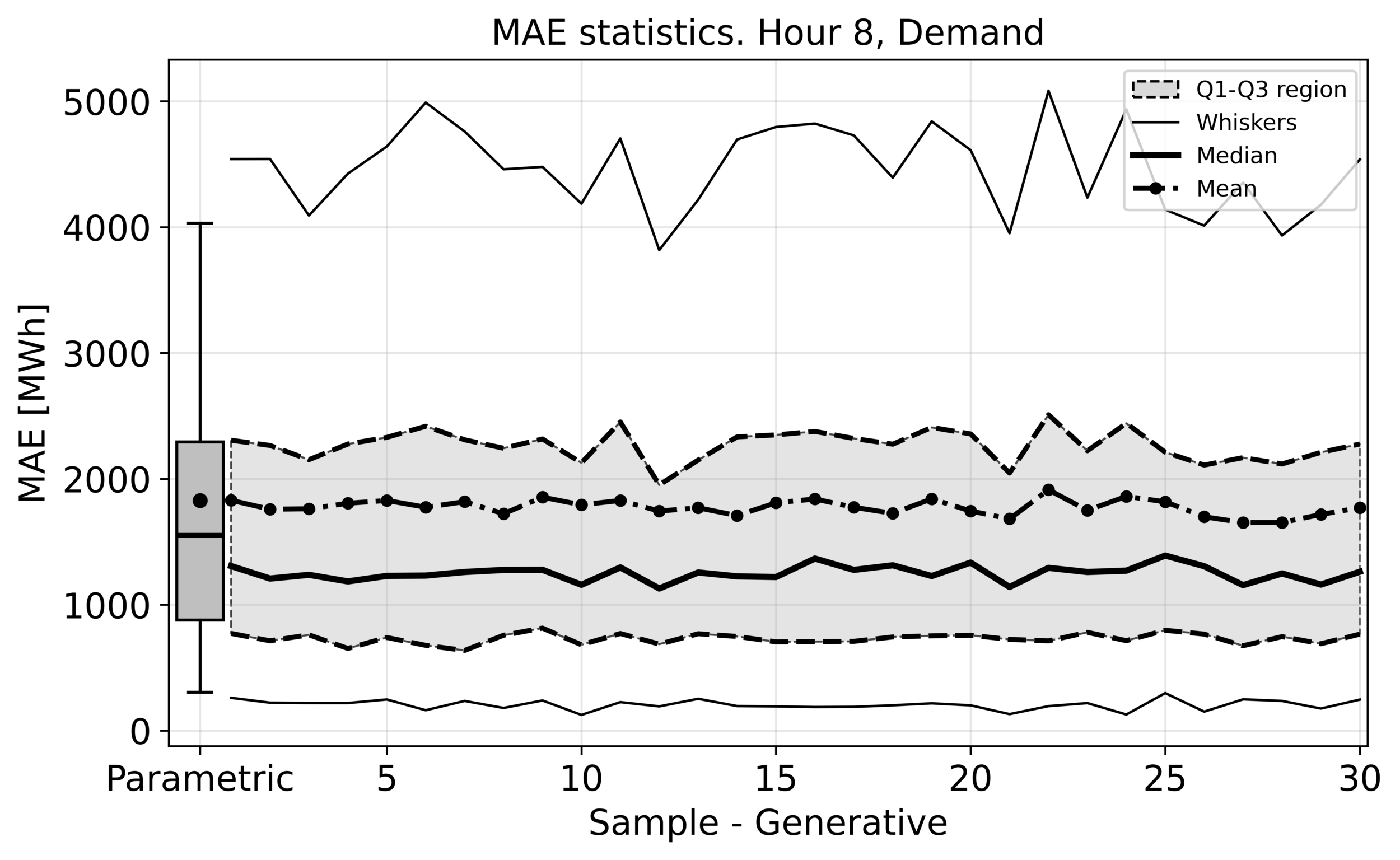}
    \end{subfigure} 

        \centering
    \begin{subfigure}[b]{0.49\textwidth}
        \includegraphics[width=1\textwidth]{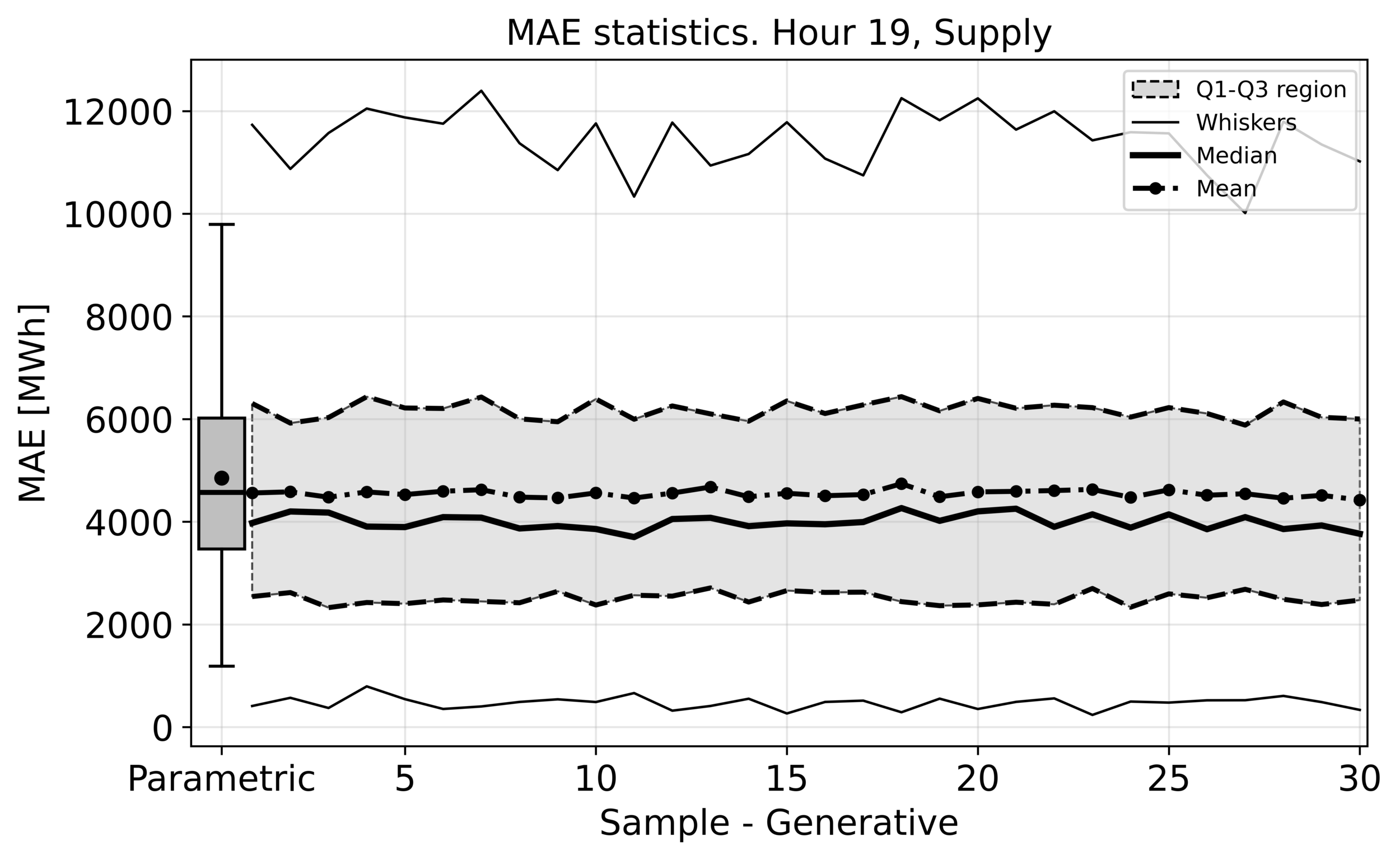}
    \end{subfigure}
    \hfill
    \begin{subfigure}[b]{0.49\textwidth}
        \includegraphics[width=1\textwidth]{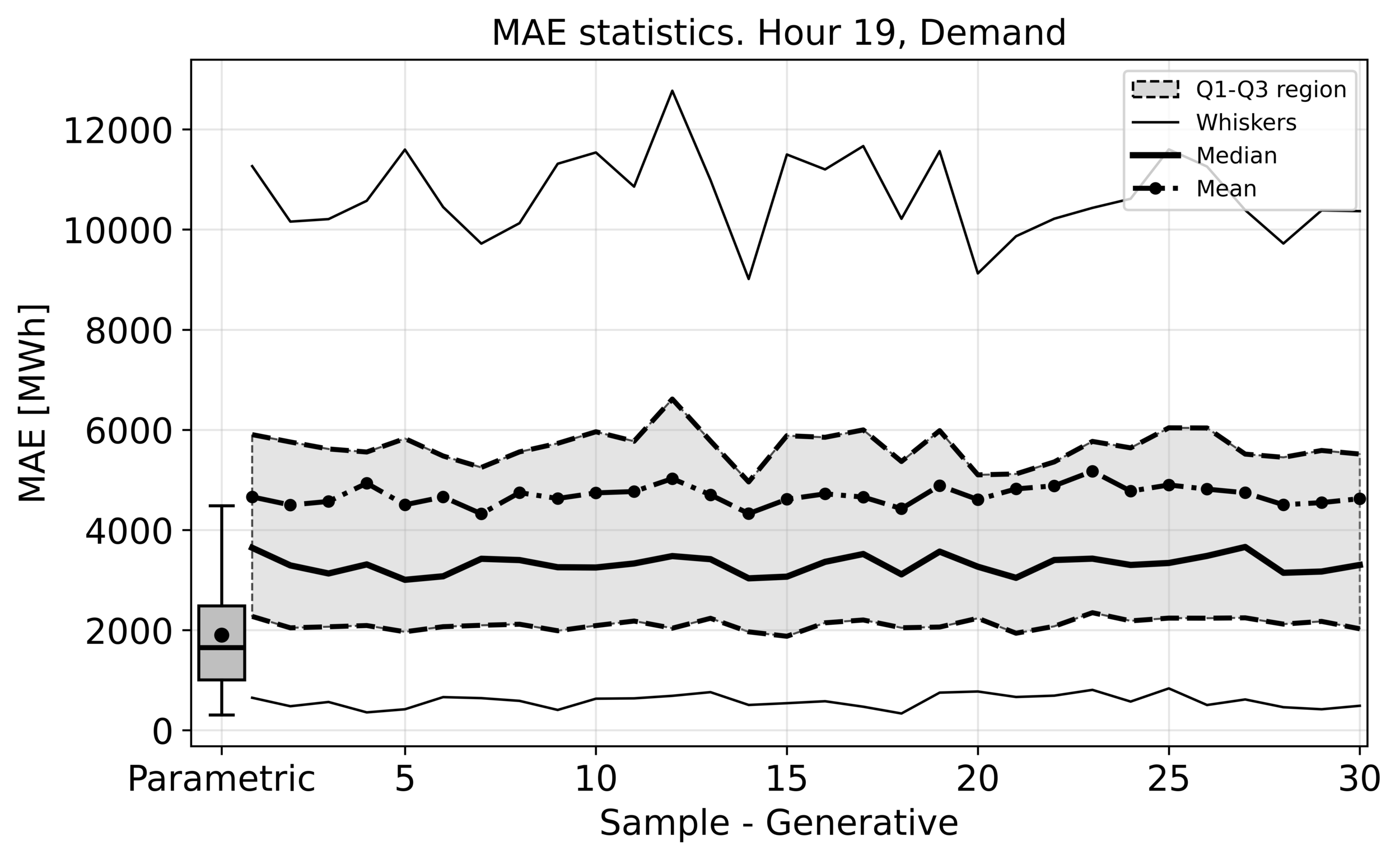}
    \end{subfigure} 

    \caption{MAE distributions against EPEX data for hours 8 and 19 on 2024\char45 10\char45 17, comparing the parametric model with 30 samples from the generative model. Supply results are shown on the left and demand results on the right.}
    \label{fig:Boxplots_samples}
\end{figure}

We evaluate the generative model on the test set by comparing generated curves with the raw observed price\char45 volume pairs from EPEX. For each of the $30$ generated samples, we compute the MAE defined in \eqref{eq:MAES parametric raw}. Figure \ref{fig:Boxplots_samples} reports boxplot statistics of the resulting MAE distributions across generated samples, showing the median, interquartile range, whiskers, and outliers. The figure also compares these distributions with the corresponding boxplots for the parametric model errors reported in Figure \ref{fig:Second_evaluation_MAE}.

\begin{figure}[H]
    \centering
    \begin{subfigure}[b]{1\textwidth}
        \includegraphics[width=1\textwidth]{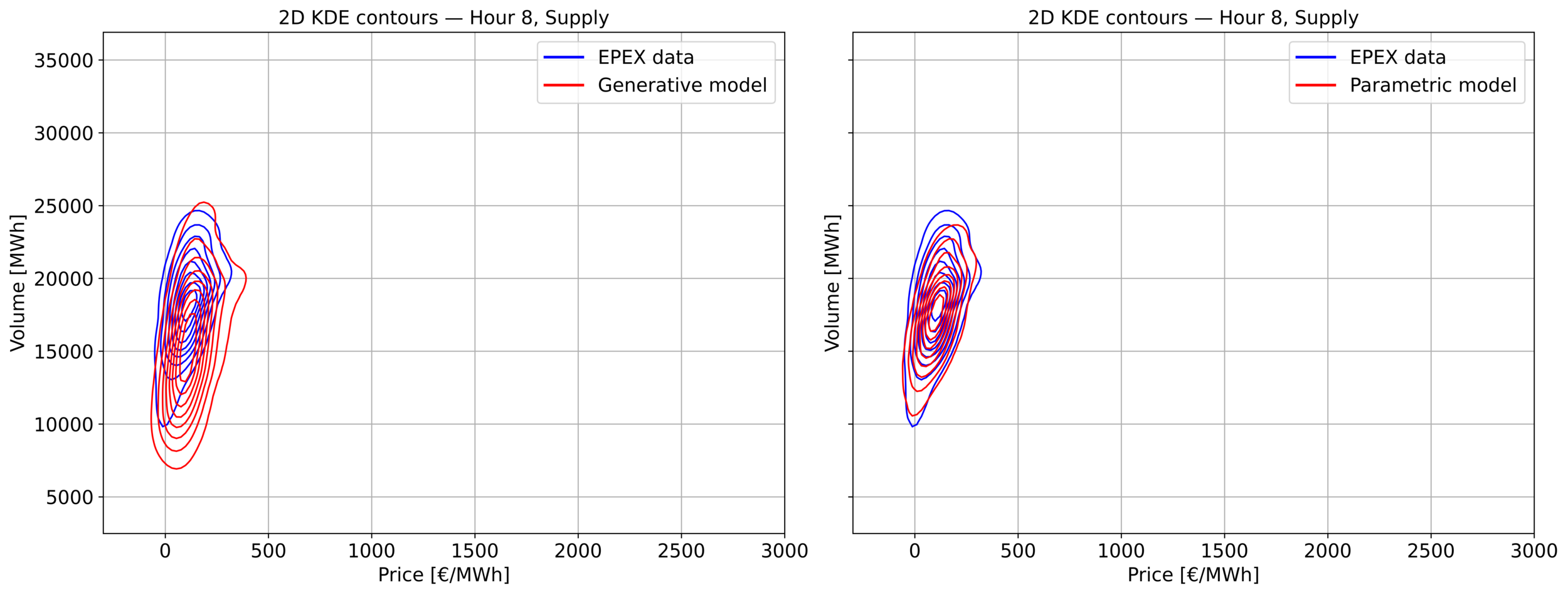}
    \end{subfigure}

    \centering
    \begin{subfigure}[b]{1\textwidth}
        \includegraphics[width=1\textwidth]{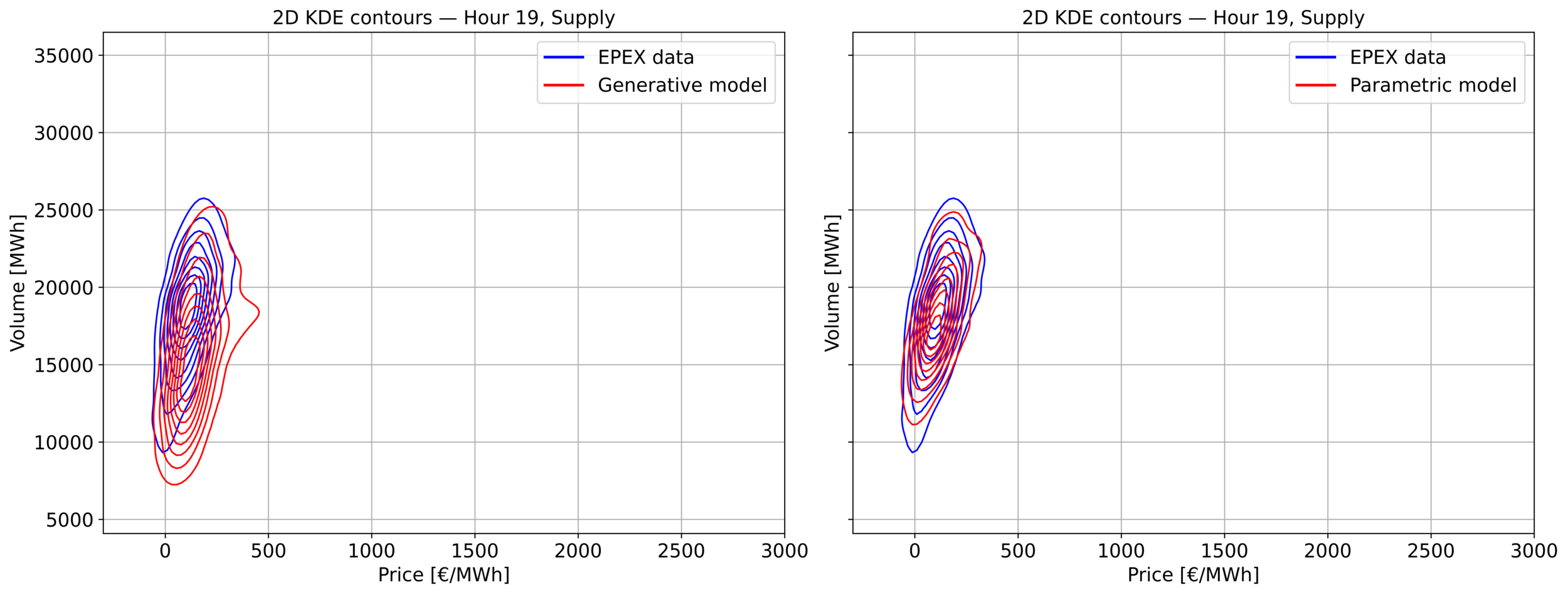}
    \end{subfigure}   
    
    \caption{KDEs of price\char45 volume supply pairs across the test dataset for EPEX data, a selected sample from the generative model, and the parametric\char45 model forecast.}
    \label{fig:KDEs_supply}
\end{figure}

\begin{figure}[H]
    \centering
    \begin{subfigure}[b]{1\textwidth}
        \includegraphics[width=1\textwidth]{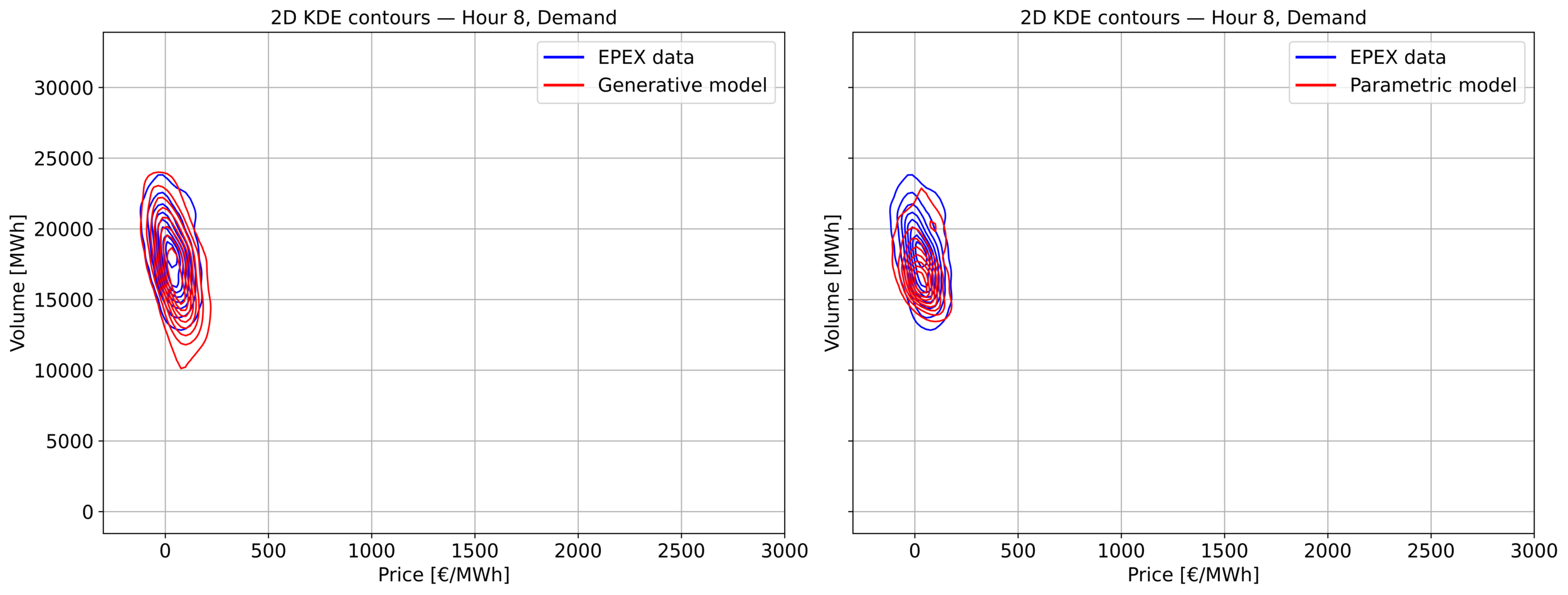}
    \end{subfigure}

    \centering
    \begin{subfigure}[b]{1\textwidth}
        \includegraphics[width=1\textwidth]{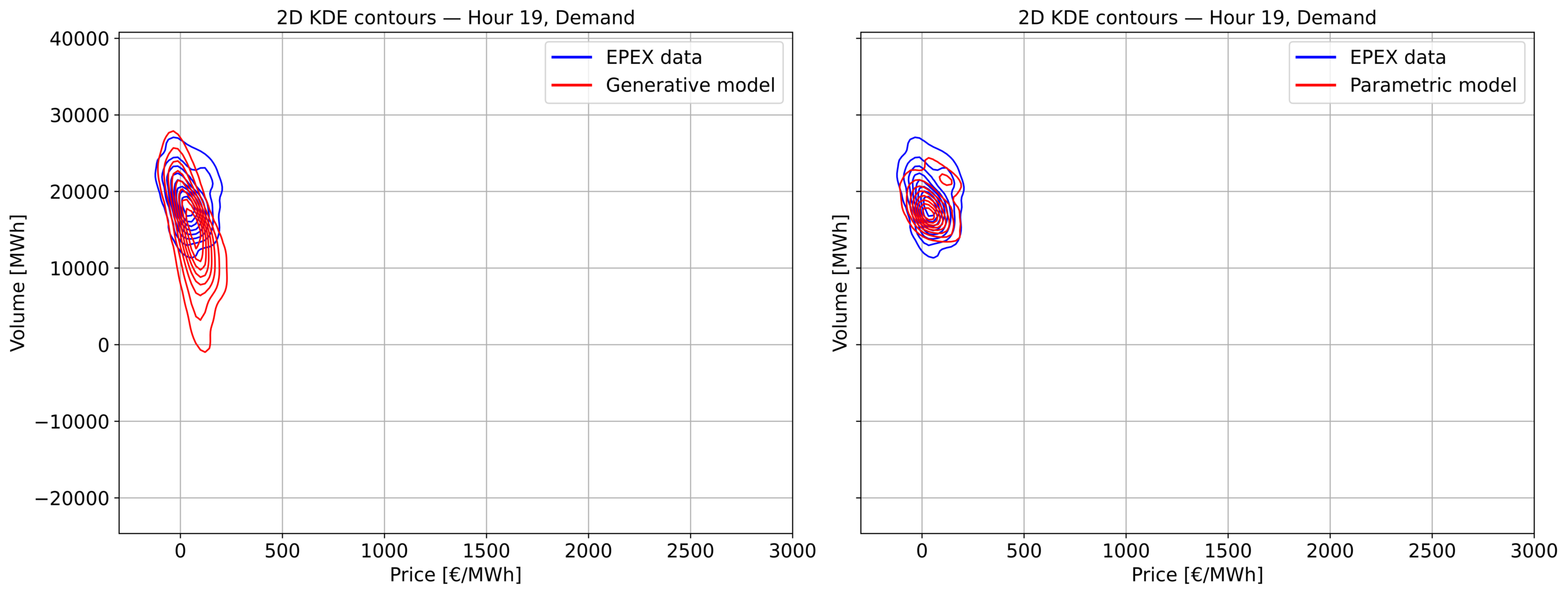}
    \end{subfigure}

    \caption{KDEs of price\char45 volume demand pairs across the test dataset for EPEX data, a selected sample from the generative model, and the parametric\char45 model forecast.}
    \label{fig:KDEs_demand}
\end{figure}

To further evaluate model performance, we compare the empirical joint distributions of price\char45 volume pairs obtained by pooling all observations for each hour across the test dataset. For a fixed hour, we collect all EPEX price\char45 volume points observed over the test dates, and compare them with the corresponding price\char45 volume pairs implied by the parametric and generative models. As a qualitative diagnostic, Figures \ref{fig:KDEs_supply} and \ref{fig:KDEs_demand} display Gaussian kernel density estimates (KDEs) of these pooled empirical distributions for the observed EPEX data, the parametric\char45 model prediction, and one selected sample from the generative model.

We complement this visual comparison with a quantitative metric: the Wasserstein distance between the corresponding empirical distributions. The Wasserstein distance compares probability distributions through optimal transport ( \cite{GalichonOTMethodsinEcono,COTFNT,peyre2025optimaltransportmachinelearners}). For empirical measures 
\[
\mu=\frac{1}{n}\sum_{i=1}^n \delta_{x_i}, \quad \nu=\frac{1}{n}\sum_{j=1}^m \delta_{y_j},
\]
induced by two point clouds $(x_i)_{i=1}^n$ and $(y_i)_{i=1}^m$, the Wasserstein distance is 
\[
W_2^2(\mu,\nu) = \min_{\pi \in \Pi(\mu,\nu)} \sum_{i=1}^{n}\sum_{j=1}^m \pi_{ij} \|x_i - y_j\|^2,
\]
where $\Pi(\mu,\nu)$ is the set of transport plans with marginals $\mu$ and $\nu$. 

We describe the computation for the supply side; the demand side is treated analogously. For each date $d$ and hour $h$, the observed EPEX supply curve provides price\char45 volume points 
\[
( (P_{d,h,\ell}^{\mathrm{EPEX}},
S_{d,h,\ell}^{\mathrm{EPEX}}))_{\ell=1}^{N_{d,h}}.
\]
Each sample $k = 1 ,\ldots, K$ from the generative model produces price\char45 volume point clouds. Hence, we use the generated points $((P_{d,h,k,j}^{\mathrm{gen}},
S_{d,h,k,j}^{\mathrm{gen}}))_{j=1}^{M_{d,h,k}}$, and compute the Wasserstein distance between 
\[ 
\mu_{d,h}^{\mathrm{EPEX}}
=
\frac{1}{N_{d,h}}
\sum_{\ell=1}^{N_{d,h}}
\delta_{\left(P_{d,h,\ell}^{\mathrm{EPEX}},
S_{d,h,\ell}^{\mathrm{EPEX}}\right)} \quad \text{and} \quad
\mu_{d,h,k}^{\mathrm{gen}}
=
\frac{1}{M_{d,h,k}}
\sum_{j=1}^{M_{d,h,k}}
\delta_{\left(P_{d,h,k,j}^{\mathrm{gen}},
S_{d,h,k,j}^{\mathrm{gen}}\right)}.
\]
Since the Wasserstein distance has no intrinsic threshold for determining whether two empirical distributions are sufficiently close, we interpret the generative distances relative to the parametric model. This model does not directly provide price\char45 volume point clouds. Thus, we evaluate it at the observed EPEX prices, obtaining $\widehat S_{d,h,\ell}^{\mathrm{par}}
=
\widehat S_{d,h}^{\mathrm{par}}
\left(P_{d,h,\ell}^{\mathrm{EPEX}}\right)$, and compute $W_2^2(
\mu_{d,h}^{\mathrm{EPEX}},
\mu_{d,h}^{\mathrm{par}}
)$ for 
\[
\mu_{d,h}^{\mathrm{par}}
=
\frac{1}{N_{d,h}}
\sum_{\ell=1}^{N_{d,h}}
\delta_{\left(P_{d,h,\ell}^{\mathrm{EPEX}},
\widehat S_{d,h,\ell}^{\mathrm{par}}\right)}.
\]
It is worth noting that because the parametric model is evaluated on the observed EPEX price grid, this removes price\char45 support errors from its Wasserstein comparison and therefore places it in a more favorable position for this metric than the generative model, which is evaluated using its own generated price\char45 volume points. 

\begin{figure}[H]
    \centering
    \begin{subfigure}[b]{1\textwidth}
        \includegraphics[width=1\textwidth]{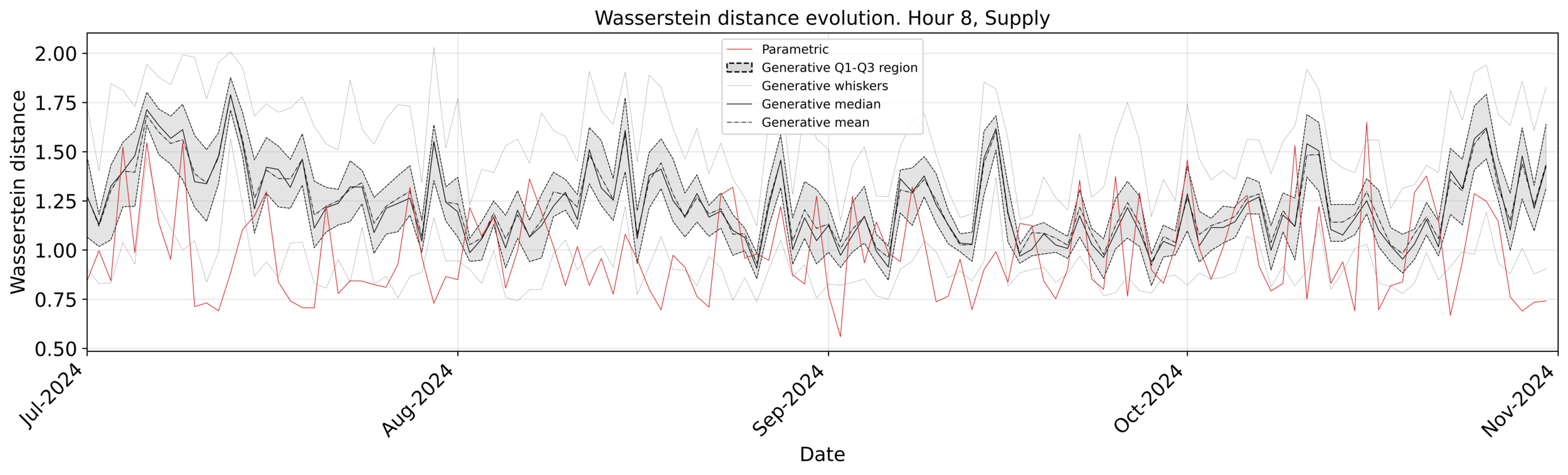}
    \end{subfigure}
 
        \centering
    \begin{subfigure}[b]{1\textwidth}
        \includegraphics[width=1\textwidth]{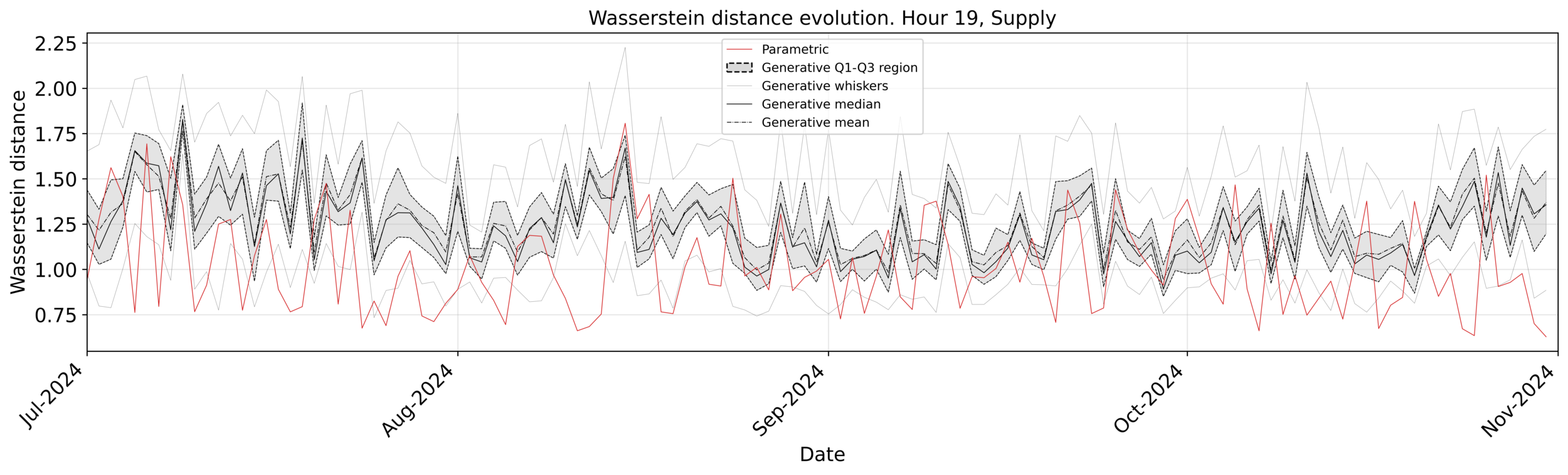}
    \end{subfigure}
      
        \centering
    \begin{subfigure}[b]{1\textwidth}
        \includegraphics[width=1\textwidth]{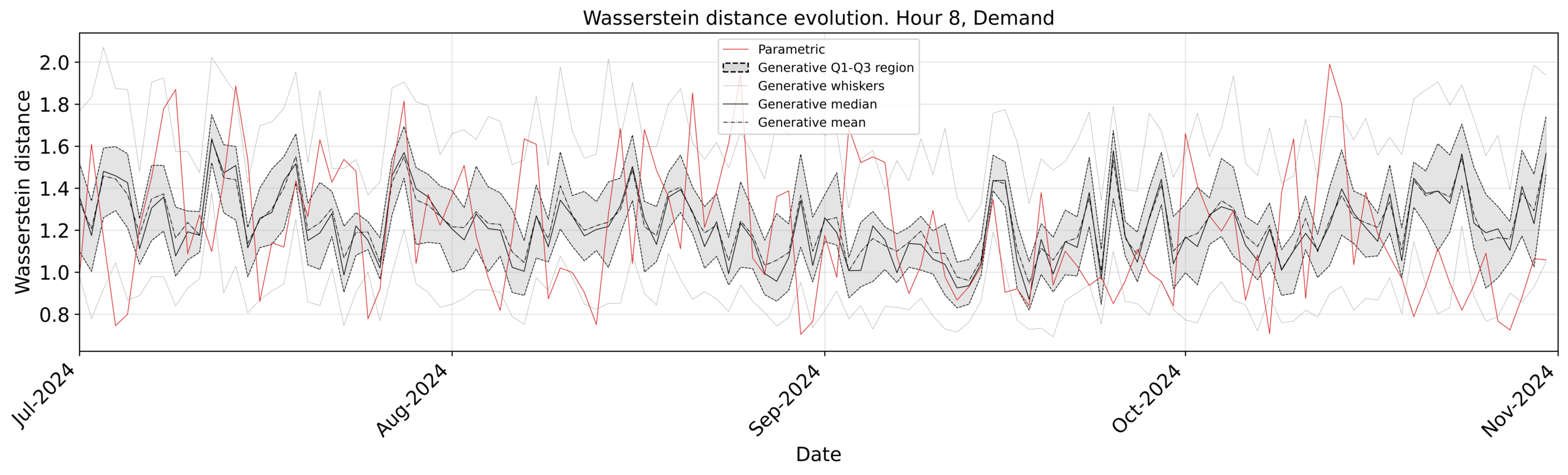}
    \end{subfigure}    

        \centering
    \begin{subfigure}[b]{1\textwidth}
        \includegraphics[width=1\textwidth]{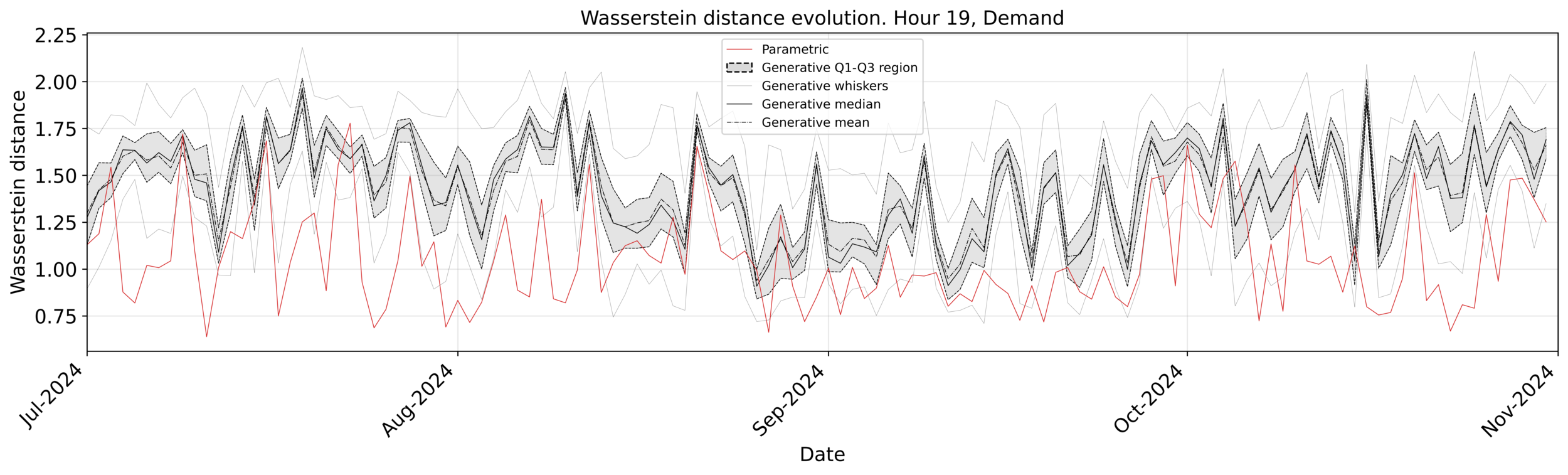}
    \end{subfigure}

    \caption{Evolution of the $2$\char45 Wasserstein distance for hours 8 and 19 from 2024\char45 07\char45 01 to 2024\char45 10\char45 31, comparing the parametric model distance with summary statistics computed over 30 samples from the generative model.}
    \label{fig:Wasserstein_samples}
\end{figure}

Figure \ref{fig:Wasserstein_samples} shows the evolution of the Wasserstein distances from July to October 2024. For the generative model, we compute one Wasserstein distance for each of the $K=30$ generated samples at every date and hour, and report the resulting boxplot statistics across samples. The corresponding parametric Wasserstein distance is shown for comparison.

The previous visual and distributional diagnostics indicate that the DDPM implementation captures important aspects of the elastic\char45 region structure, producing curve shapes that resemble those observed in the data. Its main limitation is that cumulative errors propagate into the plateau regions, leading to deviations in the estimated upper\char45 plateau values. By contrast, the XGBoost implementation, which relies on a more data\char45 intensive set of input variables (see Table \ref{tab:parametric_features_summary}), provides an accurate approximation of both the elastic and plateau regions.

This comparison suggests that the two approaches capture complementary aspects of the EPEX curves. The DDPM is particularly effective at learning the elastic part of the distribution with substantially fewer input variables, whereas the parametric model benefits from extensive conditioning information to recover the plateau levels. This distinction is important: when input variables are used as conditioning variables for generation, a richer conditioning structure effectively segments the data more finely. As a result, more observations are required to reliably estimate the corresponding conditional distributions; otherwise, each conditional distribution is approximated using only a limited number of data points.

Figure \ref{fig:Boxplots_samples} further shows that the DDPM implementation is consistent across generated samples. The samples exhibit the same overall tendency while retaining a moderate degree of variability, which is desirable for a generative model. Relative to the XGBoost benchmark, the DDPM results are comparable for hour 8 on the demand side and hour 19 on the supply side, while performance deteriorates for hour 19 on the demand side. This pattern is also reflected in the Wasserstein\char45 distance evolution reported in Figure \ref{fig:Wasserstein_samples}, where the relative performance of the two models varies across dates, with the weakest DDPM performance again observed for demand at hour 19.

The previous section evaluates whether the predicted curves resemble the observed EPEX curves. We now shift the focus from reconstruction accuracy to decision quality by asking whether the forecasts lead to effective decisions for a price\char45 maker storage agent.

\section{Decision\char45 based evaluation: price\char45 maker storage}
\label{Sec:Application Optimal strategy of a price maker storage agent}

Although forecasts are available for both supply and demand curves, we treat demand as inelastic and represent it solely through a forecasted market\char45 clearing quantity $D_h^{\mathrm m}$. This simplification is motivated by the fact that in day\char45 ahead electricity markets, short\char45 term demand is typically much less responsive to price than supply is. Moreover, the objective of the storage agent is to evaluate the impact of its actions on market prices, as reflected in the net quantity cleared against the supply curve. 

Because the storage agent is assumed to be sufficiently large to affect the market\char45 clearing price, we distinguish between $P_h$, the market price without storage intervention, and $P^{\mathrm{i}}_h$, the market price resulting from the storage agent's actions. For each hour $h \in H$, we assume the storage agent has access to:
\begin{enumerate}
    \item A forecast $D^\mathrm{m}_h$ of the aggregated demand volume to be cleared in the market.  
    \item A forecast of the supply curve $P \mapsto S_h(P)$, assumed monotone, so that a generalized inverse $S_h^{-1}$ can be used to recover the clearing price. The forecast may be obtained from either the parametric or the generative approaches developed previously.
\end{enumerate}

We consider a block optimization problem where the storage agent chooses an initial state of charge $Q_0$ and a sequence of hourly storage actions
\[
(q_h)_{h\in H}.
\]
We use the convention that $q_h>0$ corresponds to charging, that is, an increase in the quantity demanded from the market, while $q_h<0$ corresponds to discharging. At hour $h \in H$, the market\char45 clearing price is defined through the following equilibrium equation 
\[
S_h(P_h^{\mathrm i})=D_h^{\mathrm m}+q_h,
\]
which gives $P_h^{\mathrm i}(q_h)=S_h^{-1}\left(D_h^{\mathrm m}+q_h\right)$. Hence, the block optimization problem is
\[
\begin{aligned}
& \min_{Q_0,\,(q_h)_{h\in H}}
\quad
\sum_{h\in H} q_h
S_h^{-1}\left(D_h^{\mathrm m}+q_h\right)
\\
& \text{subject to} \begin{cases}
 -\bar q \le q_h \le \bar q,
\quad Q_h = Q_0+\sum_{\tilde{h} \leq h} q_{\tilde h},
\quad 0\le Q_h\le Q_{\max},
\qquad h\in H,
\\
0\le Q_0\le Q_{\max}, \quad \sum_{h\in H} q_h=0.
\end{cases}
\end{aligned}
\]
The objective represents the total net cost of the storage schedule. Therefore, the corresponding arbitrage profit is the negative of the objective. The constraint $-\bar q\le q_h\le \bar q$ imposes hourly charging and discharging limits. The variable $Q_h$ is the state of charge after hour $h$, and the constraint $0\le Q_h\le Q_{\max}$ enforces the battery capacity limits. Finally, $\sum_{h \in H} q_h=0$ imposes a cyclic condition: the battery finishes the block with the same state of charge with which it started.

When the supply curves are generated by a stochastic forecasting model, the stochastic block optimization problem becomes
\begin{align}
\label{eq:Sto bloc problem}
\begin{aligned}
& \min_{Q_0,\,(q_h)_{h\in H}}
\quad
\mathbb E\left[
\sum_{h \in H} q_h
S_h^{-1}\left(D_h^{\mathrm m}+q_h,\omega\right)
\right]
\\
& \text{subject to} \begin{cases}
-\bar q \le q_h \le \bar q, \quad Q_h = Q_0+\sum_{\tilde h\leq h} q_{\tilde h}, \quad 0\le Q_h\le Q_{\max},
\qquad h\in H,
\\
0\le Q_0\le Q_{\max}, \quad \sum_{h\in H} q_h=0.
\end{cases}
\end{aligned}
\end{align}
Given $K$ generated supply\char45 curve scenarios
\[
(S_{h,1},\ldots,S_{h,K}),
\qquad h\in H,
\]
the sample\char45 average approximation of \eqref{eq:Sto bloc problem} is
\[
\begin{aligned}
& \min_{Q_0,\,(q_h)_{h\in H}}
\quad
\frac1K
\sum_{k=1}^K
\sum_{h\in H}
q_h
S_{h,k}^{-1}\left(D_h^{\mathrm m}+q_h\right)
\\
& \text{subject to} \begin{cases}
-\bar q \le q_h \le \bar q,
\quad Q_h = Q_0+\sum_{\tilde h\leq h} q_{\tilde h}, \quad 0\le Q_h\le Q_{\max},
\qquad h\in H,
\\
0\le Q_0\le Q_{\max}, \quad \sum_{h\in H} q_h=0.
\end{cases}
\end{aligned}
\]

For the numerical implementation, we solve the previous problem independently for each delivery date $d$ in the test set. The observed market\char45 clearing volume is used as the demand input,
\[
D_{d,h}^{\mathrm m}:=D_{d,h}^{\mathrm{real}},
\]
so that the experiment evaluates the quality of the predicted supply curves conditional on the realized cleared demand volume. For the parametric model, let $\widehat S_{d,h}^{\mathrm{par}}$ denote the predicted supply curve. The parametric strategy is computed as
\[
(Q_{0,d}^{\mathrm{par}},(q_{d,h}^{\mathrm{par}})_{h \in H})
\in
\arg\min_{Q_0,(q_h)_{h\in H}}
\sum_{h\in H}
q_h
\left(\widehat S_{d,h}^{\mathrm{par}}\right)^{-1}
\left(D_{d,h}^{\mathrm m}+q_h\right),
\]
subject to the same storage constraints. Its predicted profit is
\[
\widehat \Pi_{d}^{\mathrm{par}}
=
-
\sum_{h\in H}
q_{d,h}^{\mathrm{par}}
\left(\widehat S_{d,h}^{\mathrm{par}}\right)^{-1}
\left(D_{d,h}^{\mathrm m}+q_{d,h}^{\mathrm{par}}\right).
\]

For the generative model, let $\widehat S_{d,h,k}^{\mathrm{gen}}$ denote the $k$th generated supply curve for date $d$ and hour $h$, with $k=1,\ldots,K$. We compute one common strategy for the date by minimizing the Monte Carlo average cost,
\[
(Q_{0,d}^{\mathrm{gen}},(q_{d,h}^{\mathrm{gen}})_{h\in H})
\in
\arg\min_{Q_0,(q_h)_{h\in H}}
\frac{1}{K}
\sum_{k=1}^K
\sum_{h\in H}
q_h
\left(\widehat S_{d,h,k}^{\mathrm{gen}}\right)^{-1}
\left(D_{d,h}^{\mathrm m}+q_h\right),
\]
again subject to the same storage constraints. The corresponding predicted generative profit is
\[
\widehat \Pi_{d}^{\mathrm{gen}}
=
-
\frac{1}{K}
\sum_{k=1}^K
\sum_{h\in H}
q_{d,h}^{\mathrm{gen}}
\left(\widehat S_{d,h,k}^{\mathrm{gen}}\right)^{-1}
\left(D_{d,h}^{\mathrm m}+q_{d,h}^{\mathrm{gen}}\right).
\]

After the strategies are computed from the predicted curves, they are evaluated out of sample on the realized EPEX supply curves. Let $S_{d,h}^{\mathrm{real}}$ be the realized sell curve for date $d$ and hour $h$. The realized profit of the parametric strategy is
\[
\Pi_{d}^{\mathrm{real,par}}
=
-
\sum_{h\in H}
q_{d,h}^{\mathrm{par}}
\left(S_{d,h}^{\mathrm{real}}\right)^{-1}
\left(D_{d,h}^{\mathrm m}+q_{d,h}^{\mathrm{par}}\right),
\]
while the realized profit of the generative strategy is
\[
\Pi_{d}^{\mathrm{real,gen}}
=
-
\sum_{h\in H}
q_{d,h}^{\mathrm{gen}}
\left(S_{d,h}^{\mathrm{real}}\right)^{-1}
\left(D_{d,h}^{\mathrm m}+q_{d,h}^{\mathrm{gen}}\right).
\]
This separates the optimization stage, which uses forecasted curves, from the evaluation stage, which uses realized curves. As a benchmark, we also compute an oracle strategy using the realized supply curves directly:
\[
(Q_{0,d}^{\mathrm{or}},(q_{d,h}^{\mathrm{or}})_{h\in H})
\in
\arg\min_{Q_0,(q_h)_{h\in H}}
\sum_{h\in H}
q_h
\left(S_{d,h}^{\mathrm{real}}\right)^{-1}
\left(D_{d,h}^{\mathrm m}+q_h\right),
\]
subject to the same storage constraints. The oracle profit is therefore
\[
\Pi_{d}^{\mathrm{or}}
=
-
\sum_{h\in H}
q_{d,h}^{\mathrm{or}}
\left(S_{d,h}^{\mathrm{real}}\right)^{-1}
\left(D_{d,h}^{\mathrm m}+q_{d,h}^{\mathrm{or}}\right).
\]
The final comparison is based on the realized performance of the strategies derived from the parametric and generative forecasts, measured relative to the oracle strategy. For each forecast type $m\in\{\mathrm{par},\mathrm{gen}\}$, we define the profit gap
\[
\Delta_{d}^{m}
=
\Pi_{d}^{\mathrm{or}}
-
\Pi_{d}^{\mathrm{real},m}.
\]
A smaller value of $\Delta_d^m$ indicates that the strategy derived from the forecasted curves is closer to the oracle strategy. The comparison is performed across dates using the mean and median realized profits, the mean and median profit gaps, and the share of dates on which each forecast\char45 based strategy obtains the larger realized profit.

Because the storage problem is nonlinear, the numerical comparison may be affected by local optima. To keep the comparison consistent, the parametric, generative, and oracle strategies are computed using the same numerical protocol\footnote{The numerical problem is implemented in Python using \texttt{scipy.optimize.minimize} with the Sequential Least Squares Programming (\texttt{SLSQP}) method, a tolerance $\texttt{ftol}=10^{-9}$, and a maximum number of iterations $\texttt{maxiter}=1000$.}: the same constrained nonlinear solver, feasibility constraints, tolerances, and initialization procedure. Thus, differences in realized performance are not driven by differences in the optimization setup.

In the numerical implementation, we solve the storage problem on the block $H$ defined in \eqref{eq:Selected hours}. The storage parameters\footnote{The parameters correspond to a four\char45 hour battery, meaning that the maximum energy capacity is four times the maximum charging or discharging power: $Q_{\max}/\bar q=2000/500=4$ hours. Although battery revenues are often mainly associated with ancillary services, larger batteries of this type are increasingly relevant for spot\char45 market arbitrage.} are reported in Table \ref{tab:storage_optimization_parameters}. For the generative forecast, the stochastic objective is approximated by a sample average over $K=30$ generated supply curves per date.
\begin{table}[H]
\centering
\begin{tabular}{|c|p{9cm}|c|}
\hline
Parameter & \multicolumn{1}{c|}{Description} & Value \\
\hline

$\bar q$
&
Maximum hourly charging and discharging power: $-\bar q \le q_h \le \bar q$.
&
$500$ [MW] \\
\hline

$Q_{\max}$
&
Maximum energy capacity: $0\le Q_h\le Q_{\max}$.
&
$2000$ [MWh]\\
\hline

$Q_0^{\mathrm{init}}$
&
Initial value used by the numerical solver: $Q_{\max}/2$.
&
$1000$ [MWh] \\
\hline

$q_h^{\mathrm{init}}$
&
Initial hourly storage action used by the numerical solver.
&
$0$ [MW] \\
\hline

\end{tabular}
\caption{Numerical parameters used in the storage optimization problem.}
\label{tab:storage_optimization_parameters}
\end{table}

Table \ref{tab:storage_model_comparison} reports the out\char45 of\char45 sample performance of the two forecast\char45 based strategies. The strategy derived from the generative forecast achieves a higher mean realized profit, $17\,042.8$ \EUR, compared with $8\,620.7$ \EUR{} for the strategy derived from the parametric forecast, and also obtains a higher median realized profit. Relative to the oracle strategy, the generative\char45 based strategy has smaller mean and median profit gaps. It also obtains a higher realized profit on $52.9\%$ of the dates, compared with $47.1\%$ for the parametric\char45 based strategy.

\begin{table}[H]
\centering
\begin{tabular}{|p{6.5cm}|r|r|}
\hline
\multicolumn{1}{|c|}{Metric} & \multicolumn{1}{c|}{Parametric} & \multicolumn{1}{c|}{Generative} \\
\hline

Mean realized profit
&
$8\,620.7$ \EUR
&
$17\,042.8$ \EUR \\
\hline

Median realized profit
&
$10\,299.0$ \EUR
&
$12\,437.6$ \EUR \\
\hline

Standard deviation of realized profit
&
$45\,477.7$ \EUR
&
$42\,445.8$ \EUR \\
\hline

Mean profit gap to oracle
&
$38\,677.3$ \EUR
&
$30\,255.2$ \EUR \\
\hline

Median profit gap to oracle
&
$26\,226.4$ \EUR
&
$20\,519.5$ \EUR \\
\hline
Share of dates with higher realized profit
&
$47.1\%$
&
$52.9\%$ \\
\hline

\end{tabular}
\caption{Out\char45 of\char45 sample storage performance of the parametric and DDPM models relative to the oracle benchmark.}
\label{tab:storage_model_comparison}
\end{table}

\begin{figure}[H]
    \centering        
    \includegraphics[width=1\textwidth]{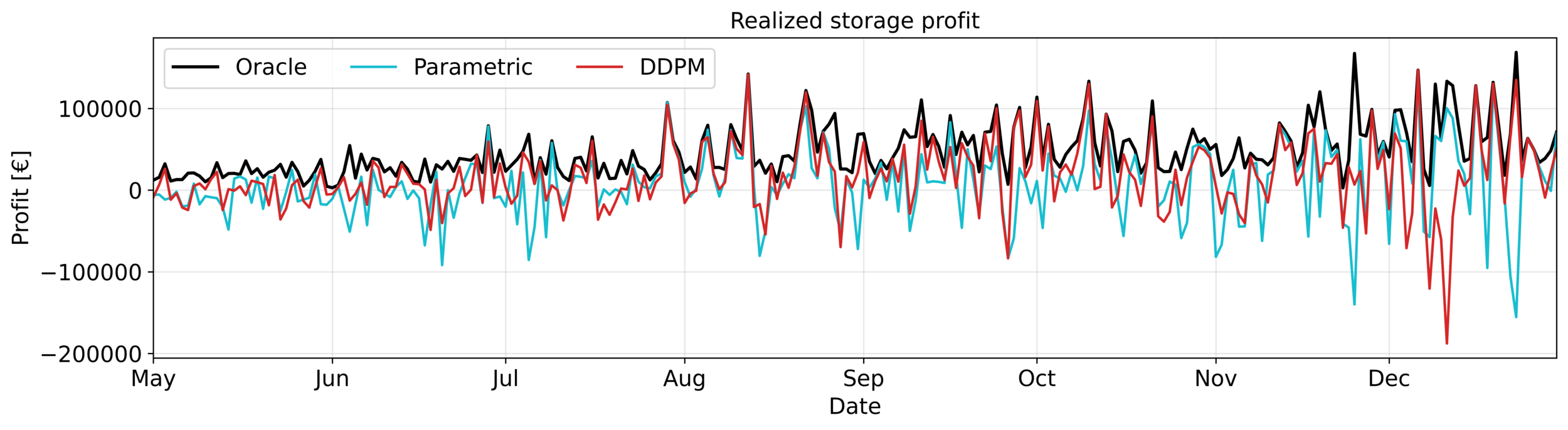}
    \caption{Daily realized storage profit.}
    \label{fig:IMG_Realized_StorageProfit}
\end{figure}

\begin{figure}[H]
    \centering        
    \includegraphics[width=1\textwidth]{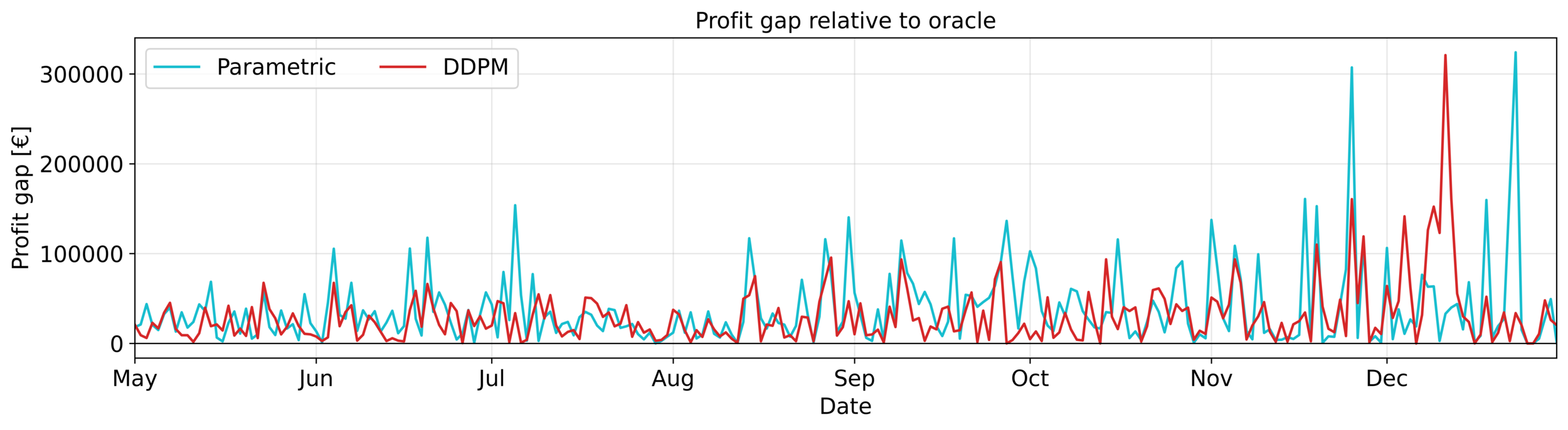}
    \caption{Daily profit gap relative to the oracle benchmark.}
    \label{fig:IMG_ProfitGap}
\end{figure}

\begin{figure}[H]
    \centering        
    \includegraphics[width=1\textwidth]{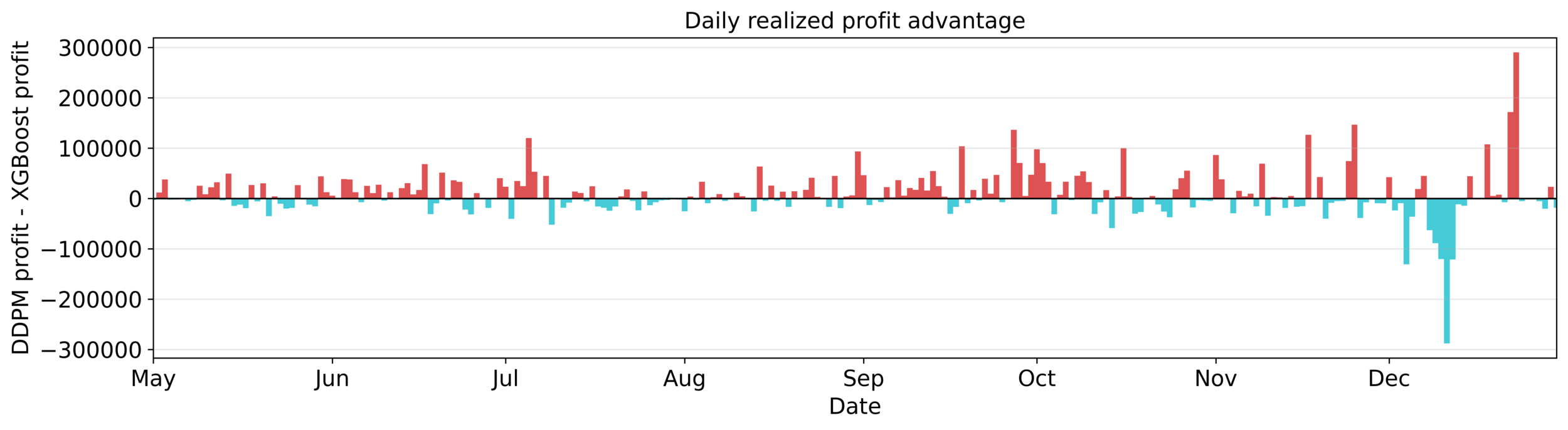}
    \caption{Daily realized profit advantage of the generative strategy over the parametric strategy. Positive values indicate higher realized profit for the generative strategy; negative values indicate higher realized profit for the parametric strategy.}
    \label{fig:IMG_Apply_ProfitAdvantage}
\end{figure}

Figure \ref{fig:IMG_Realized_StorageProfit} reports the daily realized storage profit for the oracle strategy and for the two forecast\char45 based strategies. Figure \ref{fig:IMG_ProfitGap} shows the corresponding profit gap relative to the oracle strategy, where smaller values indicate better realized performance. Figure \ref{fig:IMG_Apply_ProfitAdvantage} reports the daily realized profit difference between the generative\char45 based and parametric\char45 based strategies. Positive values indicate higher realized profit for the generative\char45 based strategy, while negative values indicate higher realized profit for the parametric\char45 based strategy. The daily advantage varies substantially over the test period, but the generative\char45 based strategy performs better on a slightly larger share of dates. The storage experiment suggests that curve accuracy should not be evaluated only through global reconstruction errors. For price\char45 maker storage, the most relevant part of the curve is the neighborhood of the clearing volume, where the agent's actions modify the residual demand faced by the market. The results indicate that even a model with less stable plateau recovery can still generate better storage decisions if it captures the local elastic structure around the clearing point more accurately, as is the case for the DDPM implementation of the generative model.

\section{Implied clearing prices}
\label{Sec:Implied price computation}

This section evaluates the implied clearing prices induced by the forecasted supply and demand curves as an additional diagnostic of forecast quality. Although the models predict curves rather than prices directly, each predicted supply\char45 demand intersection defines a market\char45 clearing price. For each date and hour, the parametric model yields a single implied price, while the generative model yields a distribution of implied prices across generated supply\char45 demand pairs.

\begin{figure}[H]
    \centering
    \begin{subfigure}[b]{1\textwidth}
        \includegraphics[width=1\textwidth]{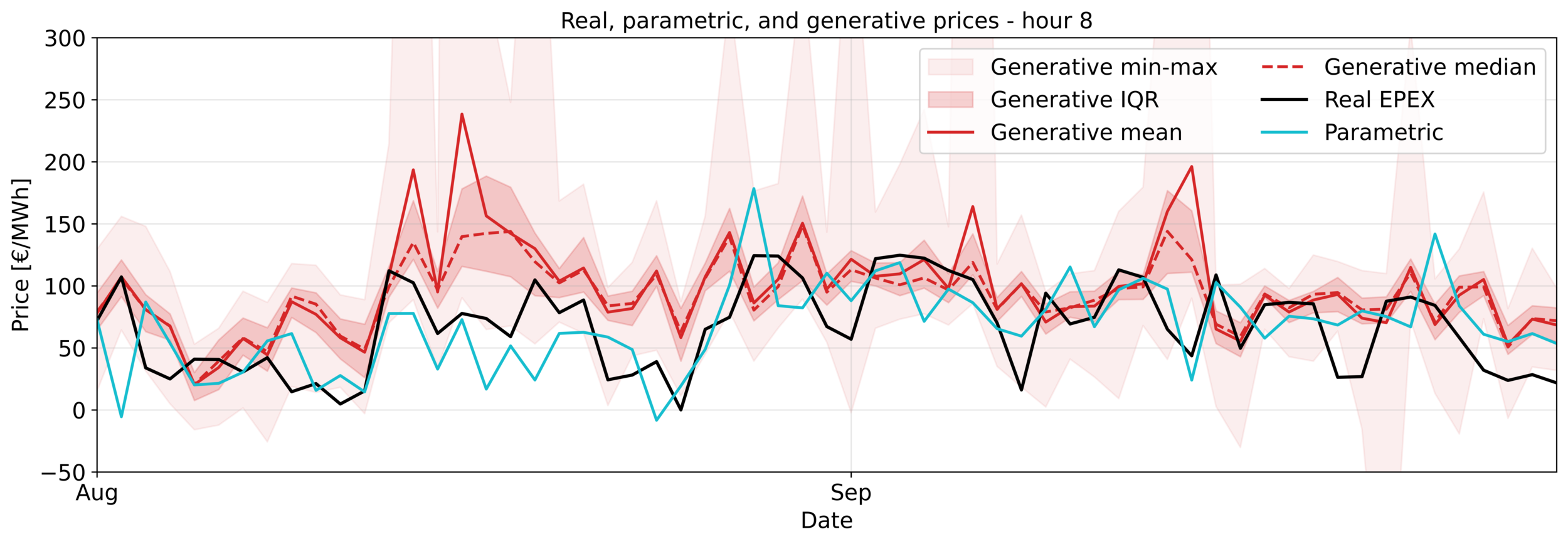}

    \end{subfigure}

    \centering
    \begin{subfigure}[b]{1\textwidth}
        \includegraphics[width=1\textwidth]{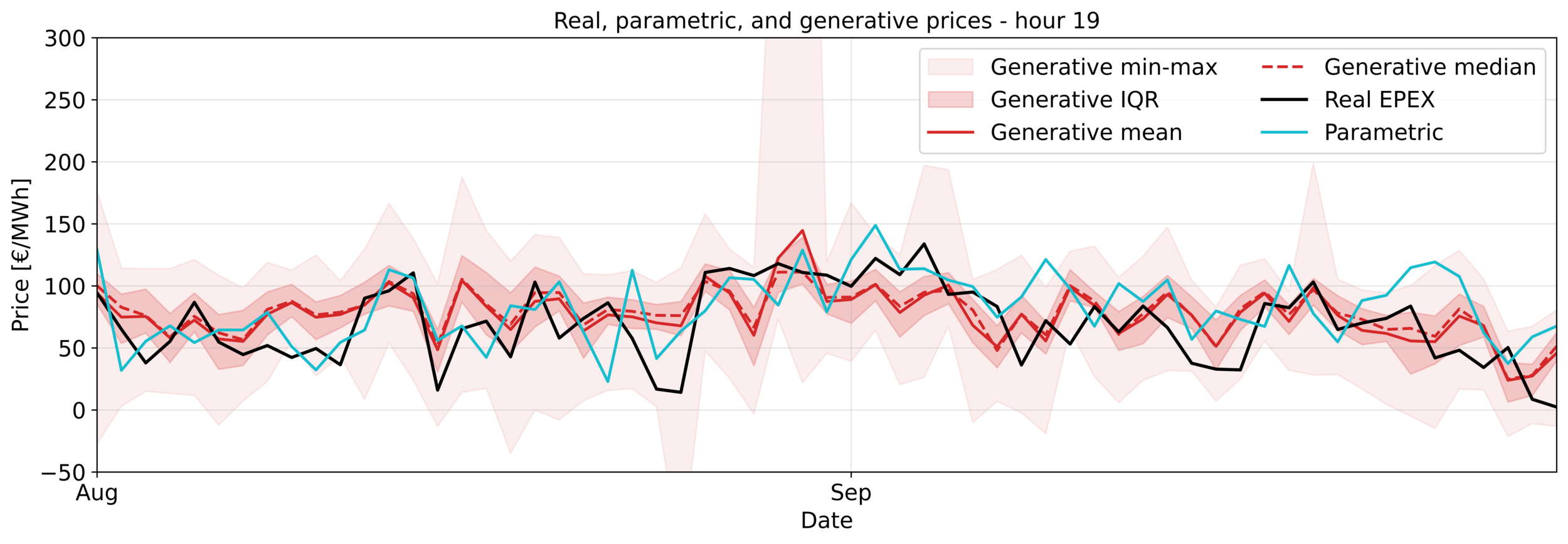}
    \end{subfigure}   
    
    \caption{Realized and forecast\char45 implied market\char45 clearing prices for hours $8$ and $19$ over August and September. The generative forecast is summarized by its min\char45 max range, interquartile range (IQR), mean, and median across generated samples.}
    \label{fig:IMG_Price_Intersections_h8h19}
\end{figure}

Figure \ref{fig:IMG_Price_Intersections_h8h19} displays the implied prices for hours $8$ and $19$ over the test dates of August and September 2024. For the generative model, the implied price distribution is computed from $K=30$ generated curve pairs. We represent this distribution by its interquartile range, together with its mean and median.

The generative price forecasts exhibit higher variability than the parametric ones. This is expected, since the DDPM implementation produces probabilistic curve scenarios rather than a single curve realization. On several dates, the generated price distribution does not exactly match the realized EPEX price at the mean or median. Nevertheless, the realized price often remains within the sample range of generated implied prices and, in many cases, within or close to the interquartile band. This suggests that the generative model captures a plausible set of market\char45 clearing outcomes, even when the central forecast misses the exact realized value.

The two forecasts should therefore be read differently. The parametric model gives a point price prediction, whereas the generative model gives a distribution of implied prices. Hence, for the generative model, performance is not only measured by the mean or median forecast, but also by whether the realized price lies within the generated price range. The wider bands reflect uncertainty in the generated order\char45 level structure and the resulting market intersection.

The price\char45 intersection exercise confirms the pattern observed in the storage application. The generative model is more volatile in price space, partly because small local changes in the elastic region can move the intersection price substantially. However, this same sensitivity is also what makes the generative forecasts useful for price\char45 maker storage decisions: they provide a richer description of the local curve geometry around the clearing point, whereas the parametric model gives a smoother and more stable point estimate.

\section{Conclusions and Future Research Directions}
\label{Sec:ConclusionsAndFuture}

This paper developed two methodologies for modelling aggregated supply and demand curves in the EPEX SPOT Day\char45 Ahead market. The first is a low\char45 dimensional parametric representation, implemented with eXtreme Gradient Boosting, which summarizes each curve through plateau levels, elastic\char45 region boundaries, and polynomial coefficients. The second is a high\char45 dimensional order\char45 level representation, implemented with conditional Denoising Diffusion Probabilistic Models, which describes curves through price arrivals and volume\char45 increment marks. While the parametric model provides deterministic point forecasts, the generative model samples from a conditional distribution of plausible curves.

The empirical results show that the two approaches capture different aspects of the curve structure. The parametric model recovers the plateau regions more accurately, especially at high price levels, because these regions are explicitly encoded in its parametrization. However, this low\char45 dimensional structure tends to smooth the transition through the elastic region and can miss the sharp slopes observed in empirical curves. The generative model exhibits the opposite behavior. By modelling the order\char45 level structure, it better reproduces the transition between inelastic and elastic regimes and captures steep local changes around the elastic segment. Its main limitation is the higher variability of the generated plateau levels, which can lead to overestimation or underestimation once the curve reaches the high\char45 price plateau.

This distinction is reflected in the price\char45 maker storage application. Since the storage agent's price impact is determined by local changes around the market\char45 clearing volume, an accurate representation of the elastic region is particularly important for decision performance. The DDPM\char45 based implementation benefits from its finer order\char45 level description of this region, despite its less stable plateau recovery. In the out\char45 of\char45 sample storage experiment, the generative strategy achieves higher realized profits and smaller gaps to the oracle benchmark than the parametric strategy. This illustrates the value of distributional curve forecasts for applications in which price impact, uncertainty, and local market sensitivity matter.

The implied price computation provides a complementary diagnostic. Although neither methodology is designed primarily for price forecasting, predicted supply and demand curves induce market\char45 clearing prices through their intersections. The resulting price forecasts capture broad price patterns but are not uniformly accurate, especially for the generative model, whose induced price distribution can be volatile across samples. This behavior is expected, since small local changes in the elastic region may produce large changes in the intersection price. Nevertheless, the realized price often lies within the sample range of the generated implied prices, even when the mean or median implied price does not match it exactly.

Several extensions follow naturally from this work. First, the generative representation can be extended from the selected block of eight delivery hours to the full 24\char45 hour cycle. Second, the treatment of plateau levels in the generative model should be refined, for example by adding constraints, post\char45 processing, or separate modelling components for inelastic regions. Third, the induced distribution of market\char45 clearing prices could be calibrated more systematically, turning curve generation into a probabilistic price\char45 forecasting tool. Finally, a systematic benchmark against other generative frameworks, including Generative Adversarial Networks, Variational Autoencoders, and Normalizing Flows, is part of ongoing work. Such a benchmark will clarify when distributional curve models provide practical value beyond point forecasts, particularly in decision problems where local price impact is more relevant than global curve reconstruction accuracy.

\appendix

\section{Fuel prices computation}
\label{subsec:Fuel prices computation}

The short\char45 run marginal cost of electricity generation from fuel $\mathrm{F}$ in [\EUR/MWh] is computed by the formula
\[
C_{\mathrm{F}}
\;=\;
P_{\mathrm{F}}\,h_{\mathrm{F}}
\;+\;
e_{\mathrm{F}}\,\tau_{\mathrm{CO}_2},
\]
where
\begin{itemize}
  \item \(P_{\mathrm{F}}\) is the commodity price of fuel $\mathrm{F}$ (e.g.\ \EUR/ton for coal, \EUR/barrel for oil, or \EUR/MWh for gas),
  \item \(h_{\mathrm{F}}\) is the heat rate (units of fuel input per MWh of energy output),
  \item \(e_{\mathrm{F}}\) is the CO\(_2\) emission factor in [tCO\(_2\)/MWh],
  \item \(\tau_{\mathrm{CO}_2}\) is the CO\(_2\)\char45 tax in [\EUR/tCO\(_2\)].
\end{itemize}
For example, the coal\char45 based cost becomes
\begin{equation*}
C_{\mathrm{coal}}
= P_{\mathrm{coal}}\times0.42 \;+\; 0.986\,\tau_{\mathrm{CO}_2}.
\end{equation*}
$\tau_{\mathrm{CO}_2}$ is obtained from \hyperlink{https://www.rte-france.com/en/eco2mix/co2-emissions}{https://www.rte-france.com/en/eco2mix/co2-emissions}, while the fuel prices are extracted from \hyperlink{https://https://www.investing.com}{https://www.investing.com}. The thermal coal is priced using the API2 CIF ARA, quoted in USD per tonne. The oil is priced using the Brent Blend, quoted in USD per barrel. The gas is priced using the Dutch Title Transfer Facility (TTF), quoted in\EUR per MWh of energy in gas. As for the heat rates, we select an efficiency of 0.42, 1.5, and 2.4 for Coal, Oil, and Gas, respectively. 

\section*{Acknowledgements}
We thank Peter Tankov (CREST, ENSAE) and Roxana Dumitrescu (CREST, ENSAE) for insightful comments on an earlier draft of this paper. This study was carried out in the framework of ``Energy for Climate'' interdisciplinary research center and was supported financially by the ``Decarbonize energy'' program of the Institut Polytechnique de Paris, as well as by the FIME Research Initiative of the Europlace Institute of Finance.
\section*{Data availability}

This study uses the following data sources:
\begin{itemize}
  \item \textit{Copernicus Climate Change Service ERA5} (via the Climate Data Store): public meteorological data.
  \item \textit{EPEX SPOT} Day\char45 ahead market data (prices, volumes, aggregated curves): proprietary, obtained under a commercial license.
  \item \textit{ENTSO\char45 E} Day\char45 ahead load forecasts: public, via the ENTSO\char45 E Transparency Platform.
  \item \textit{Fuel prices} (natural gas, coal, oil): collected from \textit{Investing.com}.
  \item \textit{French CO\textsubscript{2} tax trajectory}: published by \textit{RTE}.
\end{itemize}

ERA5, ENTSO\char45 E, Investing.com, and RTE data are publicly accessible (free registration may be required). 
EPEX SPOT data are proprietary and cannot be redistributed by the authors; equivalent access can be obtained directly from EPEX SPOT/EEX under their licensing terms.

\section*{}
\noindent {\bf Declaration of generative AI and AI\char45 assisted technologies in the manuscript preparation process} During the preparation of this work the authors used ChatGPT in order to assist in language editing, rewriting paragraphs for clarity, and retrieving bibliographic information from publicly available sources. After using this tool/service, the authors reviewed and edited the content as needed and take full responsibility for the content of the published article.

\bibliographystyle{elsarticle-harv}
\bibliography{00A0_biblio}
\end{document}